\documentclass{article}

\usepackage{microtype}
\usepackage{graphicx}
\usepackage{subcaption}
\usepackage{booktabs}
\usepackage{hyperref}
\usepackage{listings}
\usepackage{siunitx}
\usepackage{xurl}

\usepackage{enumitem}
\setlist[enumerate]{topsep=0pt,parsep=0pt,itemsep=2pt}

\newcommand{\unknownchar}{\mbox{\scriptsize\ensuremath\square}}

\usepackage{iclr2026/iclr2026_conference}
\iclrfinalcopy
\usepackage{times}
\usepackage{inconsolata}
\usepackage{makecell}
\usepackage{amsmath}
\usepackage{amssymb}
\usepackage{mathtools}
\usepackage{amsthm}

\usepackage{tikz}
\usepackage{pgfplots}
\usepackage{pgfplotstable}
\pgfplotsset{compat=1.17}
\pgfplotsset{every axis plot/.append style={line width=1pt},}
\pgfplotsset{every axis/.append style={
    label style={font=\small},
    tick label style={font=\small}}
}

\DeclareMathOperator{\Lomax}{Lomax}
\newcommand{\nsp}{{n_s}} % number of sparse features
\newcommand{\nde}{{n_d}} % number of dense features
\newcommand{\xsp}{z} % sparse features
\newcommand{\xde}{x} % dense features

\title{Automated Interpretability Metrics Do Not Distinguish Trained and Random Transformers}

\author{
    Thomas Heap \\
    University of Bristol \\
    Bristol, UK \\
    \texttt{thomas.heap@bristol.ac.uk} \\
    \And
    Tim Lawson \\
    University of Bristol \\
    Bristol, UK \\
    \And
    Lucy Farnik \\
    University of Bristol \\
    Bristol, UK \\
    \And
    Laurence Aitchison \\
    University of Bristol \\
    Bristol, UK \\
}

\begin{document}
\maketitle
% \printAffiliationsAndNotice{}

\begin{abstract}
Sparse autoencoders (SAEs) are widely used to extract sparse, interpretable latents from transformer activations.
We test whether commonly used SAE quality metrics and automatic explanation pipelines can distinguish trained transformers from randomly initialized ones (e.g., where parameters are sampled i.i.d. from a Gaussian).
Over a wide range of Pythia model sizes and multiple randomization schemes, we find that, in many settings, SAEs trained on randomly initialized transformers produce auto-interpretability scores and reconstruction metrics that are similar to those from trained models.
These results show that high aggregate auto-interpretability scores do not, by themselves, guarantee that learned, computationally relevant features have been recovered.
We therefore recommend treating common SAE metrics as useful but insufficient proxies for mechanistic interpretability and argue for routine randomized baselines and targeted measures of feature `abstractness.'
\end{abstract}

% \begin{abstract}
%   Sparse autoencoders (SAEs) are an increasingly popular technique for interpreting the internal representations of transformers.
%   In this paper, we apply SAEs to `interpret' random transformers, i.e.,\ transformers where the parameters are sampled i.i.d. from a Gaussian rather than trained on text data.
%   We find that in certain circumstances, particularly for larger models, SAEs trained on random transformers produce latents with auto-interpretability scores and quality metrics that are surprisingly similar to those from trained transformers.
%   These findings highlight that aggregate auto-interpretability scores are not always reliable proxies for the complexity of learned computations.
%   Our work underscores the need for more nuanced evaluation methods and the routine use of randomized baselines in mechanistic interpretability research.
%   % Further, we find that many quantitative SAE quality metrics are broadly similar for random and trained transformers.
%   % We discuss a number of interesting questions that this work raises for the use of purely quantitative metrics of SAE quality in the context of mechanistic interpretability.
% \end{abstract}

\section{Introduction}
\label{sec:introduction}

Sparse autoencoders (SAEs) are a popular tool in mechanistic interpretability research, with the aim of disentangling the internal representations of neural networks by learning sparse, interpretable features from network activations \citep{elhage_toy_2022,sharkey_taking_2022,cunningham_sparse_2023,bricken_towards_2023}.
An autoencoder with a high-dimensional hidden layer is trained to reconstruct activations while enforcing sparsity \citep{gao_scaling_2024,templeton_scaling_2024,lieberum_gemma_2024}, with the aim of discovering the underlying concepts or `features' learned by the network \citep{park_linear_2023,wattenberg_relational_2024}.
Developing better SAEs relies on quantitative evaluation metrics like auto-interpretability scores that measure agreement between generated explanations and activation patterns \citep{bills_language_2023,paulo_automatically_2024,karvonen_saebench_2024}.

For an interpretability method to be considered robust, its evaluation metrics should distinguish features learned through training from artifacts arising from the data or model architecture.
A key sanity check is therefore to compare the method's output on a trained model against a strong null model, such as one with randomly initialized weights \citep{adebayo_sanity_2020}. 
We apply this sanity check to SAEs and find that several common quantitative metrics do not always clearly distinguish between the trained and randomized settings.
In particular, we found that SAEs trained on transformers with random parameters can yield latents with auto-interpretability scores \citep{bills_language_2023,paulo_automatically_2024} that are surprisingly similar to those from a fully trained model.

This result raises important questions about what we can glean from applying these metrics of SAE quality.
High auto-interpretability scores alone do not guarantee that an SAE has identified complex, learned computations.
Instead, such scores may sometimes reflect simpler statistical properties of the training data \citep{dooms_tokenized_2024} or architectural inductive biases that are present even without training.
Indeed, one could argue that a randomly initialized network still performs a basic form of computation, such as preserving or amplifying the sparse structure of its inputs (Section~\ref{sec:toy_models}).
From this perspective, SAEs might faithfully interpret this simple, inherent computation.
% This highlights a critical challenge: if our goal is to understand the mechanisms learned through training, our evaluation metrics must be able to distinguish features corresponding to these simple architectural priors from those that enable complex, task-oriented capabilities.

While some SAE features from trained models clearly arise from learned computation, the commonly used aggregate metrics are often insufficient for determining whether a given SAE has learned these more complex features.
% This challenges the widespread interest in understanding the mechanisms of trained models with SAEs \citep{sharkey_open_2025}, highlighting the need to develop techniques that better capture computations \citep[e.g. ][]{farnik2025jacobian}.
These results have important implications for mechanistic interpretability research.
In particular, we suggest that more rigorous methods to distinguish between artifacts and genuinely learned computations are needed, and that interpretability techniques should be carefully validated against appropriate null models.

Finally, we speculate about why these patterns might emerge.
At a high level, there are two hypotheses:
(1) the input data already exhibits superposition, and randomly initialized neural networks largely preserve this superposition; and (2) randomly initialized neural networks amplify or even introduce superposed structure to the input data (e.g., given dense input generated i.i.d. from a Gaussian).
% \begin{enumerate}
%   \item The input data is already superposed, and randomly initialized NNs largely preserve this superposition.
%   \item Randomly initialized NNs amplify or even introduce superposed structure to the input data (e.g., given dense input generated i.i.d. from a Gaussian).
% \end{enumerate}
We present toy models to demonstrate the plausibility of these hypotheses in Section~\ref{sec:toy_models} but defer conclusions as to the mechanism responsible to future work.

\section{Related Work}
\label{sec:related_work}

\paragraph{Sparse dictionary learning}

Under a different name, `superposition' in visual data is one of the foundational observations of computational neuroscience.
\citet{olshausen_emergence_1996,olshausen_sparse_1997} showed that the receptive fields of simple cells in the mammalian visual cortex can be explained as a result of sparse coding, i.e., representing a relatively large number of signals (sensory information) by simultaneously activating a small number of elements (neurons).
Coding theory offers a perspective on efforts to extract the `underlying signals' responsible for neural network activations \citep{marshall_understanding_2024}.

Sparse dictionary learning (SDL) approximates a set of input vectors by linear combinations of a relatively small number of learned basis vectors.
The learned basis is usually overcomplete: it has a greater dimension than the inputs.
SDL algorithms include Independent Component Analysis (ICA), which finds a linear representation of the data such that the components are maximally statistically independent \citep{bell_information-maximization_1995,hyvarinen_independent_2000}.
Sparse autoencoders (SAEs) are a simple neural network approach \citep{lee_efficient_2006,ng_sparse_2011,makhzani_k-sparse_2014}.
Typically, an autoencoder with a single hidden layer that is many times larger than the input activation vectors is trained with an objective that imposes or incentivizes sparsity in its hidden layer activations to try to find this structure.
A \textbf{latent} is a single neuron (dimension) in the autoencoder's hidden layer.

\paragraph{Mechanistic interpretability}

Recently, it has become common to understand `features' or concepts in language models as low-dimensional subspaces of internal model activations \citep{park_linear_2023,wattenberg_relational_2024,engels_not_2024}.
If such sparse or `superposed' structure exists, we expect to be able to `intervene on' or `steer' the activations, i.e., to modify or replace them to express different concepts and so influence model behavior \citep{meng_locating_2022,zhang_towards_2023,heimersheim_how_2024,makelov_sparse_2024,obrien_steering_2024}.

SAEs are a popular approach for discovering features, where one typically trains a single autoencoder to reconstruct the activations of a single neural network layer, e.g., the transformer residual stream \citep{sharkey_taking_2022,cunningham_sparse_2023,bricken_towards_2023}.
Many SAE architectures have been suggested, which commonly vary the activation function applied after the linear encoder \citep{makhzani_k-sparse_2014,gao_scaling_2024,rajamanoharan_jumping_2024,lieberum_gemma_2024}.
SAEs have also been trained with different objectives \citep{braun_identifying_2024,farnik2025jacobian} and applied to multiple layers simultaneously \citep{yun_transformer_2021,lawson_residual_2024,lindsey_sparse_2024}.

Besides reconstruction errors and preservation of the underlying model's performance, SAEs have been evaluated according to whether they capture specific concepts \citep{gurnee_finding_2023,gao_scaling_2024} or factual knowledge \citep{huang_ravel_2024,chaudhary_evaluating_2024}, and whether these can be used to `unlearn' concepts \citep{karvonen_evaluating_2024}.

% \citet{chaudhary_evaluating_2024} evaluate SAEs on their ability to disentangle different types of factual knowledge in language models. Using the RAVEL benchmark. 
% \citet{huang_ravel_2024} test whether SAEs can identify separate features that encode a city's country versus its continent. They find that current SAEs consistently fail at this disentanglement task, often performing worse than raw neurons. 

\paragraph{Automatic neuron description}

SAEs often learn tens of thousands of latents, which are infeasible to describe by hand.
\citet{yun_transformer_2021} find the tokens that maximally activate a dictionary element from a text dataset and manually inspect activation patterns.
Instead, researchers typically collect latent activation patterns over a text dataset and prompt a large language model to explain them \citep[e.g.][]{bills_language_2023,foote_n2g_2023}.
These methods
% The methods of manually inspecting activation patterns and auto-interpretability
have been widely adopted \citep[e.g.][]{cunningham_sparse_2023,bricken_towards_2023,gao_scaling_2024,templeton_scaling_2024,lieberum_gemma_2024}.

\citet{bills_language_2023} generate an explanation for the activation patterns of a language-model neuron over examples from a dataset, simulate the patterns based on the explanation, and score the explanation by comparing the observed and simulated activations.
This method is commonly known as auto-interpretability (as in self-interpreting).
\citet{paulo_automatically_2024} introduce classification-based measures of the fidelity of automatic descriptions that are inexpensive to compute relative to simulating activation patterns and an open-source pipeline to compute these measures.
\citet{choi_scaling_2024} use best-of-$k$ sampling to generate multiple explanations based on different subsets of the examples that maximally activate a neuron.
Importantly, they fine-tune Llama-3.1-8B-Instruct on the top-scoring explanations
% (with the highest correlations between the observed and simulated activations)
to obtain inexpensive `explainer' and `simulator' models.

\paragraph{Polysemanticity}

\citet{lecomte_what_2024} noted that neurons may become polysemantic incidentally.
A polysemantic neuron (basis dimension) of a network layer represents multiple interpretable concepts \citep{elhage_toy_2022,scherlis_polysemanticity_2023}; unsurprisingly, individual neurons in a randomly initialized network may be polysemantic.
By contrast, our work studies \emph{superposition} \citep{elhage_toy_2022,lawrencec_superposition_2024}, which pertains to the representations learned across a whole network layer as opposed to any individual neuron.
In particular, superposition allows a network layer as a whole to represent a larger number of (sparse) features than the layer has (dense) neurons by sparse coding (only a few concepts are active at a time, i.e., a given token position).

\paragraph{Training only the embeddings}

\citet{zhong2024algorithmic} showed that transformers learn surprising algorithmic capabilities when only the embeddings are trained and no other parameters.
These results demonstrate that the behavior of a randomly initialized transformer can be shaped to a surprising extent by training only a few parameters.
However, our setting is very different: besides considering SAEs, we randomize \emph{all} the parameters, including the embeddings, in our `Step-0' and `Re-randomized incl. embeddings' variants.
Our `Re-randomized excl. embeddings' variant uses pre-trained embeddings, but we do not train those embeddings with fixed, randomized weights. Instead, we freeze the pre-trained embeddings and randomize the other weights (Section~\ref{sec:results}).

\paragraph{Random transformers for board games} \citet{karvonen2024measuring} found that SAEs were considerably better at extracting meaningful structure from chess games using pre-trained transformers, as opposed to those with random weights.
However, the data from board games is wildly different from language data.
In particular, there is reason to expect that language is sparse (e.g., a particular concept such as `serendipitous' appears only rarely), and that this sparse structure is `aligned' with conceptual meaning.
In contrast, in board games, this is not necessarily true: a useful concept such as a knight fork does not necessarily turn up sparsely in board games.

\paragraph{Random one-layer transformers} \citet{bricken_towards_2023} found that auto-interpretability scores discriminated effectively between random and trained one-layer transformers.
Similarly, we found that auto-interpretability scores for randomized models were relatively low for smaller models (e.g., Pythia-70m) but that the gap was narrowed for larger models (e.g., Pythia-6.9b).
% This agrees with our finding that, it is the largest models where autointerp gives similar or even better scores for random than trained models.
% In contrast for smaller models, the autointerp scores in the randomized setting do seem to be lower than for the trained transformers.

\section{Results}
\label{sec:results}

\begin{figure*}
  \begin{center}
  \includegraphics[width=0.8\textwidth]{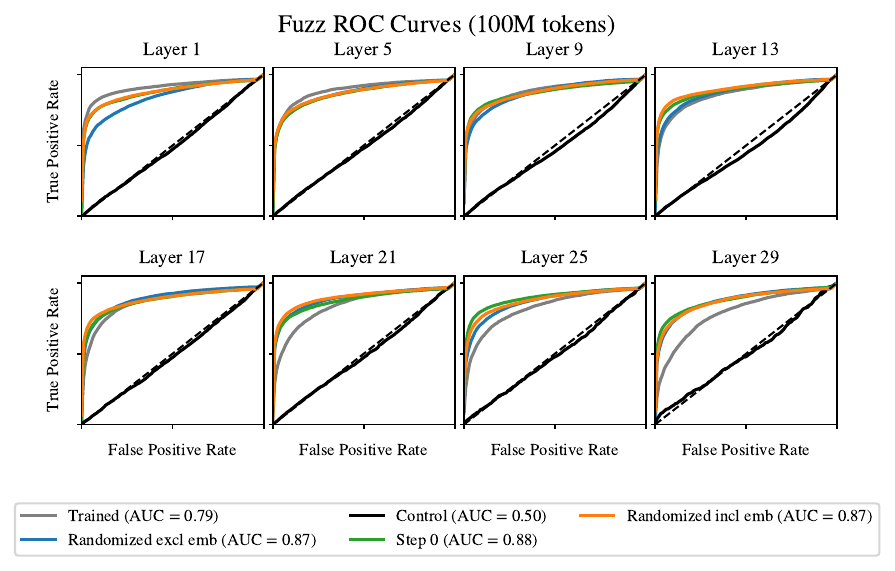}
  \end{center}
  \caption{`Fuzzing' ROC curve vs. layer for Pythia-6.9b (100 latents sampled per SAE). The trained model (gray line) and randomized variants (colored) overlap, whereas the control (black) is near chance (dotted). This suggests aggregate AUROC alone is insufficient to attribute latents to learned computation. See Figure~\ref{fig:all_metrics} for other metrics/model sizes and Appendix~\ref{app:uncertainty} for multiple random seeds.
  % ROC curves for `fuzzing' auto-interpretability for Pythia-6.9b over 100 SAE latents. These results demonstrate the similarity in performance between the SAE variants. Notably, the trained variant appears to degrade in performance for the later model layers.
  }
  \label{fig:pythia_6.9b_fuzzing}
\end{figure*}

We trained per-layer SAEs on the residual stream activation vectors of transformer language models from the Pythia suite, with between 70M and 7B parameters \citep{biderman_pythia_2023}. %as well as Gemma 2 2B and Llama 3.1 8B \citep{riviere_gemma_2024,grattafiori_llama_2024}.
We compared SAEs trained on different variants of the underlying transformers:

\begin{itemize}
    \item \textbf{Trained:} The usual, trained model.
    \item \textbf{Re-randomized incl. embeddings:} All the model parameters, including the embeddings, are re-initialized by sampling Gaussian noise with mean and variance equal to the values for each of the original, trained weight matrices.
    \item \textbf{Re-randomized excl. embeddings:} As above, except the embedding and unembedding weight matrices are not re-initialized, i.e., are the same as the original, trained model.
    \item \textbf{Step-0:} For Pythia models, the \texttt{step0} revisions are available, which are the original model weights at initialization, i.e., before any learning \citep{biderman_pythia_2023}.
    \item \textbf{Control:} The original, trained model, except where the input token embeddings are replaced at inference time by sampling i.i.d. standard Gaussian noise for each token, such that a given token does not have a consistent embedding vector. For this variant, we expect auto-interpretability to perform at the level of chance.
\end{itemize}

% \textbf{Trained}: The usual, trained model.

% \textbf{Re-randomized incl. embeddings}: All the model parameters, including the embeddings, are re-initialized by sampling Gaussian noise with mean and variance equal to the values for each of the original, trained weight matrices.

% \textbf{Re-randomized excl. embeddings}: As above, except the embedding and unembedding weight matrices are not re-initialized, i.e., are the same as the original, trained model.

% \textbf{Step-0}: For Pythia models, the \texttt{step0} revisions are available, which are the original model weights at initialization, i.e., before any learning \citep{biderman_pythia_2023}.

% \textbf{Control}: The original, trained model, except where the input token embeddings are replaced at inference time by sampling i.i.d. standard Gaussian noise for each token, such that a given token does not have a consistent embedding vector. We expect auto-interpretability to perform at the level of chance for this variant.

For our primary experiments, we trained SAEs on 100M tokens from the RedPajama dataset \citep{weber_redpajama_2024}
using an activation buffer size of 10M tokens (see Appendix~\ref{app:1b_tokens} for a subset of experiments that demonstrate similar results with SAEs trained on one billion tokens).
% \footnote{\href{https://huggingface.co/datasets/togethercomputer/RedPajama-Data-1T-Sample}{\texttt{https://huggingface.co/datasets/togethercomputer/RedPajama-Data-1T-Sample}}}
For models with fewer than 410M parameters, we trained an SAE at every layer;
% for larger models, we skipped some layers due to available compute resources.
for Pythia-1b, we trained SAEs at every second layer; and for Pythia-6.9b, we trained SAEs at every fourth layer.
%Finally, for Gemma 2 2B, and for Llama 3.2 8B, we trained at every fourth layer again.

Unless otherwise stated, we trained $k$-sparse autoencoders (also known as TopK SAEs; \citealt{makhzani_k-sparse_2014,gao_scaling_2024}), with an expansion factor of $R=64$ and sparsity $k=32$.
We confirm that our results are robust with respect to these hyperparameters by training SAEs on Pythia-160m with expansion factors equal to powers of 2 between 16 and 128, and sparsities of 16 and 32 (Figure~\ref{fig:hyper_params}).
The training implementation is based on \citet{eleutherai_sparsify_2025}; our evaluations are based on \citet{eleutherai_delphi_2025} and \citet{karvonen_saebench_2024}.

\paragraph{Auto-interpretability} 

Feature explanations that identify a concept can be input to a classifier that predicts whether the concept appears in the text inputs.
Such a classifier may be evaluted by traditional metrics, like the area under the receiver operating characteristic (ROC) curve (AUROC).
% There is a choice of classification tasks for evaluating feature explanations, e.g., `fuzzing' and `detection' scoring \citep{paulo_automatically_2024}.
\citet{paulo_automatically_2024} proposed `fuzzing' and `detection' classification tasks to evaluate feature explanations.
For `fuzzing' scoring, both positive and negative examples of tokens (i.e., with non-zero and zero activation values, respectively) for a given latent are delimited with special characters, and a language model is prompted to identify which examples have been correctly delimited for the latent given its explanation. 
For `detection', a language model is asked to identify which examples contain activating tokens for each feature.
\citet{bills_language_2023} originally proposed `simulation' scoring, based on the correlation between predicted and observed activations, but this method is expensive to compute.

Except where noted, we report `fuzzing' scores as a measure of auto-interpretability, because this measure has been demonstrated to correlate with simulation scoring \citep{paulo_automatically_2024}.
We include similar AUROC curves for the `detection' scoring method in Appendix~\ref{app:roc_curves}.
For each trained SAE (i.e., underlying model, variant, and layer), we randomly sampled 100 features to obtain auto-interpretability scores.
The implementation is based on \citet{paulo_automatically_2024}.
We use the \texttt{Meta-Llama-3.1-70B-Instruct-AWQ-INT4} model to generate explanations and make predictions (larger than the 8B models used by \citet{choi_scaling_2024} and open-source, unlike \citealt{bills_language_2023}).

\begin{figure*}
\centering\includegraphics[width=0.95\textwidth]{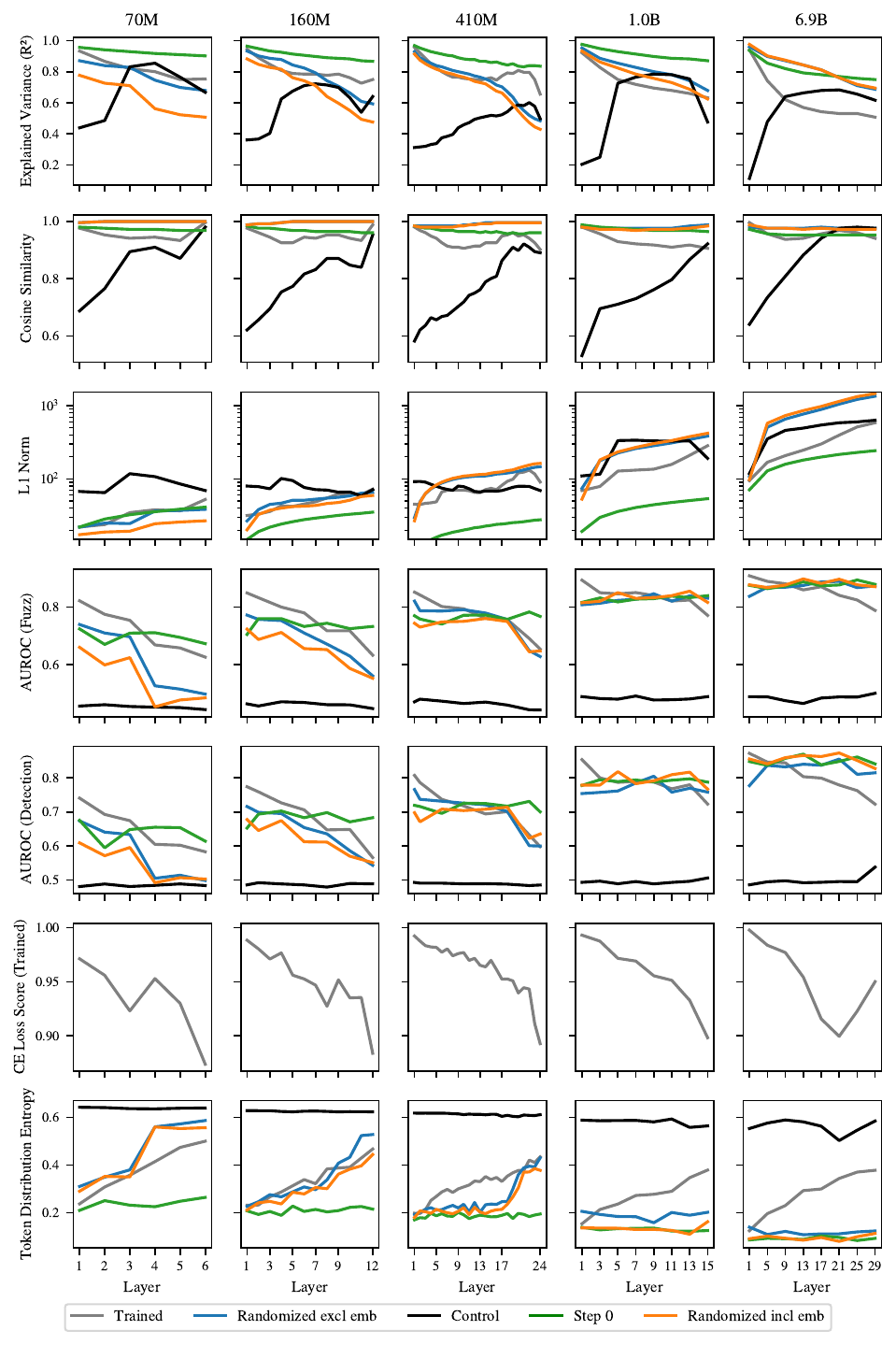}
  \caption{Comparison of sparse autoencoder performance across Pythia models (70M to 6.9B parameters). The different SAE variants show remarkably similar trends across model scales, with larger models exhibiting more consistent behavior across layers. All variants save for control achieve comparable performance despite fundamentally different initialization approaches.}
  \label{fig:all_metrics}
\end{figure*}

We found that the auto-interpretability scores were far more similar between the trained and randomized models than with the control (Figures \ref{fig:pythia_6.9b_fuzzing} and \ref{fig:all_metrics}). 
%In the 410m case all variants approach the random baseline, but this does not seem to occur in the 6.9b case. In fact comparing a larger number of model sizes (Fig. ~\ref{fig:pythia_70m_fuzzing}, \ref{fig:pythia_160m_fuzzing} and \ref{fig:pythia_1b_fuzzing} in the appendix) reveals that SAEs trained on models of size 1b and up do not show this behavior. 
The similarity between the ROC scores for trained and randomized transformers demonstrates that `fuzzing' auto-interpretability alone, applied to SAE latent explanations, may not meaningfully distinguish between these underlying models.

% \begin{figure*}[t]
%   \begin{center}
%   \includegraphics[width=0.8\textwidth]{figures/token_100M/pythia-410m-deduped_64_k32/fuzz_roc_curves_100M.pdf}
%   \end{center}
%   \caption{ROC curves for `fuzzing' auto-interpretability for Pythia-410m over 100 SAE latents. These results demonstrate the similarity in performance between the SAE variants, as well as the overall degradation in performance as the layer index increases. The auto-interpretability scores here fail to distinguish between trained and randomized models.}
%   \label{fig:pythia_410m_fuzzing}
% \end{figure*}

\paragraph{Evaluation} 

We considered standard SAE evaluation metrics alongside the auto-interpretability AUROC for Pythia models with between 70M and 6.9B parameters.
As above, we broadly found that the randomly initialized and re-randomized models (Figure~\ref{fig:all_metrics}; blue, green, orange lines) were more similar to the trained model (Figure~\ref{fig:all_metrics}; gray lines) than to our control (Figure~\ref{fig:all_metrics}; black lines).
% We can pull out some interesting, lower-level patterns from these plots, though we are in the most part unable to offer any explanations.

Notably, the cosine similarity between the original activation and the SAE reconstruction and explained variance are often far lower for the random control than the other models, and its reconstruction errors tend to increase across layers while the remaining variants decrease. 
For the random control, this can perhaps be explained by the fact that a Gaussian is the highest entropy distribution with fixed mean and variance \citep{jaynes2003probability}; we speculate that Gaussian vectors are the `least structured', in some sense, and thus hardest for SAEs to reconstruct.
As Gaussian-distributed activations are propagated through successive layers, we would expect the activations to become less Gaussian and perhaps more `sparse', i.e., easier to reconstruct (Section~\ref{sec:toy_models}).

Interestingly, the randomized variants (blue and orange lines) are more similar to the trained model than the variant at initialization (green line).
This is especially evident if we look at the $L^1$ norm values in larger models.
We speculate that this pattern arises because parameter norms may differ greatly between a trained model and its state at initialization. 
In contrast, our randomization procedure was specifically designed to preserve parameter norms with respect to the trained model.
The scale of parameters at different layers may be important, e.g., to control the growth of activations as they progress through the residual stream \citep{liu-etal-2020-understanding}.
In the AUROC plots, we find that for all but the control variant, AUROC increases with model size.
We speculate that features become more specific as SAE size increases: in smaller SAEs, each latent must explain more of the input, making classification tasks easier for larger SAEs.

Figure~\ref{fig:all_metrics} (row five) shows the cross-entropy (CE) loss score, or loss recovered, against model layer.
This is the increase in the loss when the original model activations are replaced by their SAE reconstructions, divided by the increase when the activations are replaced by zeros (`ablated').
The results show that the `trained' variant SAEs perform similarly to others from the literature \citep[e.g.,][]{kissane_interpreting_2024,rajamanoharan_improving_2024,mudide2024efficient}.
Importantly, the CE loss score only makes sense for the trained variant: for any of the randomized variants, the loss is very poor, regardless of whether the original or reconstructed activations are used.

\paragraph{Latent explanation complexity}

Despite sometimes similar auto-interpretability scores and evaluation metrics, we had expected that SAEs applied to trained vs. randomized transformers would discover qualitatively different features.
In particular, we expected SAEs trained on the randomized variants to learn relatively simple features based on characteristics of the input text, but not more complex, abstract features as with trained transformers \citep{templeton_scaling_2024}.
% Given the difficulty of qualitatively evaluating many features, we simply quantified this notion in terms of the distribution of latent activations over token IDs.
For qualitative examples, we provide a random sample of features and the corresponding maximally activating dataset examples for each variant of Pythia-6.9b in Appendix~\ref{app:6.9b_example_features}, and more detailed information in Appendix~\ref{app:feature_dashboards}.

Anecdotally, we have observed that a significant proportion of SAE latents have non-zero activations only on a single token or a small number of distinct tokens within a text dataset \citep[e.g.,][]{lin_announcing_2024,dooms_tokenized_2024}.
Hence, a simple measure of the complexity of an SAE latent given a set of maximally activating examples is the degree to which the latent activates on a single token ID or multiple distinct IDs.
Specifically, we quantify the number of token IDs in terms of the entropy of the observed distribution of latent activations over tokens: the greater the entropy, the more `spread out' the latent activations, and the less token-specific the latent.
We take this distribution to be the total latent activation per token across the set of maximally activating examples used to generate explanations for auto-interpretability.
% We provide examples of high and low entropy features in appendix ~\ref{app:entropy_8_high_entropy} and ~\ref{app:entropy_8_low_entropy}.
% We provide a breakdown of entropy across layers and variants in appendix ~\ref{app:entropy_breakdown}, and scatter plots showing the relationship between the ``Fuzzing'' AUROC score and entropy in appendix ~\ref{app:ent_fuzz_auroc}.
We show the relationship between entropy and `fuzzing' AUROC score for individual latents in Appendix~\ref{app:ent_fuzz_auroc}.
% While the distribution of latent activations over token IDs is a crude approximation of the complete distribution, it serves to quantify the diversity of the token activations commonly used to generate latent explanations in the auto-interpretability pipeline.
% We expected this entropy-based measure to filter out latents whose explanation has the form `instances of the word X in various contexts' or similar.

We include the entropy of the observed distributions of latent activations over token IDs in the last row of Figure~\ref{fig:all_metrics}.
The negative control variant displays a consistently high entropy, which is to be expected given that the embedding for a given token ID is sampled i.i.d. from a Gaussian on each occurrence of the token, i.e., a token does not have a consistent embedding vector (Section~\ref{sec:results}).
For the trained variant, the entropy increases across layers, i.e., the further into the model, the less likely the maximally activating examples for each latent contain activations concentrated on a single token.
This is also expected: at later layers, we expect more abstract features that are less similar to token embeddings.
Finally, the entropy for randomized models tends to be lower than for either the trained or control variants, indicating that latents are activated specifically at one or a few IDs.

In combination with the preceding results, this suggests that standard SAE quality and auto-interpretability metrics are missing an important aspect of SAE features: their `abstractness'.
While the token distribution entropy is not a direct measure of `abstractness', it suggests that the randomized variants, viewed in the context of their similar auto-interpretability scores to the trained variant, remain able to learn simple, single-token features.
However, unlike the trained variant, the features of the randomized variants do not become more complex as the layer index increases.

\section{A toy model of superposition in random networks}
\label{sec:toy_models}

We speculated in Section~\ref{sec:introduction} that the apparently high degree of sparsity and interpretability in the activations of randomized transformers might be because the input data exhibits superposition, which neural networks preserve, or neural networks somehow amplify or even introduce superposition into the input data.
In this section, we examine both possibilities through the lens of toy models.
We find some evidence to support each potential cause, but we leave the question of which predominates in the case of randomized transformers and the results detailed in the main text to future work.

\begin{figure}
    \centering
    \begin{subfigure}[t]{0.24\linewidth}
        \centering
        \includegraphics[width=\textwidth]{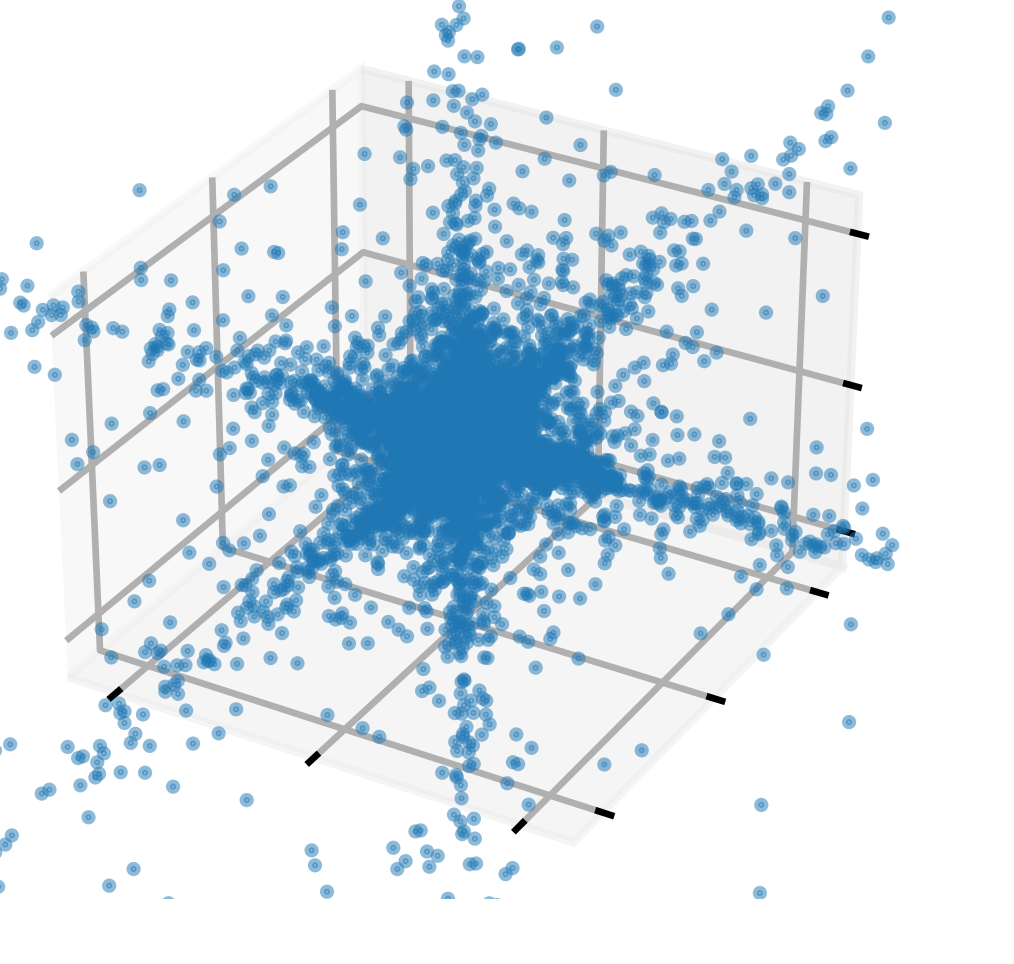}
        \caption{Superposed inputs}
        \label{fig:sparse_inputs}
    \end{subfigure}
    \hspace{-0.025\linewidth} 
    \begin{subfigure}[t]{0.24\linewidth}
        \centering
        \begin{tikzpicture}
            \begin{axis}[
                scale only axis,
                width=1in,height=1in,
                enlargelimits=false,
                xmin=-25,xmax=+25,
                ymin=-25,ymax=+25,
                xtick align=outside,
                ytick align=outside,
                tick pos=left,
            ]
            \addplot graphics [
                xmin=-25,xmax=+25,
                ymin=-25,ymax=+25,
            ] {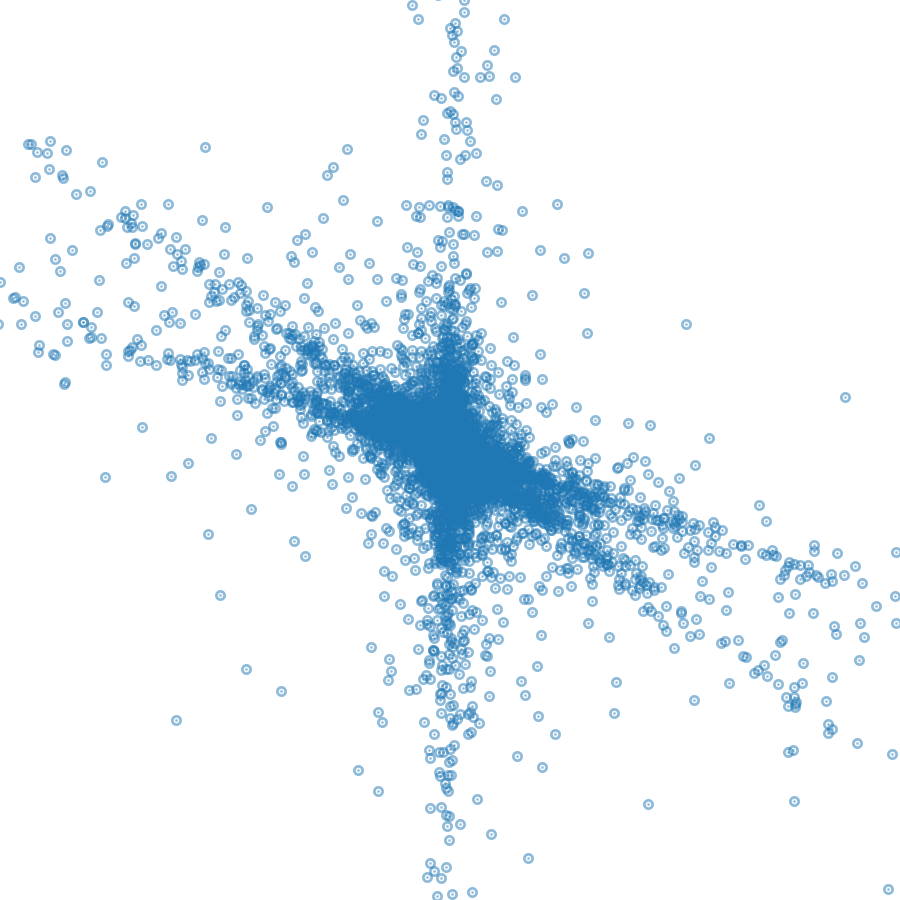};
            \end{axis}
        \end{tikzpicture}
        \caption{MLP inputs}
        \label{fig:dense_inputs}
    \end{subfigure}
    \hfill % Normal spacing
    \begin{subfigure}[t]{0.24\linewidth}
        \centering
        \begin{tikzpicture}
            \begin{axis}[
                scale only axis,
                width=1in,height=1in,
                enlargelimits=false,
                xmin=-12.5,xmax=+25,
                ymin=-12.5,ymax=+25,
                xtick align=outside,
                ytick align=outside,
                tick pos=left,
            ]
            \addplot graphics [
                xmin=-12.5,xmax=+25,
                ymin=-12.5,ymax=+25,
            ] {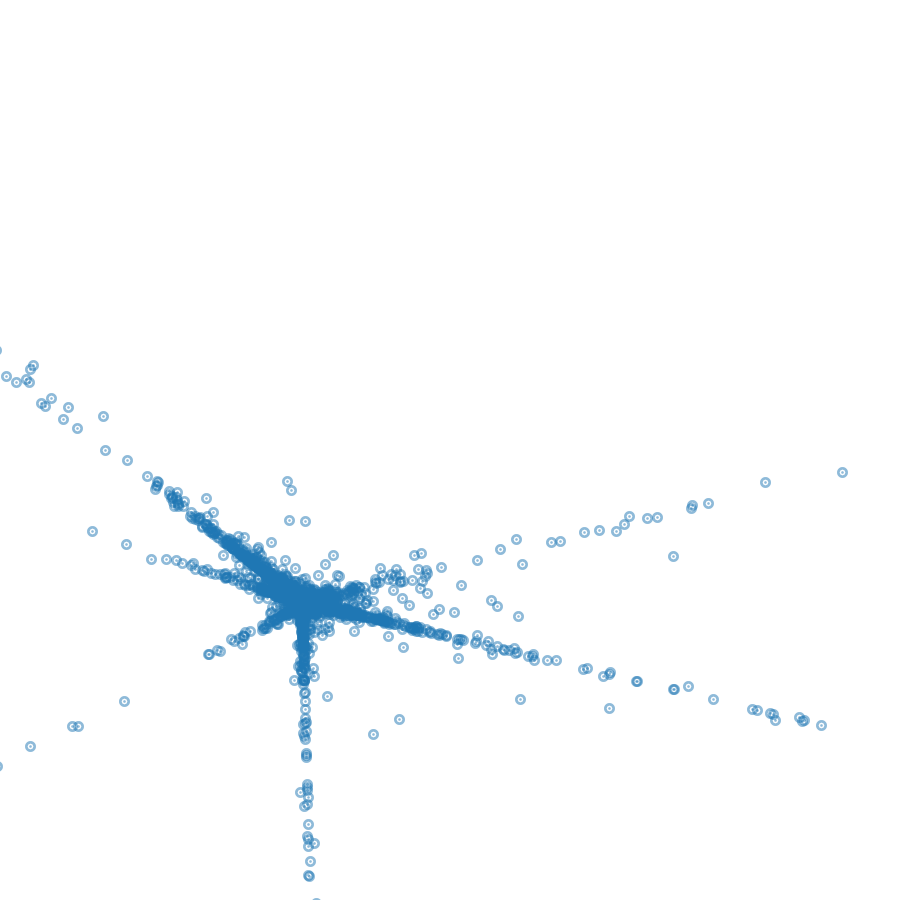};
            \end{axis}
        \end{tikzpicture}
        \caption{MLP outputs}
        \label{fig:dense_outputs}
    \end{subfigure}
    \hspace{0.02\linewidth} 
    \begin{subfigure}[t]{0.24\linewidth}
        \centering
        \includegraphics[width=\textwidth]{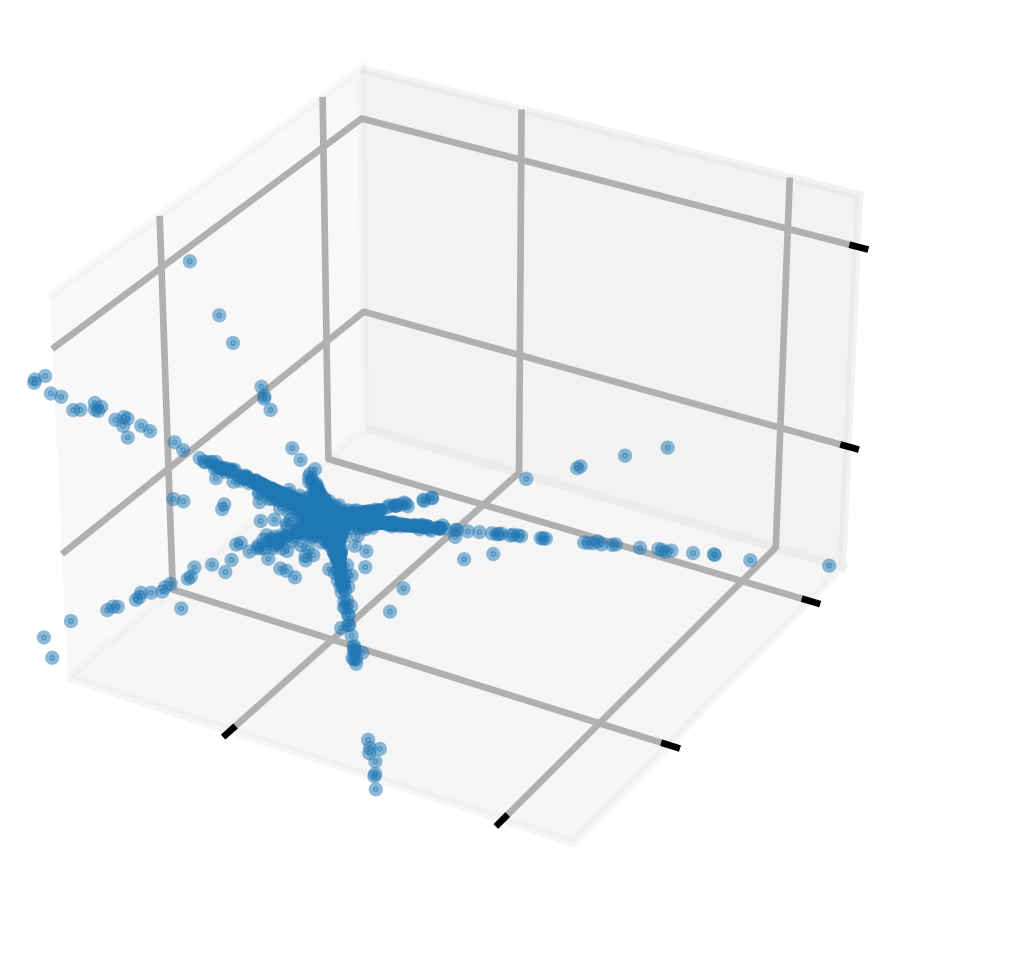}
        \caption{Superposed outputs}
        \label{fig:sparse_outputs}
    \end{subfigure}
    \caption{
        An example of the effect of a randomly initialized neural network on superposed input data.
        We take 10K samples of $\nsp=3$ sparse input features from a Lomax distribution with shape $\alpha=1$ and scale $\lambda=1$ and project these to $\nde=2$ dense input features by an i.i.d. standard normal matrix.
        Then, we pass the dense outputs to a two-layer MLP with ReLU activation and hidden size of $4\nde$ and recover $\nsp=3$ sparse outputs by the inverse of the previously generated projection matrix.
    }
    \label{fig:toy_outputs}
\end{figure}

\subsection{Matrix multiplications preserve superposition}
\label{sec:toy_models_0}

First, we consider a simplified model to demonstrate that multiplication by a weight matrix $W$ preserves superposition.
Imagine that we generate superposed input data $\xde$ by first generating $\nsp$ i.i.d. `sparse' features $\xsp$ from a heavy-tailed Lomax distribution $\xsp \sim \Lomax(\alpha, \lambda)$.
We can project the higher-dimensional, sparse $\xsp$ down to lower-dimensional, dense $\xde$ with a matrix $D$, then add Gaussian noise with a small variance $\Sigma$,
$\xde \sim \mathcal{N}(\xde; D\xsp, \Sigma)$.
% \begin{align}
% \xde \sim \mathcal{N}(\xde; D\xsp, \Sigma).
% \end{align}
Importantly, if we multiply $\xde$ by some matrix $W$, then $\xde'=W\xde$ is \emph{also} superposed: it is generated by the same model as $\xde$, except with different noise covariances and mappings from $\xsp$ to $\xde$, namely
$\xde' \sim \mathcal{N}(\xsp; WD\xsp, W \Sigma W^T)$.
% \begin{align}
% \xde' \sim \mathcal{N}(\xsp; WD\xsp, W \Sigma W^T).
% \end{align}

% \paragraph{Multi-layer perceptrons can preserve superposition}

We can see that the same intuition might extend to neural networks with nonlinearities by visualizing the results of passing the dense activations through a simple feed-forward network (MLP).
Figure~\ref{fig:toy_outputs} shows an example where $\nsp=3$ sparse features are projected down to $\nde=2$ dense features, and the MLP outputs appear superposed despite the non-linearity.
Moreover, it suggests that NNs might amplify superposition rather than only preserving it: comparing the inputs (Figures~\ref{fig:sparse_inputs} and \ref{fig:dense_inputs}) to the outputs (Figures~\ref{fig:sparse_outputs} and \ref{fig:dense_outputs}), there are fewer points between the `arms' of the outputs.

\subsection{Do random NNs preserve or amplify superposition?}
\label{sec:toy_models_1}

We investigated this suggestion by generating toy data with the same procedure as \citet{sharkey_taking_2022}, i.e., sampling ground-truth features on a hypersphere and generating correlated feature coefficients such that only a small number are active (Appendix~\ref{app:toy_models_data}).
We then passed these inputs to a two-layer MLP at initialization and trained SAEs on both the inputs and outputs individually.
As a control, we used Gaussian-distributed inputs with a mean and standard deviation equal to the superposed toy data.
We used standard SAEs with an $L^1$ sparsity penalty (Appendix~\ref{app:toy_models_training}).

\begin{figure}[t]
    \centering
    \includegraphics{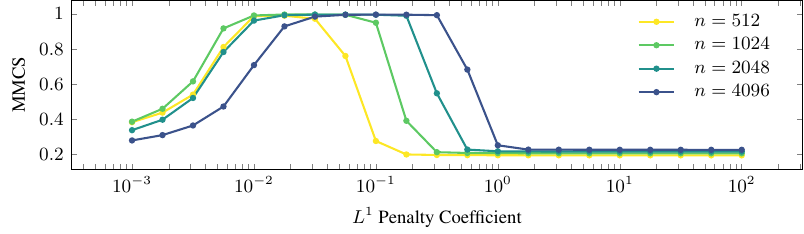}
    \caption{
        The mean max cosine similarity (MMCS) between the features learned by a standard SAE (decoder weight vectors) and the data-generating features against the $L^1$ penalty coefficient in the training loss, following \citet{sharkey_taking_2022}.
        There is a `Goldilocks zone' where SAEs near-perfectly recover the data-generating features, given enough latents to represent them.
    }
    \label{fig:l1_coef_mmcs}
\end{figure}

\begin{figure}
    \centering
    \includegraphics[width=\textwidth]{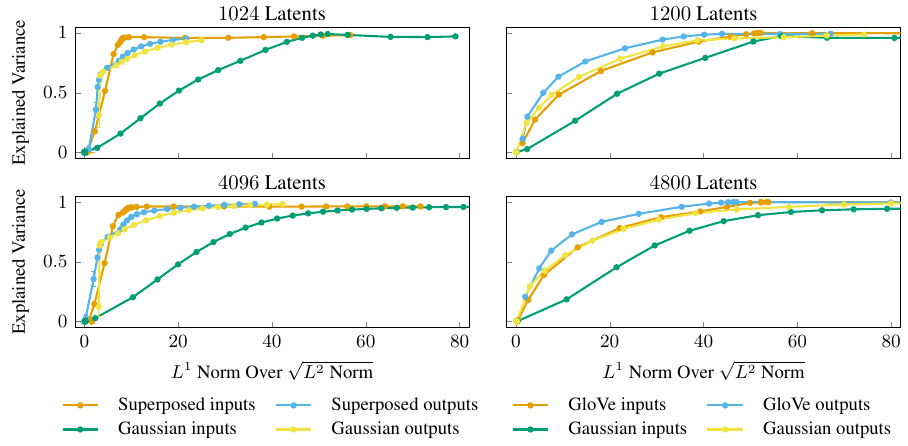}
    \begin{subfigure}[b]{0.49\textwidth}
        \centering
        \caption{Toy datasets following \citet{sharkey_taking_2022}.}
        \label{fig:sharkey_l1_over_sqrt_l2_onecolumn}
    \end{subfigure}
    % \hfill
    \begin{subfigure}[b]{0.49\textwidth}
        \centering
        \caption{$300$-dim. GloVe vectors \citep{pennington_glove_2014}.}
        \label{fig:glove.6B.300_onecolumn}
    \end{subfigure}
    \caption{Pareto frontiers of the explained variance against the $L^1$ norm divided by the square root of the $L^2$ norm (sparsity) for datasets perhaps exhibiting superposition, Gaussian controls with the same mean and variance, and the corresponding outputs when these are passed to a randomly initialized two-layer MLP (Section~\ref{sec:toy_models_1}. Each point denotes a choice of $L^1$ penalty coefficient.}
\end{figure}

Following \citet{sharkey_taking_2022}, we confirmed that SAEs can recover the ground-truth features that generated the data (Figure~\ref{fig:l1_coef_mmcs}).
In particular, we measured the mean max cosine similarity (MMCS): for every data-generating feature, we found its maximum cosine similarity with the features learned by the SAE (its decoder weight vectors) and took the average over data-generating features.
However, the MMCS only applies to the MLP inputs, where we have access to the data-generating features -- a different approach is required to analyze the MLP outputs.
To this end, we took the ability of SAEs to achieve low reconstruction error with high sparsity as a proxy for the degree to which the training data exhibits superposition.
Specifically, we vary the $L^1$ penalty coefficient to obtain Pareto frontiers of the explained variance against sparsity measures (Figure~\ref{fig:sharkey_l1_over_sqrt_l2_onecolumn}).

As expected, we found that SAEs achieved much greater sparsity at a given level of explained variance for the superposed inputs relative to the Gaussian control (Figure~\ref{fig:sharkey_l1_over_sqrt_l2_onecolumn}; orange and blue-green).
Interestingly, the difference between the superposed outputs, i.e., the outputs of the MLP given the superposed inputs, and the Gaussian outputs is much smaller, with only slightly greater sparsity at a given level of explained variance.
This suggests that the outputs of randomly initialized MLPs have a relatively high level of sparsity insensitive to the input distribution.
%This suggests that random neural networks tend to amplify the degree to which superposition is exhibited by their input data.
We consider other sparsity measures and hyperparameters in Appendix~\ref{app:toy_models}.

% \begin{figure}[t]
%     \centering
%     \includegraphics{figures/sharkey_l1_over_sqrt_l2_onecolumn.pdf}
%     \caption{
%         Pareto frontiers of the explained variance against the $L^1$ norm divided by the square root of the $L^2$ norm (sparsity) for toy datasets generated to exhibit superposition, Gaussian controls with the same mean and variance, and the corresponding outputs when these are passed to a randomly initialized two-layer MLP (Section~\ref{sec:toy_models_1}).
%         Each point denotes a choice of $L^1$ penalty coefficient, and the error bars denote the standard error in the mean taken over random seeds.
%     }
%     \label{fig:sharkey_l1_over_sqrt_l2_onecolumn}
% \end{figure}

\subsection{Do token embeddings exhibit superposition?}
\label{sec:toy_models_2}

To the extent that randomly initialized neural networks preserve or amplify superposition, our results (Section~\ref{sec:results}) could be explained by the degree to which the inputs to transformer language models exhibit superposition.
We study this question by applying the procedure described in Section~\ref{sec:toy_models_1} to language data.
In particular, we train SAEs on pre-trained GloVe word vectors, the embedding matrices of Pythia models, the results of passing these inputs to a randomly initialized two-layer MLP, and Gaussian controls.
The setup is unchanged from Section~\ref{sec:toy_models_1}, except that the number of data points is fixed by the number of word embeddings or tokens, and we use a single random seed.

% \begin{figure}[t]
%     \centering
%     \includegraphics{figures/glove.6B.300_onecolumn.pdf}
%     \caption{
%         Pareto frontiers of the explained variance against the $L^1$ norm divided by the square root of the $L^2$ norm (sparsity) for $300$-dimensional GloVe word vectors \citep{pennington_glove_2014}, Gaussian controls with the same mean and variance, and the corresponding outputs when these are passed to a randomly initialized two-layer MLP (Section~\ref{sec:toy_models_1}).
%         Each point denotes a choice of $L^1$ penalty coefficient.
%     }
%     \label{fig:glove.6B.300_onecolumn}
% \end{figure}

We find that the gap between the Pareto frontiers of the GloVe word vectors and the corresponding Gaussian controls (Figure~\ref{fig:glove.6B.300_onecolumn}) is smaller than that observed for the toy superposed datasets described in Section~\ref{sec:toy_models_1} (Figure~\ref{fig:sharkey_l1_over_sqrt_l2_onecolumn}).
More interestingly, we again see that the Pareto frontiers for both inputs improve when they are passed to a randomly initialized two-layer MLP, emphasizing the possibility that random NNs `sparsify' their inputs (i.e., increase the degree of apparent superposition).

\section{Limitations}
\label{sec:limitations}
% In this work, we demonstrate that SAEs can meaningfully interpret randomly initialized models.
In this work, we demonstrate that auto-interpretability measures can produce apparently meaningful, interpretable results for SAEs trained on randomly initialized models, which are unlikely to exhibit computationally interesting features.
Given the impossibility of testing across all datasets and model architectures, we strategically focused on the Pythia family of models, widely adopted in mechanistic interpretability research \citep[e.g.,][]{paulo2025sparseautoencoderstraineddata, ghilardi2024efficienttrainingsparseautoencoders, mueller2024missedcausesambiguouseffects}, and the RedPajama-V2 dataset, representing typical pre-training data for language models and SAEs.

While we used the default model for generating explanations in the EleutherAI auto-interpretability framework \citep{eleutherai_delphi_2025}, exploring alternative models could yield valuable insights into aggregate behaviors and the quality of generated explanations.
% Our intent was not to establish complete equivalence between SAEs trained on random versus trained transformers, only that SAEs can be used to interpret random transformers and highlight the significant similarities that emerge.
Importantly, we do not claim that SAEs fail to capture information from trained Transformers above and beyond randomly initialized transformers; only that aggregate auto-interpretability measures do not necessarily indicate the existence of interesting underlying features.

% Could say something about:
% - the model we used to generate explanations (i.e. was it large enough?). but it's the default for the EleutherAI repo!
% - simulation scoring (almost certainly similar)
% - other transformer models (again almost certainly similar)
% - subsequent developments to the EleutherAI auto-interp repo - contrastive examples

\section{Conclusion}
\label{sec:conclusion}

In this work, we applied sparse autoencoders to both trained and randomly initialized transformers and evaluated them with a suite of common quantitative metrics.
Our central empirical finding is that, under certain conditions, these metrics -- particularly aggregate auto-interpretability scores -- can be surprisingly similar in both settings.
While we observe that features derived from trained transformers are qualitatively more complex and abstract, especially in later layers, these aggregate metrics often fail to capture this distinction.

This result does not imply that SAEs trained on real models fail to learn meaningful computational features.
Rather, it reveals a limitation in our current evaluation methods.
High aggregate auto-interpretability scores are insufficient proof for the discovery of complex, learned computations: they may instead reflect simpler structure inherent in the data or model architecture that is preserved even by random weights.
Our analysis of token distribution entropy, while preliminary, serves as a proof-of-concept: it successfully revealed differences in feature `abstractness' that aggregate auto-interpretability scores missed.
Future work should focus on developing more robust metrics that can quantify the computational significance of the features SAEs discover.
Our work reaffirms the importance of benchmarking interpretability techniques against strong, appropriately constructed null models, such as the randomly initialized transformers used here.
Without such baselines, it is difficult to confidently attribute discovered features to the process of learning.

\clearpage
\bibliographystyle{abbrvnat}
\bibliography{main}

\newpage
\appendix
\onecolumn

\section{Broader Impact}
\label{sec:impact}
This work investigates a method currently used for mechanistic interpretability of LLMs, yielding results that challenge certain assumptions about sparse autoencoders. By demonstrating that SAEs can produce similar aggregate auto-interpretability scores for both random and trained transformers, our findings raise important questions about what these SAE evaluation methods are actually capturing.

By better understanding the metrics of SAE quality, we hope that this work will contribute to a more informed search of better SAE-like methods and thus help to make these models more interpretable and to mitigate the potential harm these models could cause. Since our work is an empirical study of the capabilities of a presently used method, and it shows that the method provides interpretation of both random and trained transformers, we think the risk that this work could lead to negative social impact is minimal.

\section{Auto-interpretability ROC curves}
\label{app:roc_curves}

\begin{figure*}[!htb]
  \begin{center}
  \includegraphics[width=0.85\textwidth]{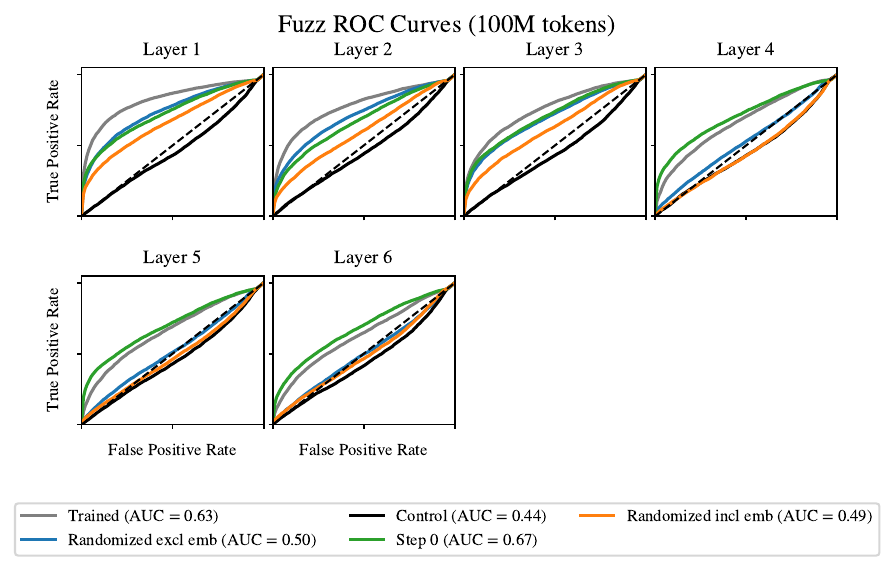}
  \end{center} 
  \caption{ROC curves for `fuzzing' auto-interpretability for Pythia-70m over 100 SAE latents. These results demonstrate the similarity in performance between the SAE variants, as well as the overall degradation in performance as the layer index increases.}
  \label{fig:pythia_70m_fuzzing}
\end{figure*}

\begin{figure*}[!htb]
  \begin{center}
  \includegraphics[width=0.85\textwidth]{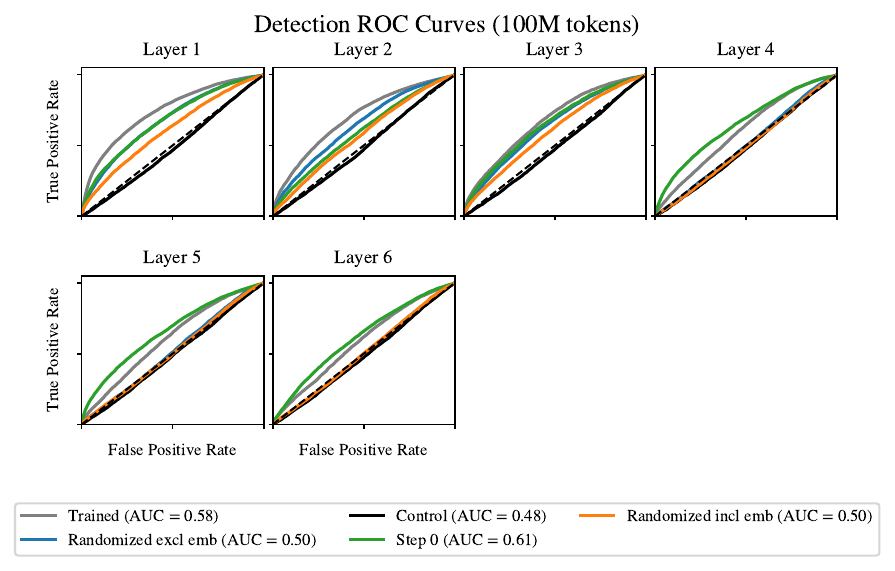}
  \end{center} 
  \caption{ROC curves for `detection' auto-interpretability for Pythia-70m over 100 SAE latents. These results demonstrate the similarity in performance between the SAE variants, as well as the overall degradation in performance as the layer index increases.}
  \label{fig:pythia_70m_detection}
\end{figure*}

\begin{figure*}[!htb]
  \begin{center}
  \includegraphics[width=0.85\textwidth]{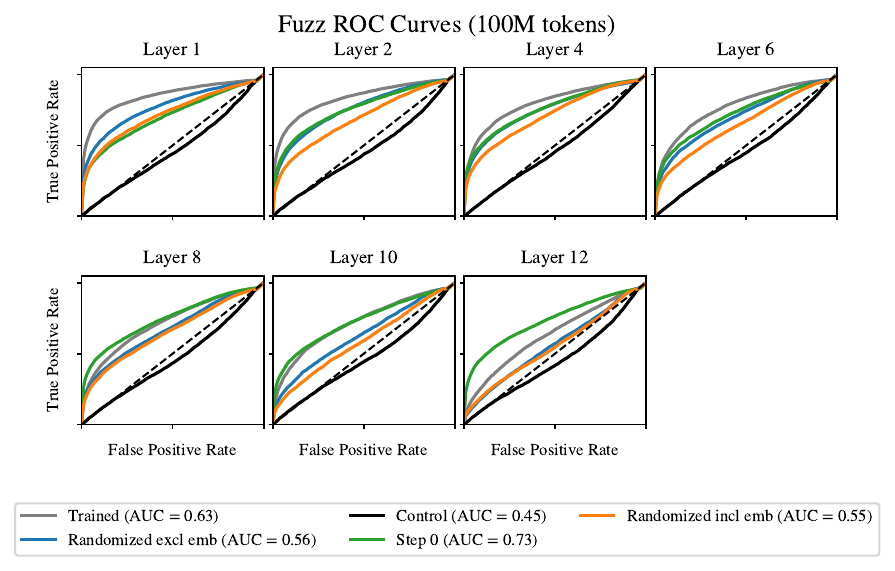}
  \end{center}
  \caption{ROC curves for `fuzzing' auto-interpretability for Pythia-160m over 100 SAE latents. These results demonstrate the similarity in performance between the SAE variants, as well as the overall degradation in performance as the layer index increases.}
\label{fig:pythia_160m_fuzzing}
\end{figure*}

\begin{figure*}[!htb]
  \begin{center}
  \includegraphics[width=0.85\textwidth]{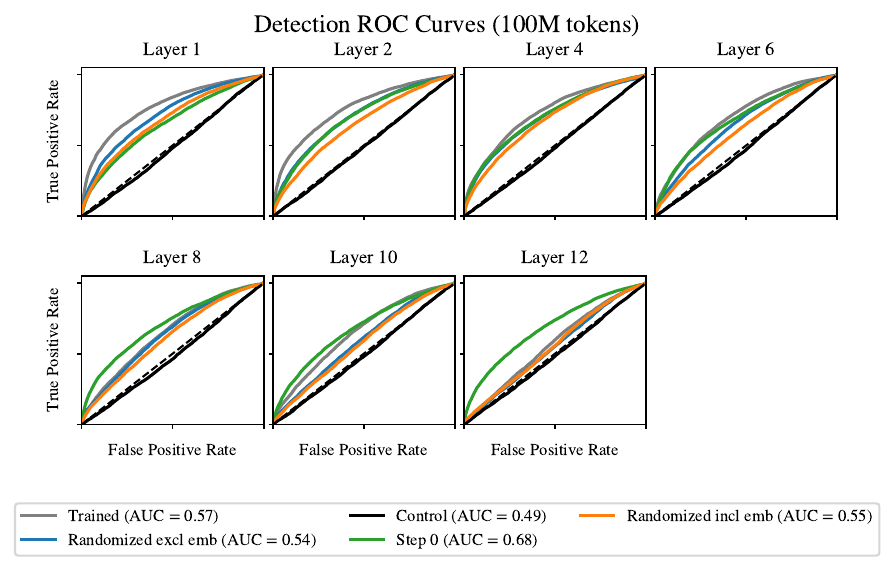}
  \end{center}
  \caption{ROC curves for `detection' auto-interpretability for Pythia-160m over 100 SAE latents. These results demonstrate the similarity in performance between the SAE variants, as well as the overall degradation in performance as the layer index increases.}
\label{fig:pythia_160m_detection}
\end{figure*}

\begin{figure*}[!htb]
  \begin{center}
  \includegraphics[width=0.85\textwidth]{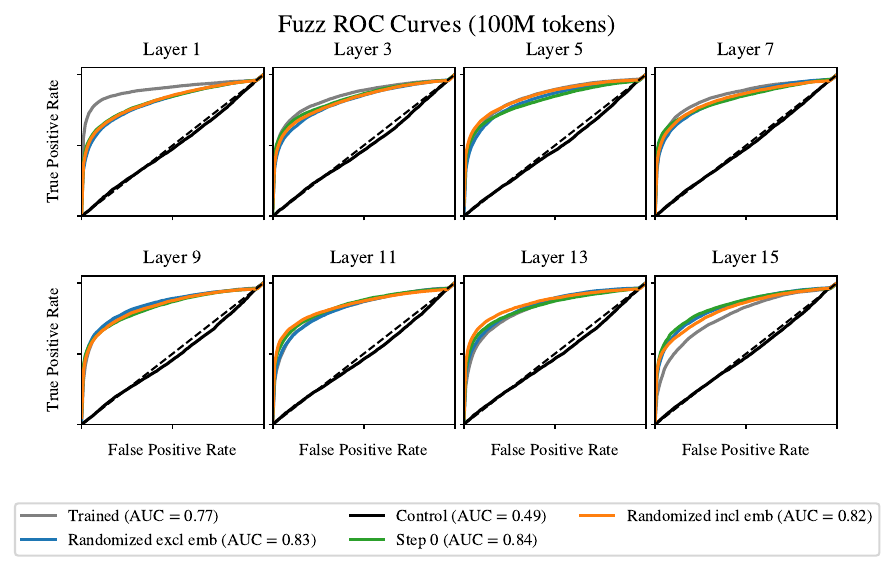}
  \end{center}
  \caption{ROC curves for `fuzzing' auto-interpretability for Pythia-1b over 100 SAE latents. These results demonstrate the similarity in performance between the SAE variants, although here we do not observe an overall degradation in quality.}
  \label{fig:pythia_1b_fuzzing}
\end{figure*}

\begin{figure*}[!htb]
  \begin{center}
  \includegraphics[width=0.85\textwidth]{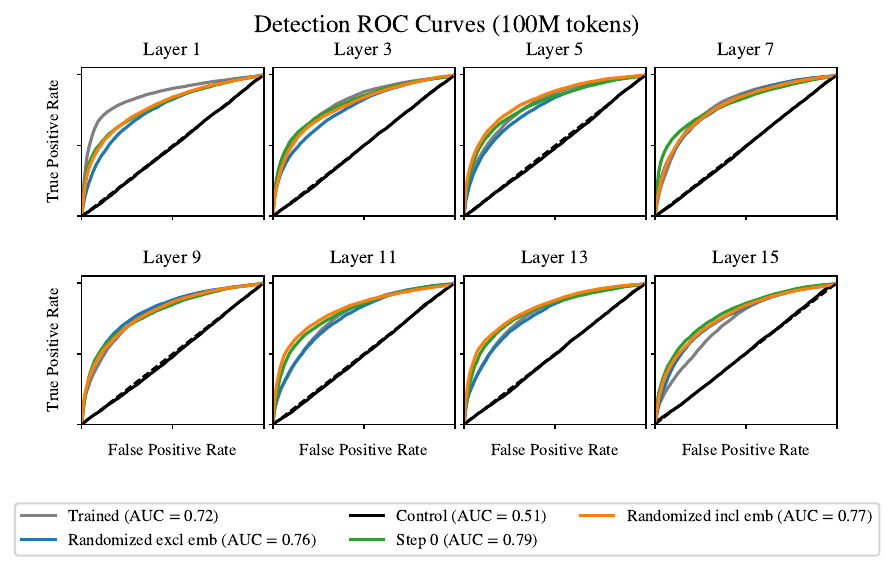}
  \end{center}
  \caption{ROC curves for `detection' auto-interpretability for Pythia-1b over 100 SAE latents. These results demonstrate the similarity in performance between the SAE variants, although here we do not observe an overall degradation in quality.}
  \label{fig:pythia_1b_detection}
\end{figure*}

Figures ~\ref{fig:pythia_70m_fuzzing}, ~\ref{fig:pythia_160m_fuzzing}, ~\ref{fig:pythia_1b_fuzzing} show the similarity between `fuzzing' AUROC for the trained and randomized SAEs for the 70M, 160M, and 1B models.
Figures ~\ref{fig:pythia_70m_detection}, ~\ref{fig:pythia_160m_detection}, ~\ref{fig:pythia_1b_detection}, show the similarity between `detection' AUROC for the trained and randomized SAEs for the 70M, 160M, and 1B models. 

\subsection{Pythia 70m}

\newpage

\subsection{Pythia 160m}

\newpage
\subsection{Pythia 410m}

\begin{figure*}[!htb]
  \begin{center}
  \includegraphics[width=0.8\textwidth]{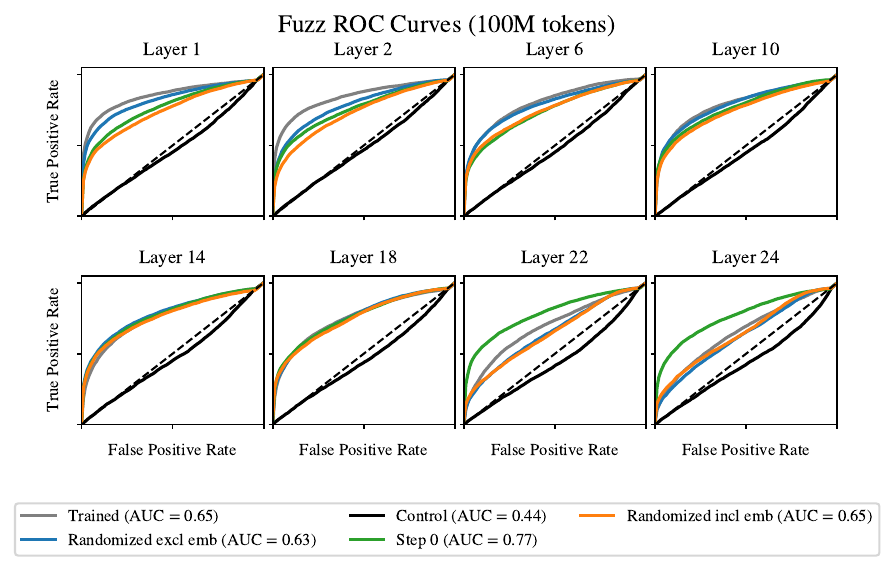}
  \end{center}
  \caption{ROC curves for `fuzzing' auto-interpretability for Pythia-410m over 100 SAE latents. These results demonstrate the similarity in performance between the SAE variants, as well as the overall degradation in performance as the layer index increases. The auto-interpretability scores here fail to distinguish between trained and randomized models.}
  \label{fig:pythia_410m_fuzzing}
\end{figure*}

\begin{figure*}[!htb]
  \begin{center}
  \includegraphics[width=0.85\textwidth]{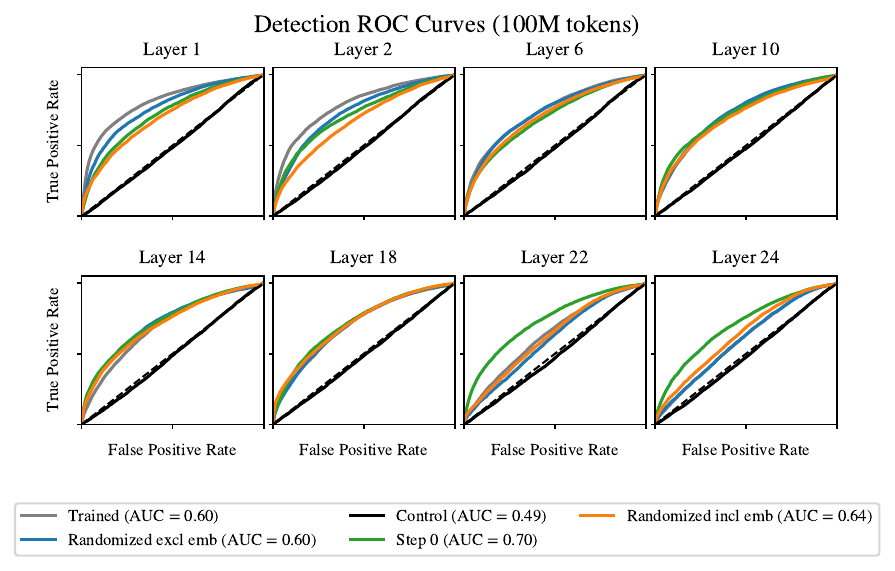}
  \end{center}
  \caption{ROC curves for `detection' auto-interpretability for Pythia-410m over 100 SAE latents. These results demonstrate the similarity in performance between the SAE variants, as well as the overall degradation in performance as the layer index increases.}
\label{fig:pythia_410m_detection}
\end{figure*}

\newpage
\subsection{Pythia-1b}

\newpage
\subsection{Pythia 6.9b}

\begin{figure*}[!htb]
  \begin{center}
  \includegraphics[width=0.85\textwidth]{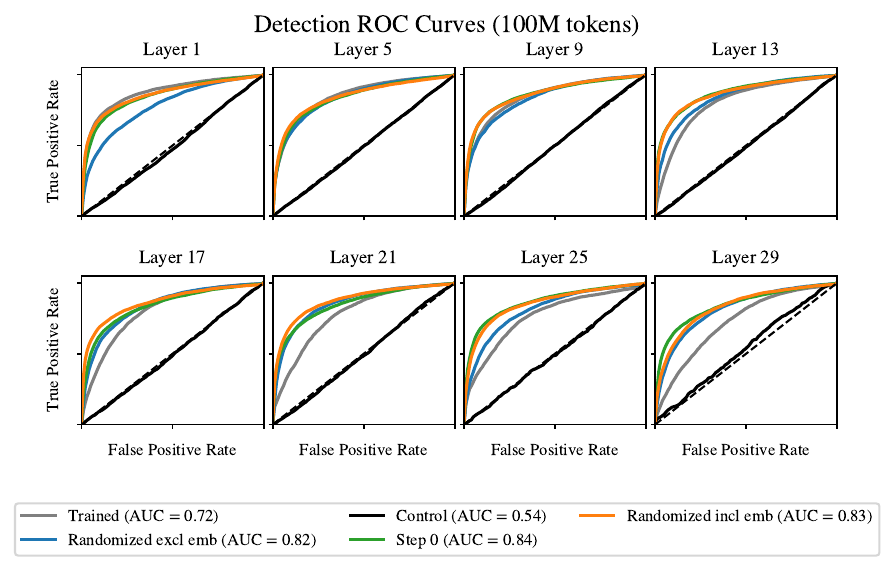}
  \end{center}
  \caption{ROC curves for `detection' auto-interpretability for Pythia-6.9b over 100 SAE latents. These results demonstrate the similarity in performance between the SAE variants.}
  \label{fig:pythia_6.9b_detection}
\end{figure*}

\clearpage

\section{Effect of increased training data}
\label{app:1b_tokens}

\begin{figure*}[!htb]
  \begin{center}
  \includegraphics[width=0.85\textwidth]{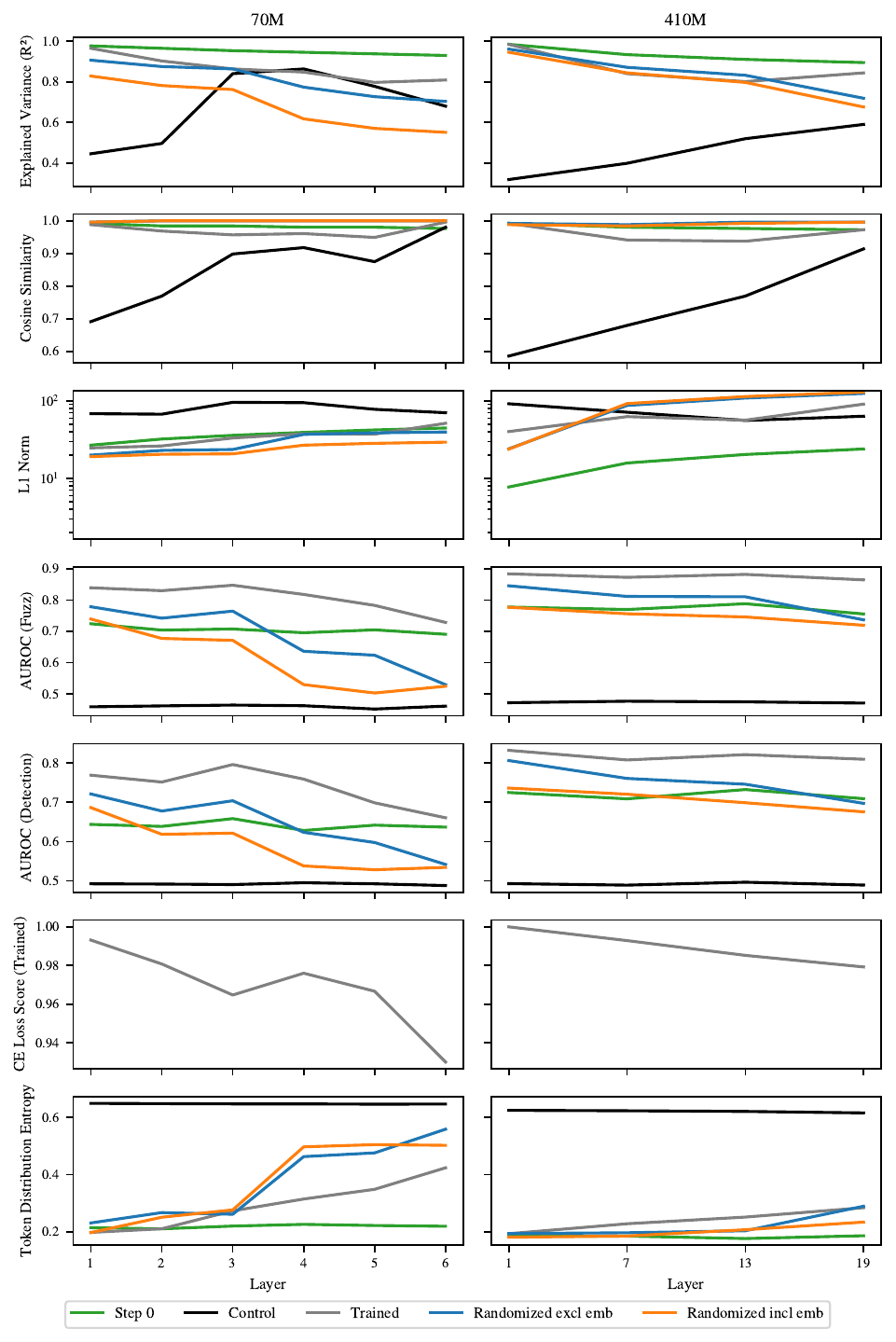}
  \end{center}
  \caption{Evaluation metrics for SAEs trained with one billion tokens on the Pythia-70m and 410m models. These results correspond to columns of Figure~\ref{fig:all_metrics}, which show the same evaluation metrics for SAEs trained on 100M tokens, and qualitatively similar behavior.}
  \label{fig:1B_metrics}
\end{figure*}

For our primary experiments, we trained SAEs on 100M tokens (Section~\ref{sec:results}).
We verified that our results were not explained by a lack of sufficient training data by repeating a subset of these experiments with SAEs trained on 1B tokens from the RedPajama dataset (Figure~\ref{fig:1B_metrics}).

\clearpage
\section{Effect of decreased training data for Pythia-1B}
\label{app:1m_tokens}

\begin{figure*}[!htb]
  \begin{center}
  \includegraphics[width=0.75\textwidth]{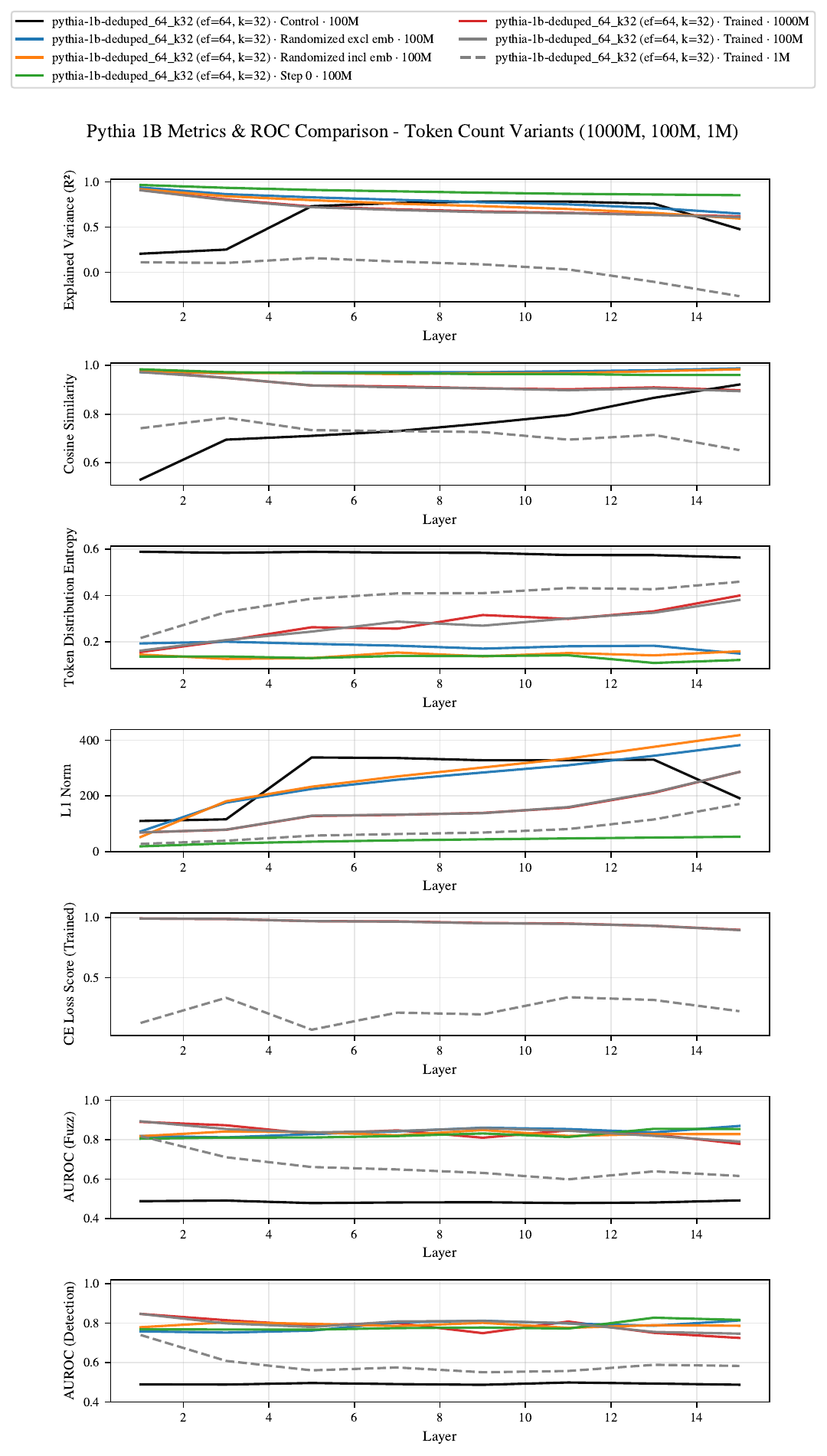}
  \end{center}
  \caption{Evaluation metrics for SAEs trained with 1M and 1B tokens on Pythia-1b.  The explained variance and CE loss score are significantly lower for the 1M model, showing that the SAEs are under-trained. Average auto-interpretability scores are slightly lower for the earliest layers, but decline sharply with increasing layer. The trends in auto-interpretability and token distribution entropy with layer index are consistent with other SAEs.}
  \label{fig:1M_metrics}
\end{figure*}

\clearpage
\section{Uncertainty plots for Pythia-70m}
\label{app:uncertainty}

We computed uncertainty for our evaluation metrics on Pythia-70m using five random seeds.

\begin{figure*}[h]
  \begin{center}
  \includegraphics[width=0.725\textwidth]{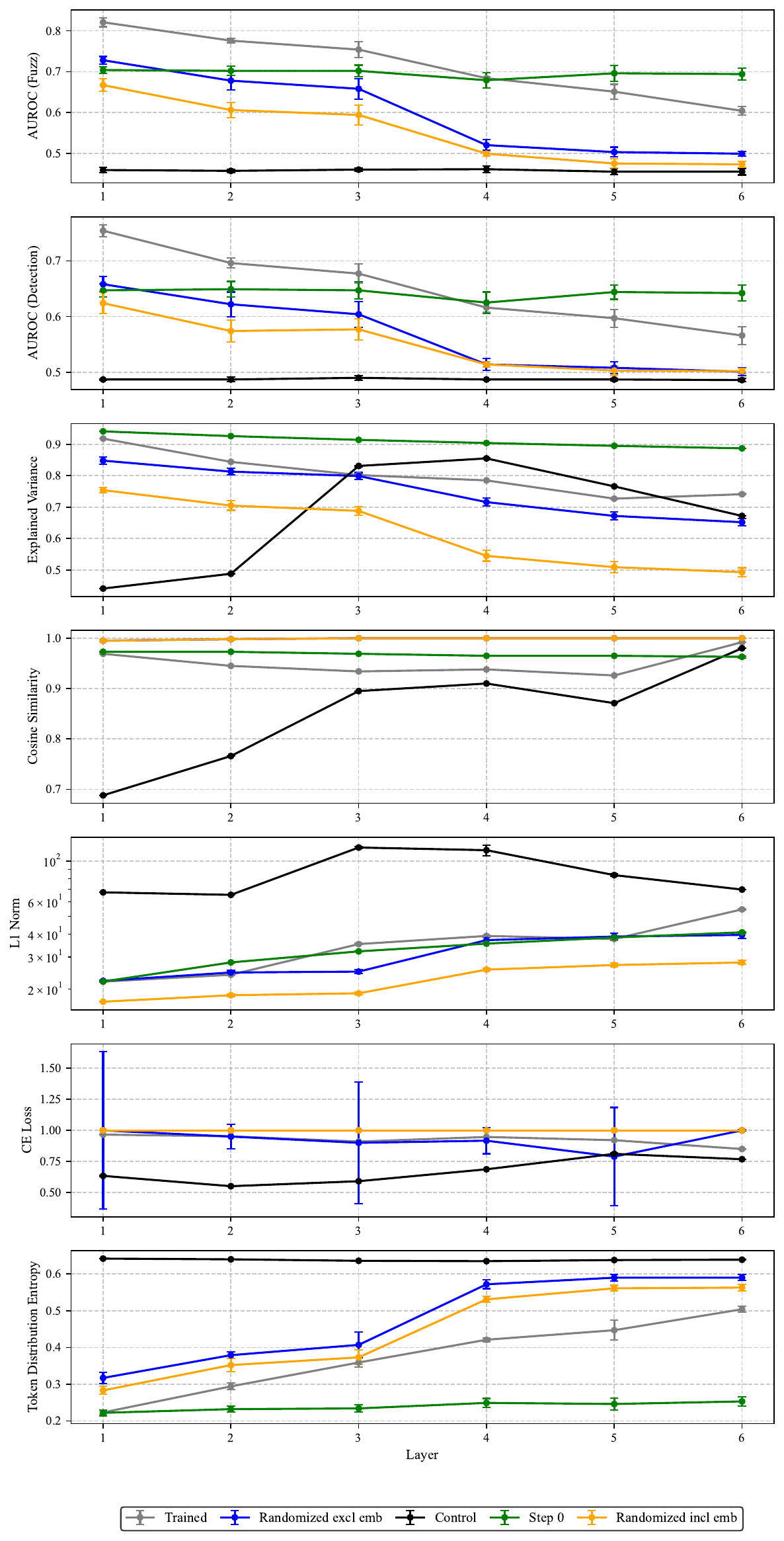}
  \end{center}
  \caption{Uncertainty for Pythia-70m metrics computed using five random seeds.}
  \label{fig:uncertainty}
\end{figure*}

\clearpage
\section{Effect of SAE hyperparameters for Pythia-160m}
\label{app:hyper_params}

% Here we demonstrate the stability of our results over a hyperparameter sweep of the expansion factor $R$ and sparsity $k$ for SAEs trained on Pythia-160m.

\begin{figure*}[!htb]
  \begin{center}
  \includegraphics[width=\textwidth]{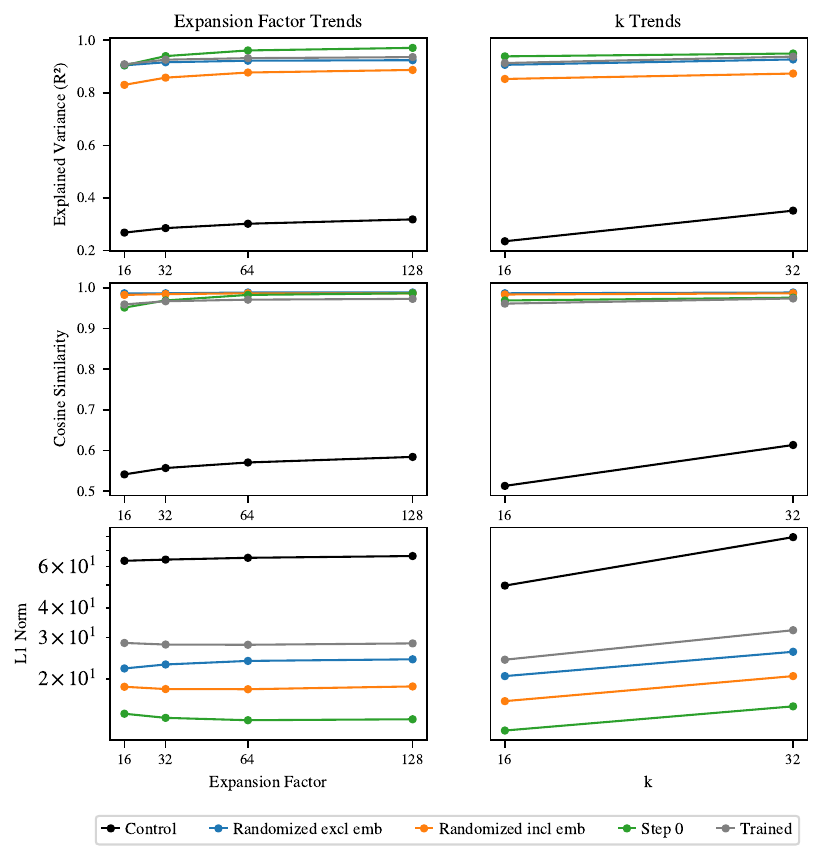}
  \end{center}
  \caption{Robustness of SAE performance to hyperparameter selection. Standard evaluation metrics remain stable across a wide range of expansion factors $R$ (16 to 128) and sparsities $k$ (16 to 32), with all initialization strategies maintaining their relative performance ordering. This stability suggests that moderate hyperparameter values (e.g., expansion factor $R=64$, sparsity $k=32$) suffice.}
  \label{fig:hyper_params}
\end{figure*}

\clearpage

\section{Effect of SAE hyperparameters for Pythia-1b}
\label{app:bad_hyperparams}

\begin{figure*}[!htb]
  \begin{center}
  \includegraphics[width=0.8\textwidth]{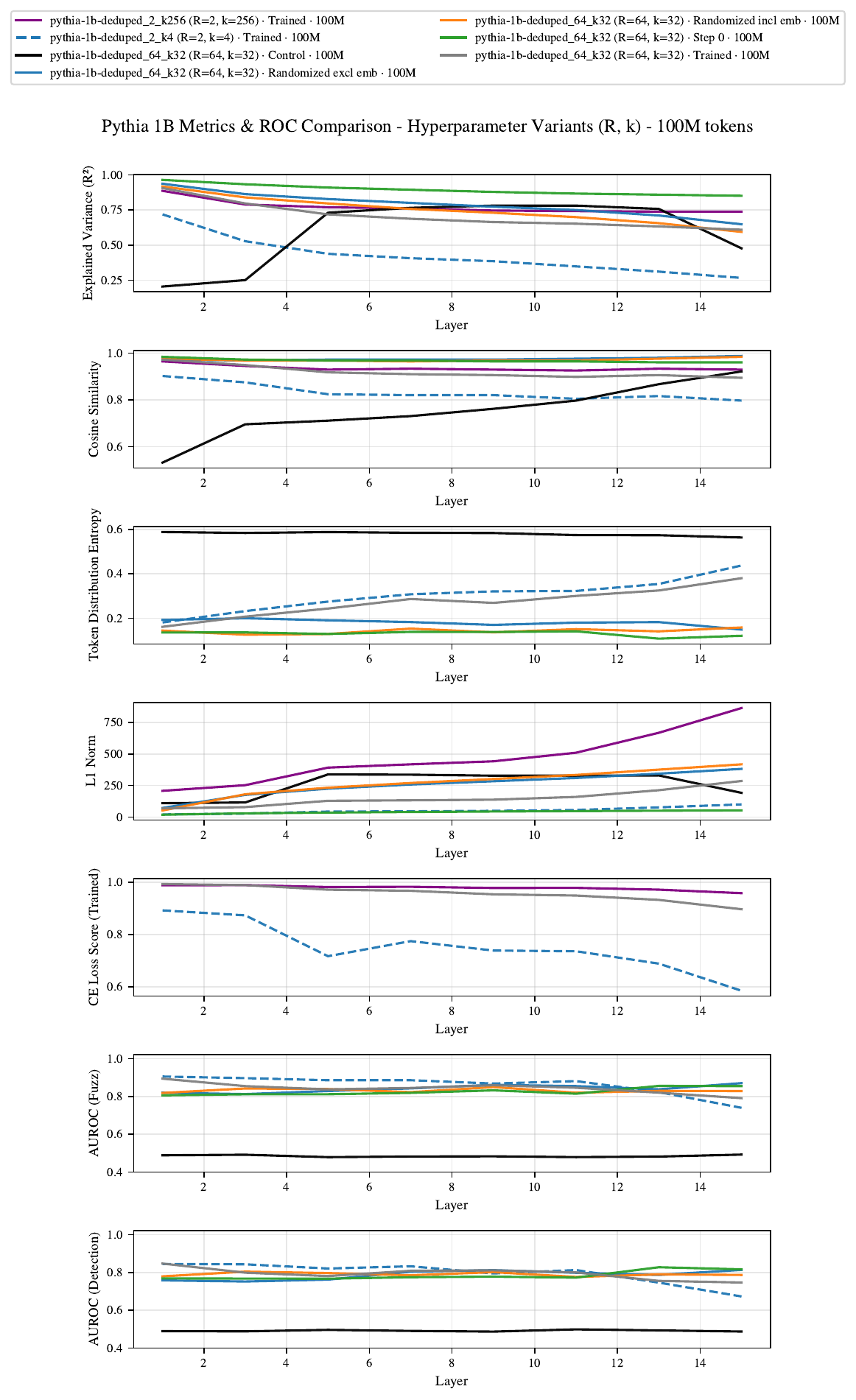}
  \end{center}
  \caption{Evaluation metrics for SAEs trained on the Pythia-1b model with different hyperparameters, including the main results from Figure~\ref{fig:all_metrics}. SAEs with a very small expansion factor $R=2$ and sparsity $k=4$ are clearly distinguished from our default hyperparameters by the explained variance and CE loss score. Importantly, the auto-interpretability scores of these SAEs remain similar to those trained with default hyperparameters on either trained or randomised models.}
  \label{fig:bad_hyperparams}
\end{figure*}

\clearpage
\section{Token distribution entropy vs. auto-interpretability}
\label{app:ent_fuzz_auroc}

\begin{figure*}[!htb]
    \centering
    \includegraphics{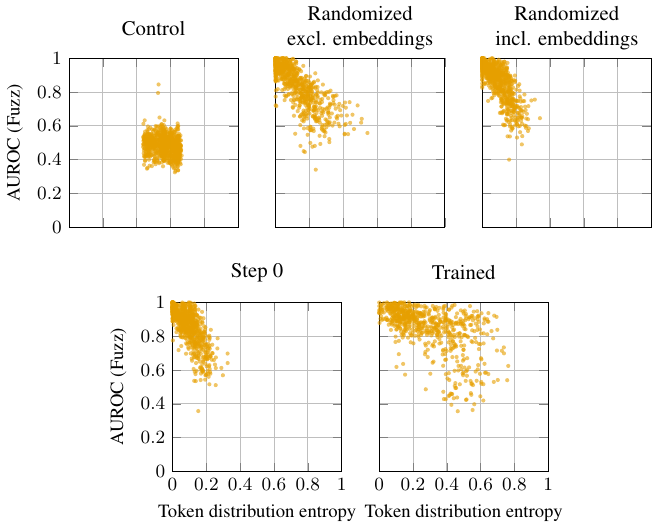}
  \caption{Scatter plots of the per-latent token distribution entropy against `fuzzing' AUROC (auto-interpretability score) for SAEs trained on multiple layers of the Pythia-6.9b model. Each point corresponds to a single latent, taken from the sample of latents used to compute the aggregate metrics displayed in Figures~\ref{fig:pythia_6.9b_fuzzing} and \ref{fig:all_metrics}.
  }
  \label{fig:ent_fuzz_auroc}
\end{figure*}

Figure~\ref{fig:ent_fuzz_auroc} clearly distinguishes the negative control, randomized variants, and the trained variant:
\begin{itemize}
    \item \textbf{Control:} Latents have a consistently high entropy (i.e., max activating examples with activation patterns spread across many tokens) and low auto-interpretability score (i.e., generated explanations that fail to adequately explain these activation patterns). No correlation between the two variables is evident.
    \item \textbf{Randomized:} For each of the randomization schemes described in Section~\ref{sec:results}, we see a negative correlation between entropy and auto-interpretability: in general, the wider variety of tokens for which a latent is activated, the less well the latent's activation patterns are explained by its generated explanation.
    \item \textbf{Trained:} There is a weaker correlation between the two variables. Crucially, in addition to the broad trend observed for the randomized variants, we also see latents with high entropy \emph{and} auto-interpretability. Some latents have activation patterns that are spread across multiple tokens, which are nevertheless consistent with the latent's generated explanation.
\end{itemize}
These results are consistent with the view that aggregate auto-interpretability scores obscure the differences between SAEs based on trained and randomized models. While randomized models with consistent token embeddings can produce `single-token' features, whose activation patterns are easy to explain, only Transformers trained on natural language produce more complex semantic features.

\clearpage
\section{A toy model of superposition}
\label{app:toy_models}

\definecolor{codegreen}{rgb}{0,0.6,0}
\definecolor{codegray}{rgb}{0.5,0.5,0.5}
\definecolor{codepurple}{rgb}{0.58,0,0.82}
\definecolor{backcolour}{rgb}{0.95,0.95,0.92}
\lstdefinestyle{mystyle}{
    backgroundcolor=\color{backcolour},   
    commentstyle=\color{codegreen},
    keywordstyle=\color{magenta},
    numberstyle=\tiny\color{codegray},
    stringstyle=\color{codepurple},
    basicstyle=\ttfamily\footnotesize,
    lineskip=-1pt,
    breakatwhitespace=false,         
    breaklines=true,                 
    captionpos=b,
    keepspaces=true,                 
    numbers=left,                    
    numbersep=5pt,                  
    showspaces=false,                
    showstringspaces=false,
    showtabs=false,                  
    tabsize=2
}
\lstset{style=mystyle}

In Section~\ref{sec:toy_models}, we trained SAEs on toy data designed to exhibit superposition \citep{sharkey_taking_2022} and GloVe word vectors \citep{pennington_glove_2014}.
In this section, we detail the data-generation procedure and training setup.

\newcommand{\R}{\mathbb{R}}
\renewcommand{\N}{\mathcal{N}}
\newcommand{\p}{\alpha}

\subsection{Data generation}
\label{app:toy_models_data}

First, we construct ground-truth features by sampling $\nsp$ points on an $\nde$-dimensional hypersphere.

For each sample, we determine the feature coefficients by generating $A \in \R^{\nsp \times \nsp}$ where $A_{ij} \sim \N(0,1)$, defining a covariance matrix $\Sigma = AA^\mathsf{T}$, sampling $\vec{\p} \in \R^\nsp$ where $\p_i \sim \N(\vec{0}, \Sigma)$, projecting $\p_i$ onto the c.d.f. of $\N(0,1)$, decaying $\p_i \to \p_i^{\lambda i}$ where $\lambda \in \R$, normalizing $\p_i \to m \p_i \big/ \nsp \sum_j \p_j$ where $m \in \R$, and performing $\nsp$ independent Bernoulli trials with $p = \p_i$.
Finally, we multiply the trial outcomes by $\nsp$ independent samples from a continuous uniform distribution $\mathcal{U}_{[0,1)}$.

The parameter $\lambda$ determines how sharply the frequency of nonzero ground-truth feature coefficients decays with the feature index $i$.
The parameter $m$ is the expected value of the number of nonzero feature coefficients for each sample.

Like \citet{sharkey_taking_2022}, we choose $\nsp = 512$, $\nde = 256$, $\lambda = 0.99$, and $m = 5$.
We include a Python implementation of this procedure in Figure~\ref{fig:sharkey_python}.

\subsection{Training}
\label{app:toy_models_training}

The SAEs described in Section~\ref{sec:toy_models} comprise a linear encoder with a bias term, a ReLU activation function, and a linear decoder without a bias term.
We use orthogonal initialization for the decoder weights and normalize the decoder weight vectors before each training step.

The training loss is the mean squared error (MSE) between the input and decoded vectors, plus the mean $L^1$ norm of the encoded vectors multiplied by a coefficient, which we vary between \qty{1e-3} and 100.

For the toy data, we train for 100 epochs on 10K data points with 10 random seeds.
For the word vectors, we train for 100 epochs on 400K data points with 1 random seed.
In both experiments, we reserve 10\% of the data points as a validation set, which we use to compute evaluation metrics.

The MLPs described in Section~\ref{sec:toy_models} comprise two layers (i.e., one hidden layer) and a ReLU activation function.
The input and output sizes are both equal to $\nde$, and the hidden size is $4\nde$.
We loosely based these choices on the feed-forward network components of transformer language models.

\begin{figure}
\begin{lstlisting}[language=Python]
def generate_sharkey(
    num_samples: int,
    num_inputs: int,
    num_features: int,
    avg_active_features: float,
    lambda_decay: float,
) -> tuple[Tensor, Tensor]:
    """
    Args:
        num_samples (int): The number of samples to generate.
        num_inputs (int): The number of input dimensions.
        num_features (int): The number of ground truth features.
        avg_active_features (float): The average number of
            ground truth features active at a time.
        lambda_decay (float): The exponential decay factor for
            feature probabilities.
    """
    features = torch.randn(num_inputs, num_features)
    features /= torch.norm(features, dim=0, keepdim=True)

    covariance = torch.randn(num_features, num_features)
    covariance = covariance @ covariance.T
    correlated_normal = MultivariateNormal(
        torch.zeros(num_features), covariance_matrix=covariance
    )

    samples = []
    for _ in range(num_samples):
        p = STANDARD_NORMAL.cdf(correlated_normal.sample())
        p = p ** (lambda_decay * torch.arange(num_features))
        p = p * (avg_active_features / (num_features * p.mean()))
        p = torch.bernoulli(p.clamp(0, 1))
        coef = p * torch.rand(num_features)

        sample = coef @ features.T
        samples.append(sample)

    return torch.stack(samples), features
\end{lstlisting}
\caption{A Python implementation of the data-generation procedure introduced by \citet{sharkey_taking_2022} and used in Section~\ref{sec:toy_models}.}
\label{fig:sharkey_python}
\end{figure}

\newcommand{\paretowidth}{\textwidth}

\begin{figure}
    \centering
    \includegraphics{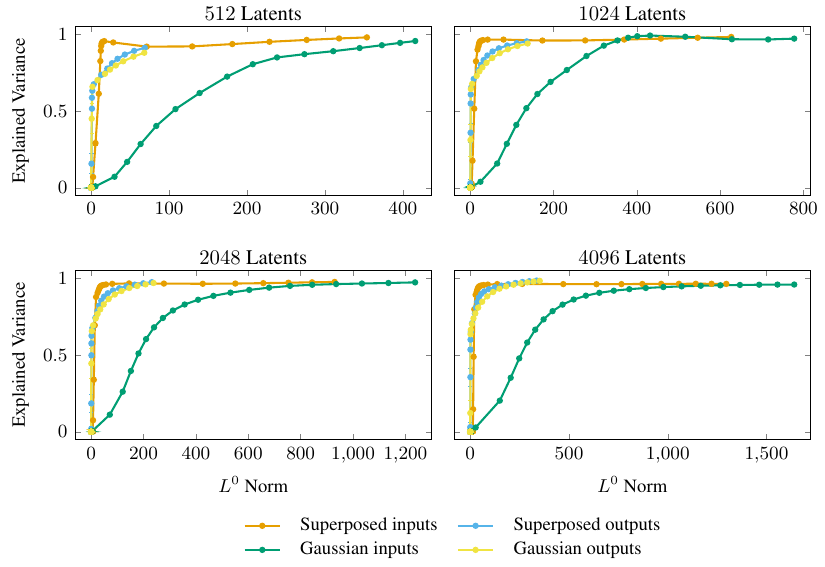}
    \caption{Pareto frontiers of the explained variance against the $L^0$ norm (sparsity) for toy datasets generated to exhibit superposition, Gaussian controls with the same mean and variance, and the corresponding outputs when these are passed to a randomly initialized two-layer MLP.}
\end{figure}

\begin{figure}
    \centering
    \includegraphics{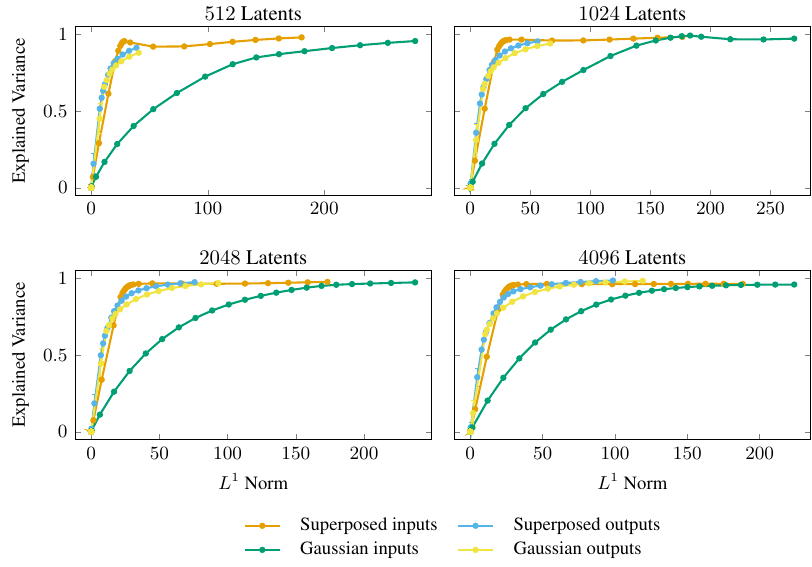}
    \caption{Pareto frontiers of the explained variance against the $L^1$ norm (sparsity) for toy datasets generated to exhibit superposition, Gaussian controls with the same mean and variance, and the corresponding outputs when these are passed to a randomly initialized two-layer MLP.}
\end{figure}

\begin{figure}
    \centering
    \includegraphics{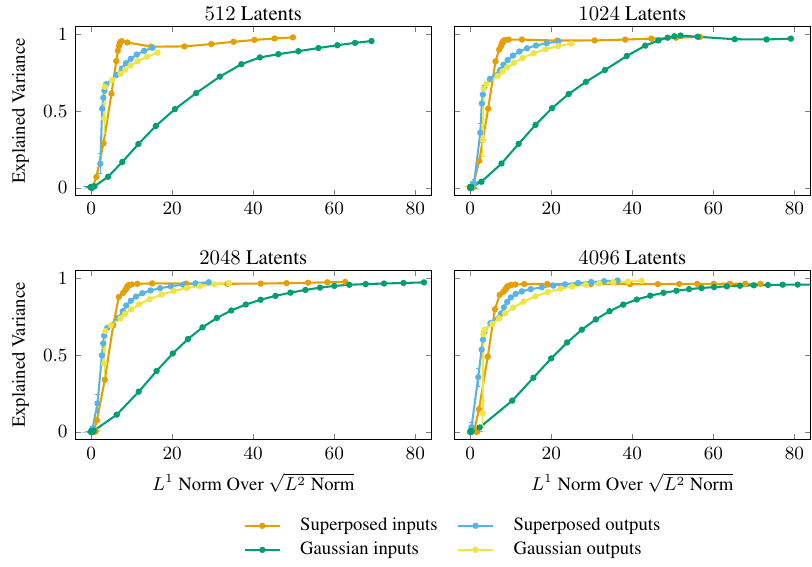}
    \caption{Pareto frontiers of explained variance against the $L^1$ norm over the square root of the $L^2$ norm (sparsity) for toy datasets generated to exhibit superposition, Gaussian controls with the same mean and variance, and the corresponding outputs when these are passed to a randomly initialized two-layer MLP.}
\end{figure}

\begin{figure}
    \centering
    \includegraphics{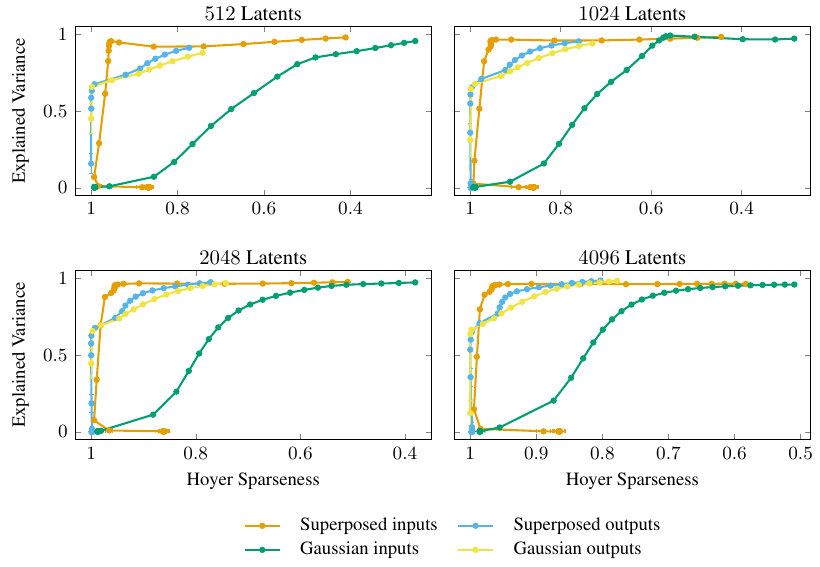}
    \caption{Pareto frontiers of explained variance against the Hoyer sparseness (sparsity) for toy datasets generated to exhibit superposition, Gaussian controls with the same mean and variance, and the corresponding outputs when these are passed to a randomly initialized two-layer MLP.}
\end{figure}

\begin{figure}
    \centering
    \includegraphics{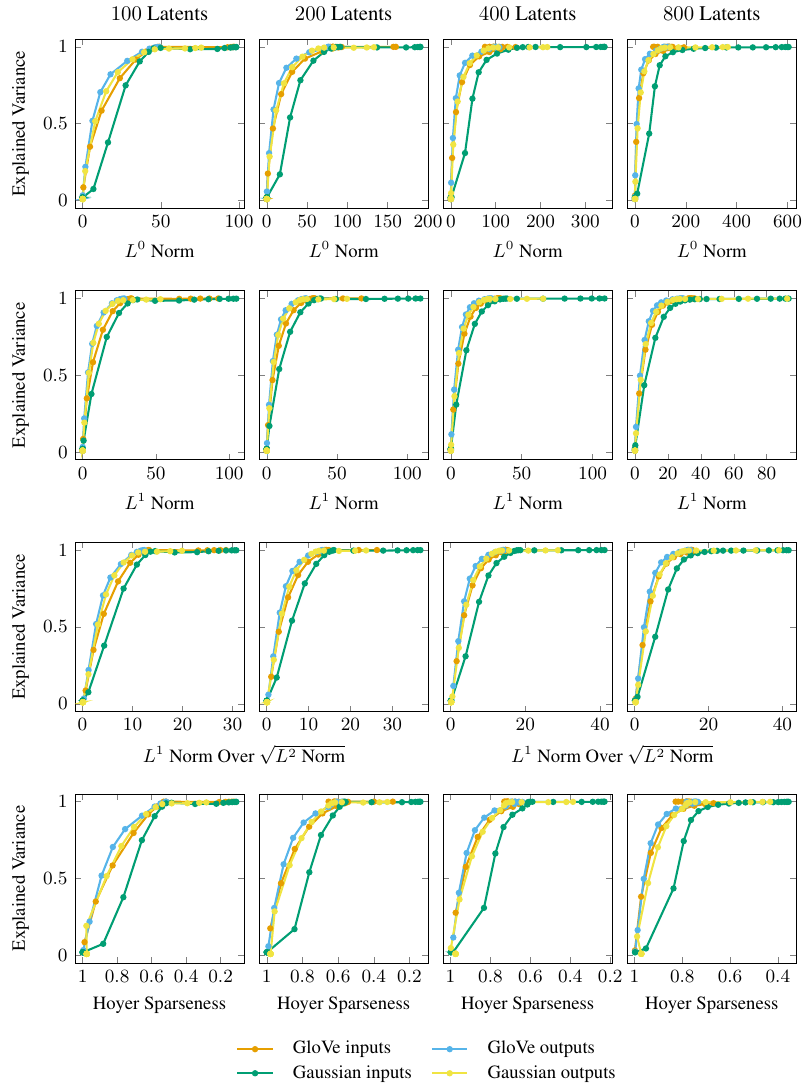}
    \caption{Pareto frontiers of explained variance against sparsity measures for 50-dimensional GloVe word vectors, Gaussian controls with the same mean and variance, and the corresponding outputs when these are passed to a randomly initialized two-layer MLP.}
\end{figure}

\begin{figure}
    \centering
    \includegraphics{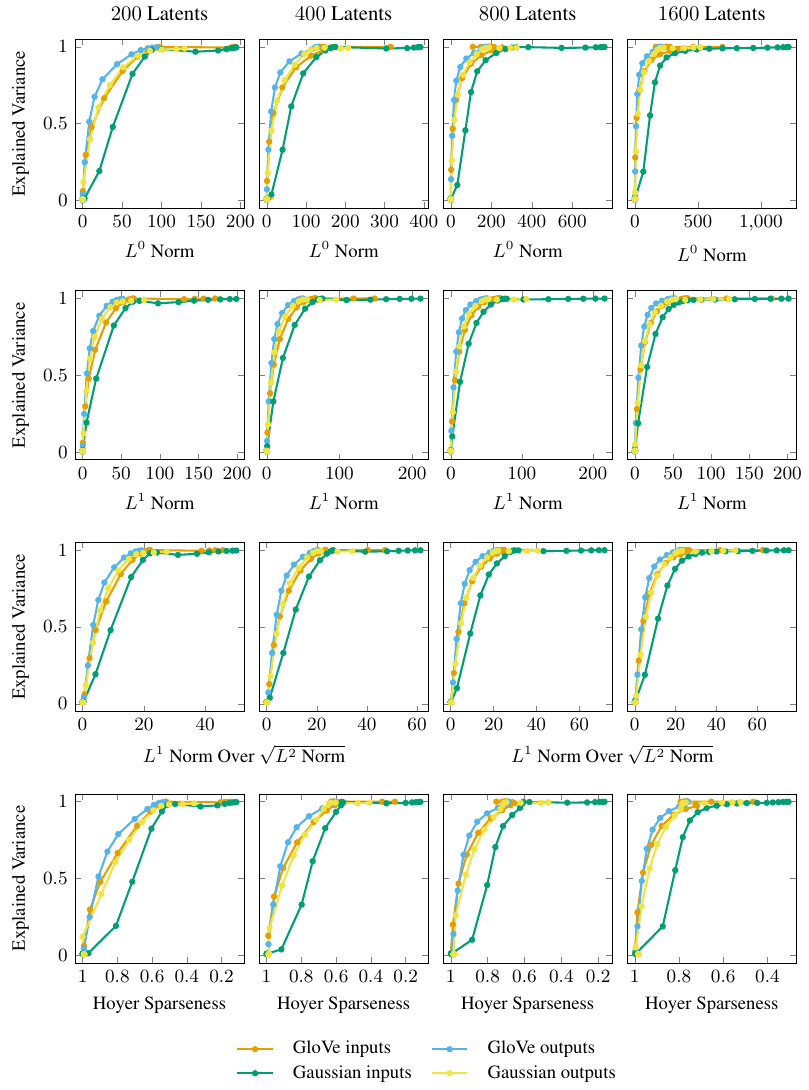}
    \caption{Pareto frontiers of explained variance against sparsity measures for 100-dimensional GloVe word embeddings, Gaussian controls with the same mean and variance, and the corresponding outputs when these are passed to a randomly initialized two-layer MLP.}
\end{figure}

\begin{figure}
    \centering
    \includegraphics{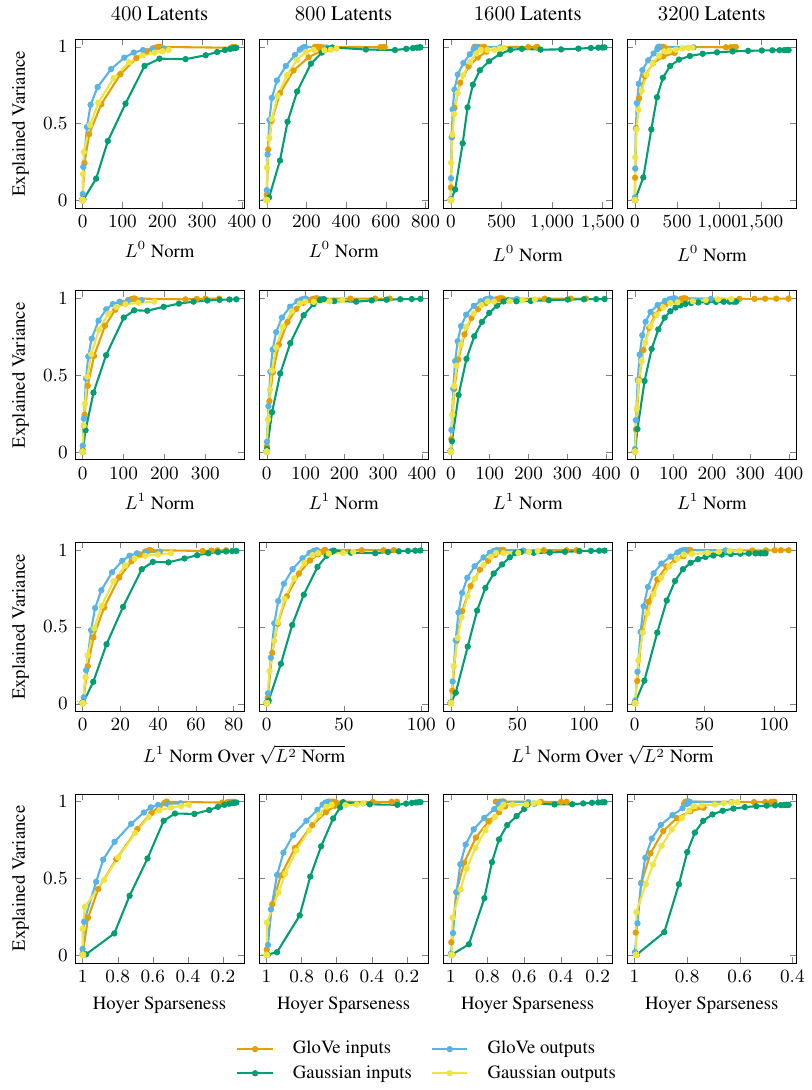}
    \caption{Pareto frontiers of explained variance against sparsity measures for 200-dimensional GloVe word embeddings, Gaussian controls with the same mean and variance, and the corresponding outputs when these are passed to a randomly initialized two-layer MLP.}
\end{figure}

\begin{figure}
    \centering
    \includegraphics{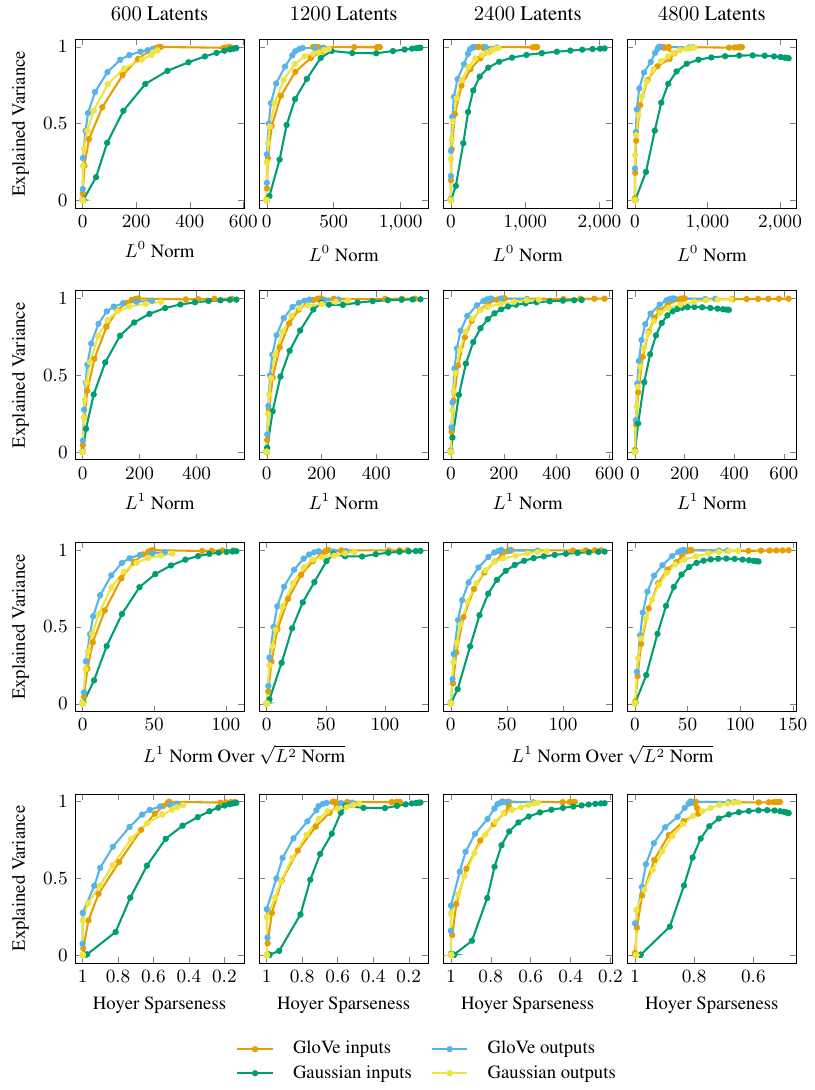}
    \caption{Pareto frontiers of explained variance against sparsity measures for 300-dimensional GloVe word embeddings, Gaussian controls with the same mean and variance, and the corresponding outputs when these are passed to a randomly initialized two-layer MLP.}
\end{figure}

\clearpage
\section{Example features for Pythia-6.9b}
\label{app:6.9b_example_features}

\subsection*{Variant: Trained}

\subsubsection*{Feature 180935 (Layer 0)}
\textbf{Interpretation:}
The term "security" is predominantly used to refer to protection, safety, and measures to prevent harm, while "oz" is likely referring to ounces, possibly in a context of measurement or quantification, although "oz" appears less frequently and often in a different context.

\textbf{Top Examples:}
\begin{enumerate}
\item
\begin{tabular}{p{0.95\textwidth}}
Text: \verb|<|endoftext|>|. If for any reason you are unhappy with our service please contact us directly so we can make it right for you.Journal of Cyber Security, Vol. \\
Activation: 4.3750 \\
Active tokens: Security
\end{tabular}
\item
\begin{tabular}{p{0.95\textwidth}}
Text: \verb|<|endoftext|>| Security Practitioner.
Tremendously passing CompTIA Advanced Security Practitioner (casp) cert has never been as easy as it \\
Activation: 4.3125 \\
Active tokens: Security   Security
\end{tabular}
\item
\begin{tabular}{p{0.95\textwidth}}
Text: in trusted hands for your Cyber Security career or staffing needs.
Call 0203 643 0248 to find out more.
Technically proficient using \\
Activation: 4.3125 \\
Active tokens: Security
\end{tabular}
\end{enumerate}
\hrulefill

\subsubsection*{Feature 93790 (Layer 8)}
\textbf{Interpretation:}
Nouns and phrases related to economic concepts, development, and business, often referring to growth, progress, and improvement.

\textbf{Top Examples:}
\begin{enumerate}
\item
\begin{tabular}{p{0.95\textwidth}}
Text: training requirements.
See “Workforce” section for additional information.
The Economic Development Transportation Fund, commonly referred to as the “Road Fund,” is an \\
Activation: 21.3750 \\
Active tokens: Development
\end{tabular}
\item
\begin{tabular}{p{0.95\textwidth}}
Text: Montréal.The Williamsburg Economic Development Authority offers a 33\% matching grant up to \$7,500 for exterior improvements to existing businesses in the City of \\
Activation: 20.2500 \\
Active tokens: Development   Authority
\end{tabular}
\item
\begin{tabular}{p{0.95\textwidth}}
Text: Correction: In a July 16 web story The Real Deal incorrectly stated that the Economic Development Corporation was “circumventing” laws with its restructuring. In \\
Activation: 20.0000 \\
Active tokens: Development
\end{tabular}
\end{enumerate}
\hrulefill

\subsubsection*{Feature 128309 (Layer 12)}
\textbf{Interpretation:}
Various types of punctuation and grammatical elements that separate words or phrases, including hyphens, commas, ellipses, prepositions, and determiners, often indicating connections, contrasts, or clarifications, and sometimes marking boundaries or transitions between clauses or ideas.

\textbf{Top Examples:}
\begin{enumerate}
\item
\begin{tabular}{p{0.95\textwidth}}
Text: Run it in JDK6, and it will print "[axons, bandrils, chumblies]".
If you are having trouble switching from \\
Activation: 8.0000 \\
Active tokens: in   JD  K
\end{tabular}
\item
\begin{tabular}{p{0.95\textwidth}}
Text: Here, we introduce the coordinate systems for three-dimensional space \unknownchar{}\unknownchar{}\unknownchar{}2. The study of 3-dimensional spaces lead us to the setting for our study \\
Activation: 7.8125 \\
Active tokens: \unknownchar{}
\end{tabular}
\item
\begin{tabular}{p{0.95\textwidth}}
Text: .path.expanduser("\textasciitilde{}/malwarehouse/") because this server doesn't have X-Windows running.
If you are looking for a simple and \\
Activation: 7.7188 \\
Active tokens: .  path  expand  user
\end{tabular}
\end{enumerate}
\hrulefill

\subsection*{Variant: Step 0}
\subsubsection*{Feature 126848 (Layer 12)}
\textbf{Interpretation:}
Nouns denoting people who train others, units or marks of measurement, and abbreviations or acronyms representing specific standards or technologies.

\textbf{Top Examples:}
\begin{enumerate}
\item
\begin{tabular}{p{0.95\textwidth}}
Text: What are the various lessons a member can access at a tennis club?
Whether you are a beginner or advanced player, trainers help you to choose the right gaming \\
Activation: 13.1250 \\
Active tokens: trainers
\end{tabular}
\item
\begin{tabular}{p{0.95\textwidth}}
Text: report include various simulation platforms and Serious Games. The report also analyzes some major allied products such as patient simulators and task trainers.
The technologies analyzed \\
Activation: 13.0625 \\
Active tokens: trainers
\end{tabular}
\item
\begin{tabular}{p{0.95\textwidth}}
Text: a stylish spring in your step when you buy from our fantastic range of men’s and women’s Asics trainers. We’ve got numerous styles from \\
Activation: 13.0000 \\
Active tokens: trainers
\end{tabular}
\end{enumerate}
\hrulefill

\subsubsection*{Feature 2125 (Layer 4)}
\textbf{Interpretation:}
The word "papers" is often used in contexts referring to written documents, such as academic papers, court documents, or printed materials, and is frequently mentioned in relation to tasks like writing, research, and education.

\textbf{Top Examples:}
\begin{enumerate}
\item
\begin{tabular}{p{0.95\textwidth}}
Text: caustic solution . As an abrasive, alumina is coated into abrasive papers and .. Pakistan. Sierra leone. Taiwan. Turkey. Venezuela. \\
Activation: 5.7188 \\
Active tokens: papers
\end{tabular}
\item
\begin{tabular}{p{0.95\textwidth}}
Text: that can be associated with interaction with other individuals. For everybody who is uncertain regardless of whether your papers is misstep no cost, buy inexpensive experienced proofreading services \\
Activation: 5.6875 \\
Active tokens: papers
\end{tabular}
\item
\begin{tabular}{p{0.95\textwidth}}
Text: who RV, often traveling in groups, often alone. You just want to have all the papers like RC, licence and insurance coverage as effectively as PUC ( \\
Activation: 5.6562 \\
Active tokens: papers
\end{tabular}
\end{enumerate}
\hrulefill

\subsubsection*{Feature 9944 (Layer 16)}
\textbf{Interpretation:}
Words or parts of words that are usually the beginning or end of a proper noun, surname, or a word of foreign origin.

\textbf{Top Examples:}
\begin{enumerate}
\item
\begin{tabular}{p{0.95\textwidth}}
Text: astic gestures.Home Music World News Music the artform Where Do Music Festivals Go Now?
Where Do Music Festivals Go Now?
Are you ready \\
Activation: 10.4375 \\
Active tokens: Fest   Fest
\end{tabular}
\item
\begin{tabular}{p{0.95\textwidth}}
Text: client streams, and encouraging existing customers to become more involved.Welcome to Fil Fest USA!!
Do you love lumpia? Can you eat a handful of them \\
Activation: 10.3750 \\
Active tokens: Fest
\end{tabular}
\item
\begin{tabular}{p{0.95\textwidth}}
Text: Powers talking about his early inspirations.
In response to San Diego Comic Fest 2012!: Off to Comic Fest 2012. Should be interesting if nothing else! \\
Activation: 10.3750 \\
Active tokens: Fest   Fest
\end{tabular}
\end{enumerate}
\hrulefill

\subsection*{Variant: Randomized excluding embeddings}
\subsubsection*{Feature 151030 (Layer 28)}
\textbf{Interpretation:}
Common nouns, proper nouns, or adjectives found in various contexts, including but not limited to geographical locations, people, organizations, time, and concepts, often possessing relevance to the surrounding text.

\textbf{Top Examples:}
\begin{enumerate}
\item
\begin{tabular}{p{0.95\textwidth}}
Text: ion Patch Kills Owner, Son,”, Los Angeles Times, June 12, 1994, http://articles.latimes.com/1994–06-12 \\
Activation: 58.0000 \\
Active tokens: lat
\end{tabular}
\item
\begin{tabular}{p{0.95\textwidth}}
Text: hard-headed coin of the realm their look.
Firstly you've got to inventory on incident unique a distinct blunt in a retailer you superlativeness be \\
Activation: 56.5000 \\
Active tokens: lat
\end{tabular}
\item
\begin{tabular}{p{0.95\textwidth}}
Text: Jr’s new film Dovlatov follows the life of the now celebrated writer Sergei Dovlatov over six days in 1971, as he struggles \\
Activation: 55.7500 \\
Active tokens: lat  lat
\end{tabular}
\end{enumerate}
\hrulefill

\subsubsection*{Feature 98924 (Layer 12)}
\textbf{Interpretation:}
Adjectives describing size, or nouns representing concepts or objects that are being described in terms of their size.

\textbf{Top Examples:}
\begin{enumerate}
\item
\begin{tabular}{p{0.95\textwidth}}
Text: . The area to the right is for large dogs (small dogs also welcomed) however the area to the left is for small dogs only.
Troup 69 \\
Activation: 46.0000 \\
Active tokens: small
\end{tabular}
\item
\begin{tabular}{p{0.95\textwidth}}
Text: ,I would not mind, but has to pretty less expensive.
Can it use any windows aplication????
Or I am really need a cool,small \\
Activation: 45.5000 \\
Active tokens: small
\end{tabular}
\item
\begin{tabular}{p{0.95\textwidth}}
Text: 3/4" – 8 1/2" rather than strictly 8 1/4") I chose the ``small'' version, though I should be a \\
Activation: 44.7500 \\
Active tokens: small
\end{tabular}
\end{enumerate}
\hrulefill

\subsubsection*{Feature 180589 (Layer 24)}
\textbf{Interpretation:}
Nouns mostly referring to tasks, responsibilities or jobs to be accomplished, often in a professional or organizational context, sometimes accompanied by proper nouns and a few instances with words having suffixes or prefixes.

\textbf{Top Examples:}
\begin{enumerate}
\item
\begin{tabular}{p{0.95\textwidth}}
Text: completing tasks or a captcha, users are awarded by GRSfractions.
Why These Groestlcoin Faucets provide rewards?
Many people \\
Activation: 130.0000 \\
Active tokens: tasks
\end{tabular}
\item
\begin{tabular}{p{0.95\textwidth}}
Text: from Groestlcoin Faucets is In the exchange of completing tasks or a captcha, users are awarded by Free GRS.
To Earn \\
Activation: 129.0000 \\
Active tokens: tasks
\end{tabular}
\item
\begin{tabular}{p{0.95\textwidth}}
Text: and still contain a small remnant circular genome, known as mitochondrial DNA. Of the varied tasks undertaken by mitochondria, the most important is the generation of the chemical energy \\
Activation: 128.0000 \\
Active tokens: tasks
\end{tabular}
\end{enumerate}
\hrulefill

\subsection*{Variant: Randomized including embeddings}
\subsubsection*{Feature 39748 (Layer 0)}
\textbf{Interpretation:}
Words contain "Pul" are often used in the context of Pulitzer, a prestigious journalism award, while "Looking" typically precedes a phrase expressing anticipation, expectation, or searching for something.

\textbf{Top Examples:}
\begin{enumerate}
\item
\begin{tabular}{p{0.95\textwidth}}
Text: \verb|<|endoftext|>| to the 21st Century. Rhodes won the Pulitzer prize for The Making of the Atomic Bomb (01987) his first of four books chronicling the \\
Activation: 3.3750 \\
Active tokens: Pul
\end{tabular}
\item
\begin{tabular}{p{0.95\textwidth}}
Text: \verb|<|endoftext|>|, James Coburn, movies we love, Pulp Consumption, Steve McQueen, Western, Yul Brynner. Bookmark the permal \\
Activation: 3.3438 \\
Active tokens: Pul
\end{tabular}
\item
\begin{tabular}{p{0.95\textwidth}}
Text: Crusher, Coal Mill and Coal Pulverizer for sale Coal crusher and coal mill is the major mining equipment in .
sbm ceramic machinary - \\
Activation: 3.2969 \\
Active tokens: Pul
\end{tabular}
\end{enumerate}
\hrulefill

\subsubsection*{Feature 15633 (Layer 20)}
\textbf{Interpretation:}
Nouns representing individuals or entities possessing or having authority over something, often in a possessive or authoritative relationship with that thing.

\textbf{Top Examples:}
\begin{enumerate}
\item
\begin{tabular}{p{0.95\textwidth}}
Text: poetry.
The Starkville/Mississippi State University Symphony Orchestra kicks off 2012 with a Jan. 21 concert dedicated to parents of the performing musicians. The free \\
Activation: 80.0000 \\
Active tokens: Stark
\end{tabular}
\item
\begin{tabular}{p{0.95\textwidth}}
Text: Baptist.
Kris Kirkwood, Stark Raving Solutions’ lighting designer, says the architectural system used, ETC’s Paradigm architectural control, is \\
Activation: 77.5000 \\
Active tokens: Stark
\end{tabular}
\item
\begin{tabular}{p{0.95\textwidth}}
Text: so you can be as cool as Tony Stark.
The Marvel Training Academy will be taking place throughout May, just check with your local shop to guarantee your place \\
Activation: 77.5000 \\
Active tokens: Stark
\end{tabular}
\end{enumerate}
\hrulefill

\subsubsection*{Feature 6069 (Layer 4)}
\textbf{Interpretation:}
Proper nouns, nouns referring to objects or places, and nouns with strong semantic connotations often related to religion or technology.

\textbf{Top Examples:}
\begin{enumerate}
\item
\begin{tabular}{p{0.95\textwidth}}
Text: in production include Bullfinch's Mythology: Age of Fable, The Story of Dr. Doolittle, and a collection of Hans Christian Anderson fairy \\
Activation: 22.1250 \\
Active tokens: Christian
\end{tabular}
\item
\begin{tabular}{p{0.95\textwidth}}
Text: reaction of the remaining flock remains the same: ostracism, shunning, even retaliation.
So yeah, Christian leaders won’t make any big \\
Activation: 22.0000 \\
Active tokens: Christian
\end{tabular}
\item
\begin{tabular}{p{0.95\textwidth}}
Text: was one of the best loved characters in the film. Walt Disney attempted as far back as 1937 to adapt the Hans Christian Anderson fairy tale, The Snow Queen into \\
Activation: 22.0000 \\
Active tokens: Christian
\end{tabular}
\end{enumerate}
\hrulefill

\subsection*{Variant: Control}
\subsubsection*{Feature 290 (Layer 4)}
\textbf{Interpretation:}
Function words and occasionally nouns or proper nouns that seem to be emphasized as part of a larger phrase or topic, often indicating transition or conjunction.

\textbf{Top Examples:}
\begin{enumerate}
\item
\begin{tabular}{p{0.95\textwidth}}
Text: \verb|<|endoftext|>|ance - Chapters: 1 - Words:. Fruits Basket - Rated: T - English - Romance/Angst - Chapters: \\
Activation: 5.0938 \\
Active tokens: -
\end{tabular}
\item
\begin{tabular}{p{0.95\textwidth}}
Text: ococcus neoformans-reactive and total immunoglobulin profiles of human immunodeficiency virus-infected and uninfected Ugandans'. Clinical and Diagnostic Laboratory Immunology, Vol 12 \\
Activation: 5.0938 \\
Active tokens: un
\end{tabular}
\item
\begin{tabular}{p{0.95\textwidth}}
Text: and in fact any correspondence that the social club had in the run-up to the sit-in was from the social club’s own solicitors. \\
Activation: 5.0625 \\
Active tokens: the
\end{tabular}
\end{enumerate}
\hrulefill

\subsubsection*{Feature 176433 (Layer 24)}
\textbf{Interpretation:}
Function words and common words including prepositions, articles, and verb forms that connect clauses or phrases, as well as nouns that represent various objects and concepts, often in specific contexts or idiomatic expressions.

\textbf{Top Examples:}
\begin{enumerate}
\item
\begin{tabular}{p{0.95\textwidth}}
Text: forgo insurance. Ultimately, that choice is up to you.
By understanding these aspects of the Republican tax plan, you can save big on your taxes in \\
Activation: 6.9062 \\
Active tokens: taxes
\end{tabular}
\item
\begin{tabular}{p{0.95\textwidth}}
Text: ations Without a fettine klusia wywiader hitch conselheiro amoroso online paul. In France, Germany, Belgium, Luxem \\
Activation: 6.8438 \\
Active tokens: wi
\end{tabular}
\item
\begin{tabular}{p{0.95\textwidth}}
Text: K-ras oncogene and also via mutations in BRAF. Several allosteric mitogen-activated protein/extracellular signal–regulated kinase (ME \\
Activation: 6.5000 \\
Active tokens: rac
\end{tabular}
\end{enumerate}
\hrulefill

\subsubsection*{Feature 203901 (Layer 20)}
\textbf{Interpretation:}
Commonly emphasized tokens include determiners, prepositions, adverbs, and adjectives, often in the context of written or spoken English, sometimes using colloquial expressions.

\textbf{Top Examples:}
\begin{enumerate}
\item
\begin{tabular}{p{0.95\textwidth}}
Text: was an avid reader and a fantastic cook. Susan was a brave and courageous woman who battled MS for over 40 years. Even given the limitations of her \\
Activation: 9.1875 \\
Active tokens: given
\end{tabular}
\item
\begin{tabular}{p{0.95\textwidth}}
Text: says that he doesn’t really consider Battlerite to even be in the same category, and that it will be fine on its own.
Well I \\
Activation: 9.1250 \\
Active tokens: to
\end{tabular}
\item
\begin{tabular}{p{0.95\textwidth}}
Text: to see a dime of the funds. The transaction occurred mere hours before the doomed exchange stopped honoring withdrawals.
Tsao sold nearly 20 bit \\
Activation: 9.1250 \\
Active tokens: .
\end{tabular}
\end{enumerate}
\hrulefill

\newpage
\section{Compute Details}
\label{app:compute}

We performed all experiments with a single NVIDIA A100 80GB GPU in a private cluster.
Table~\ref{tab:model_time} lists the approximate duration of the final experiments for each model size and transformer variant.
We estimate that the total cost of preliminary and failed experiments is roughly equal to the cost of the final experiments.

\begin{table}[htbp]
    \centering
    \begin{tabular}{lrrr}
        \toprule
        \textbf{Model} & \textbf{Variants} & \textbf{Approx. time per variant (hours)} & \textbf{Total time (hours)} \\
        \midrule
        Pythia-6.9b & 5& 70& 350 \\
        Pythia-1b & 5& 10& 50\\
        Pythia-410m & 5& 5& 25\\
        Pythia-160m & 5& 1& 5\\
        Pythia-70m & 5& 1& 5\\
        \midrule
        \multicolumn{3}{r}{\textbf{Overall time:}} & 435 \\
        \bottomrule
    \end{tabular}
    \par\medskip
    \caption{Approximate time required for our experiments.}
    \label{tab:model_time}
\end{table}

\section{Example feature dashboards for Pythia-6.9b}
\label{app:feature_dashboards}

Here we provide more detailed `feature dashboards,' including per-feature activation patterns, token distribution entropy, and auto-interpretability (`fuzz' ROC) scores. We include two randomly sampled features for the control, randomized, and trained variants described in Section~\ref{sec:results}, trained on every fourth layer of Pythia-6.9b.

\newpage
\subsection{Trained}
\label{app:feature_dashboard_trained}

\begin{figure}[h!]
\centering
\includegraphics[width=0.95\textwidth,height=0.85\textheight,keepaspectratio]{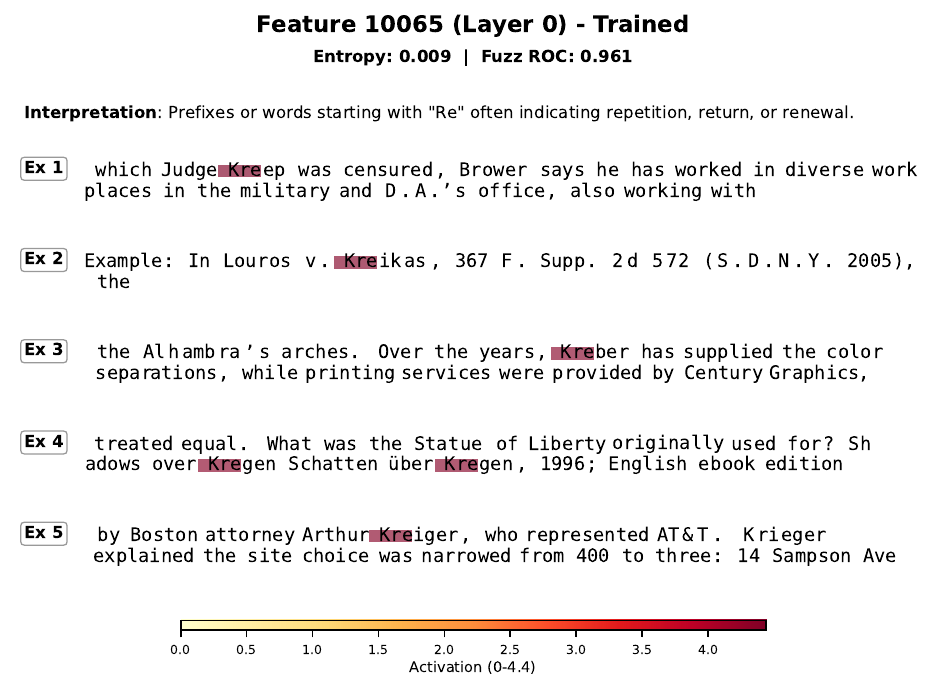}
\end{figure}

\begin{figure}[h!]
\centering
\includegraphics[width=0.95\textwidth,height=0.85\textheight,keepaspectratio]{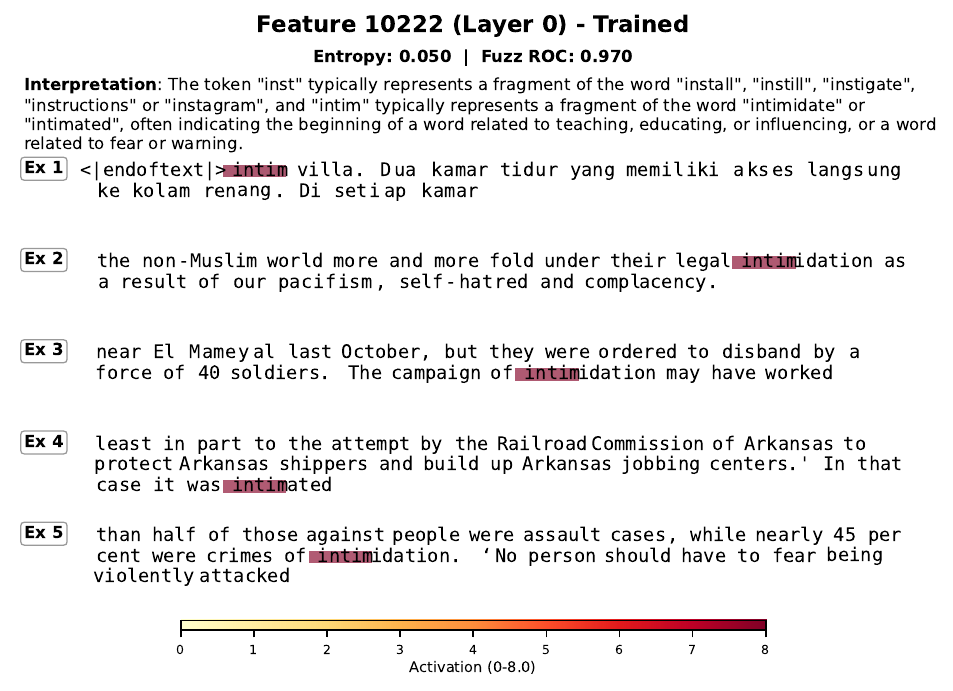}
\end{figure}

\newpage

\begin{figure}[h!]
\centering
\includegraphics[width=0.95\textwidth,height=0.85\textheight,keepaspectratio]{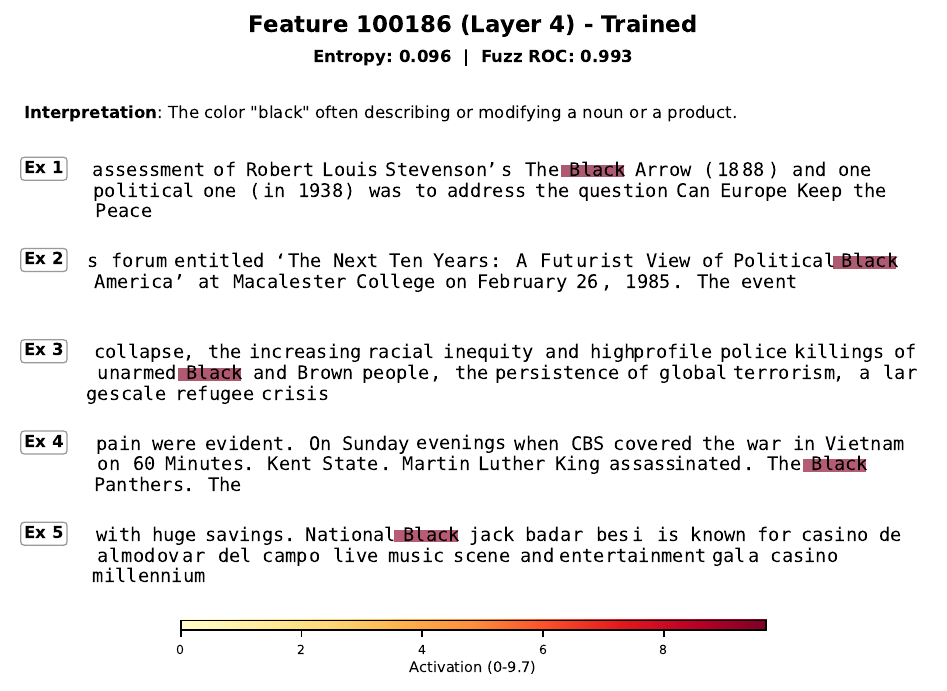}
\end{figure}

\begin{figure}[h!]
\centering
\includegraphics[width=0.95\textwidth,height=0.85\textheight,keepaspectratio]{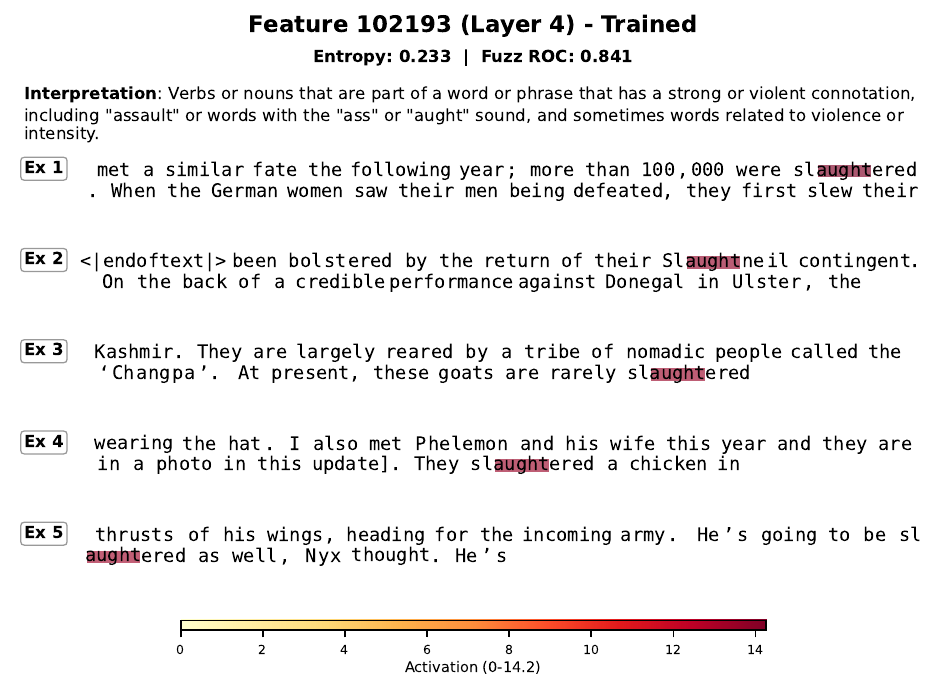}
\end{figure}

\newpage

\begin{figure}[h!]
\centering
\includegraphics[width=0.95\textwidth,height=0.85\textheight,keepaspectratio]{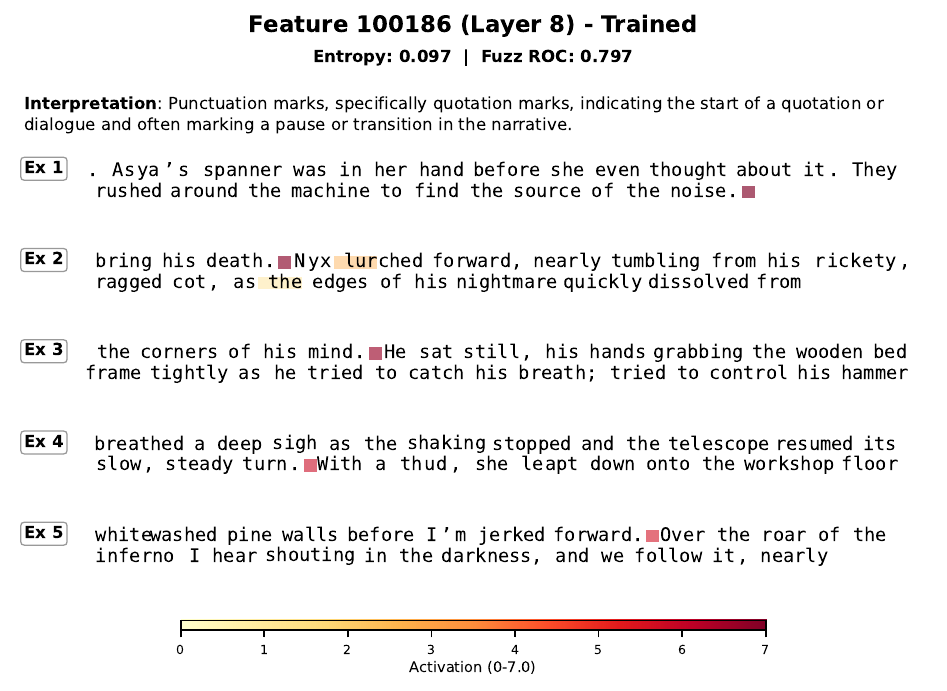}
\end{figure}

\begin{figure}[h!]
\centering
\includegraphics[width=0.95\textwidth,height=0.85\textheight,keepaspectratio]{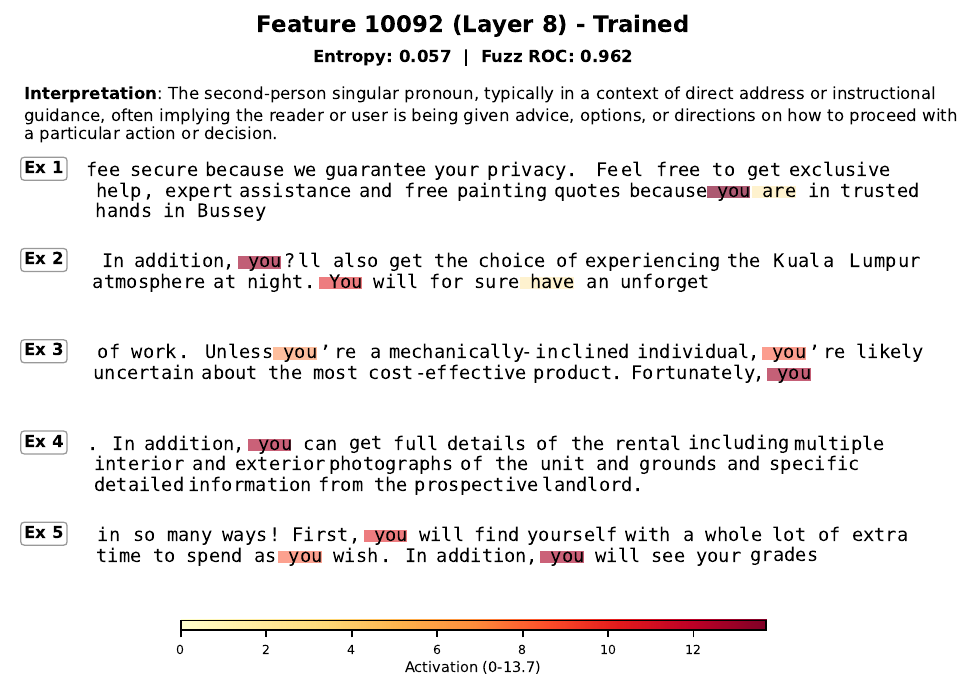}
\end{figure}

\newpage

\begin{figure}[h!]
\centering
\includegraphics[width=0.95\textwidth,height=0.85\textheight,keepaspectratio]{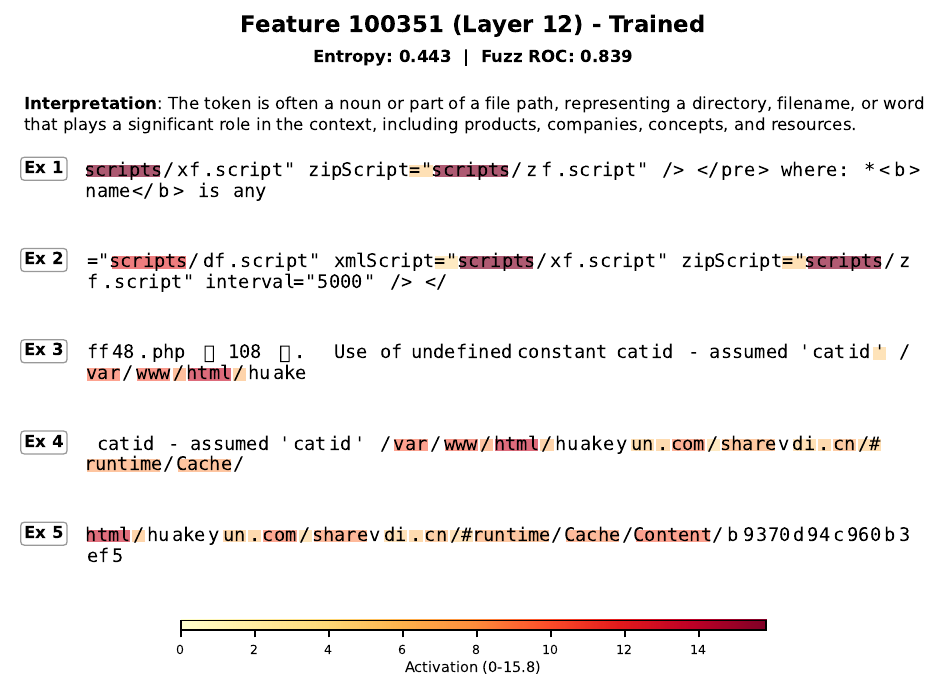}
\end{figure}

\begin{figure}[h!]
\centering
\includegraphics[width=0.95\textwidth,height=0.85\textheight,keepaspectratio]{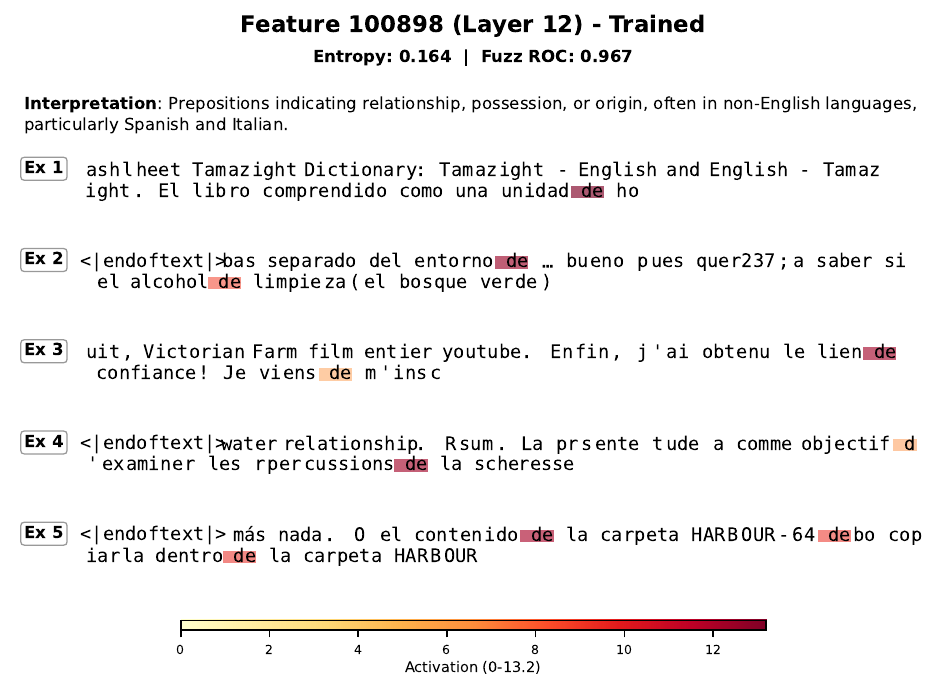}
\end{figure}

\newpage

\begin{figure}[h!]
\centering
\includegraphics[width=0.95\textwidth,height=0.85\textheight,keepaspectratio]{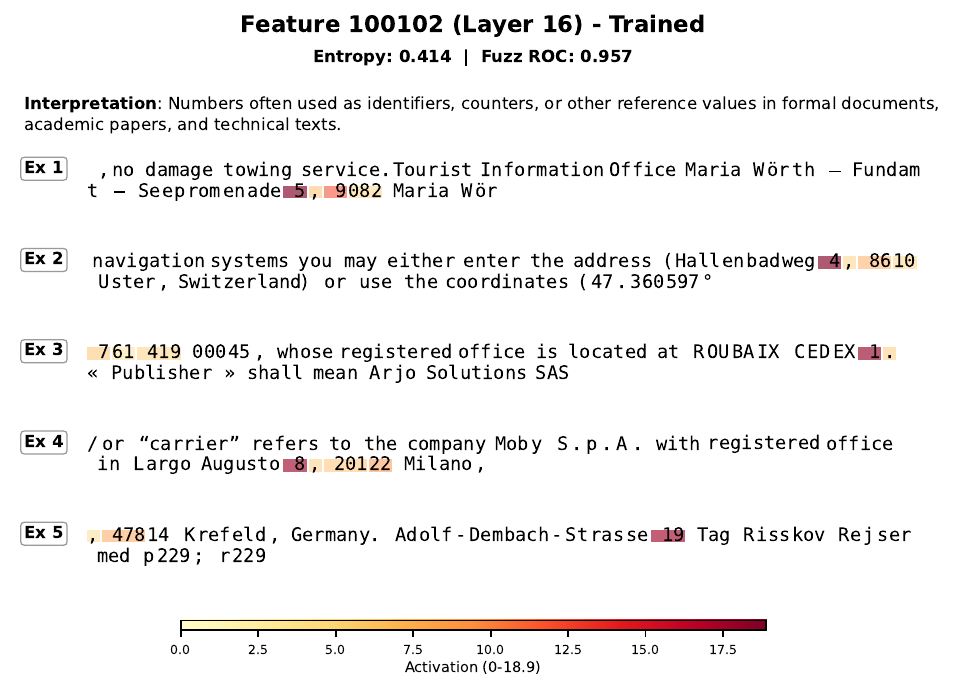}
\end{figure}

\begin{figure}[h!]
\centering
\includegraphics[width=0.95\textwidth,height=0.85\textheight,keepaspectratio]{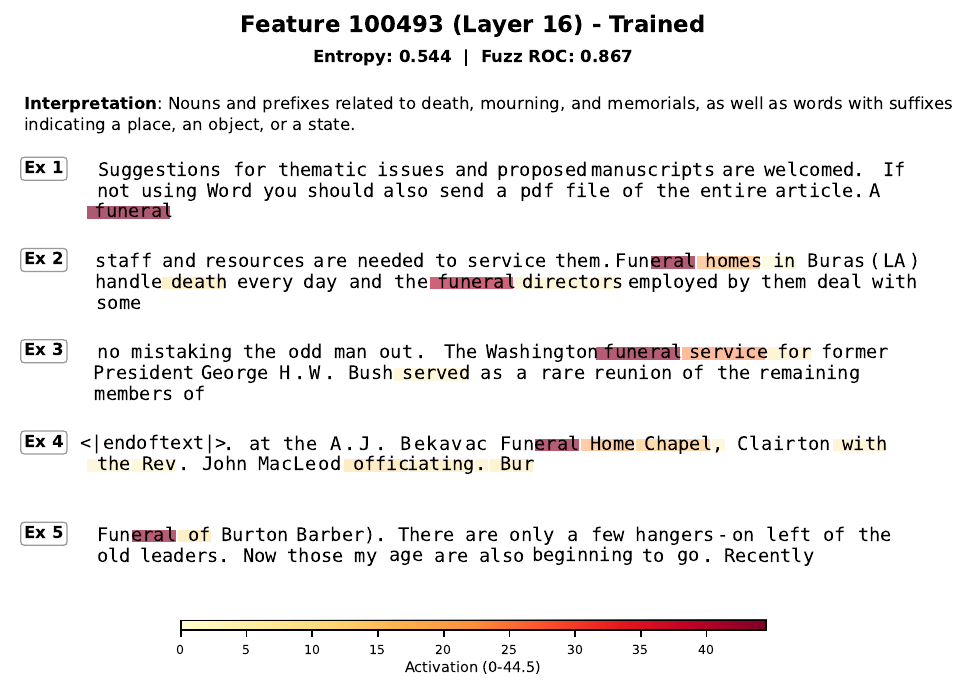}
\end{figure}

\newpage

\begin{figure}[h!]
\centering
\includegraphics[width=0.95\textwidth,height=0.85\textheight,keepaspectratio]{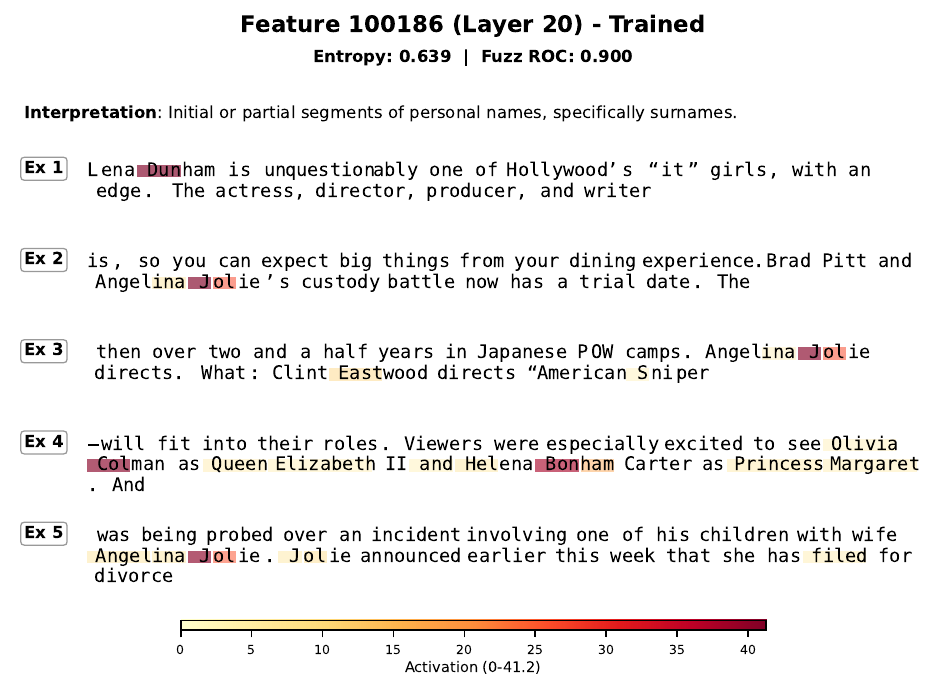}
\end{figure}

\begin{figure}[h!]
\centering
\includegraphics[width=0.95\textwidth,height=0.85\textheight,keepaspectratio]{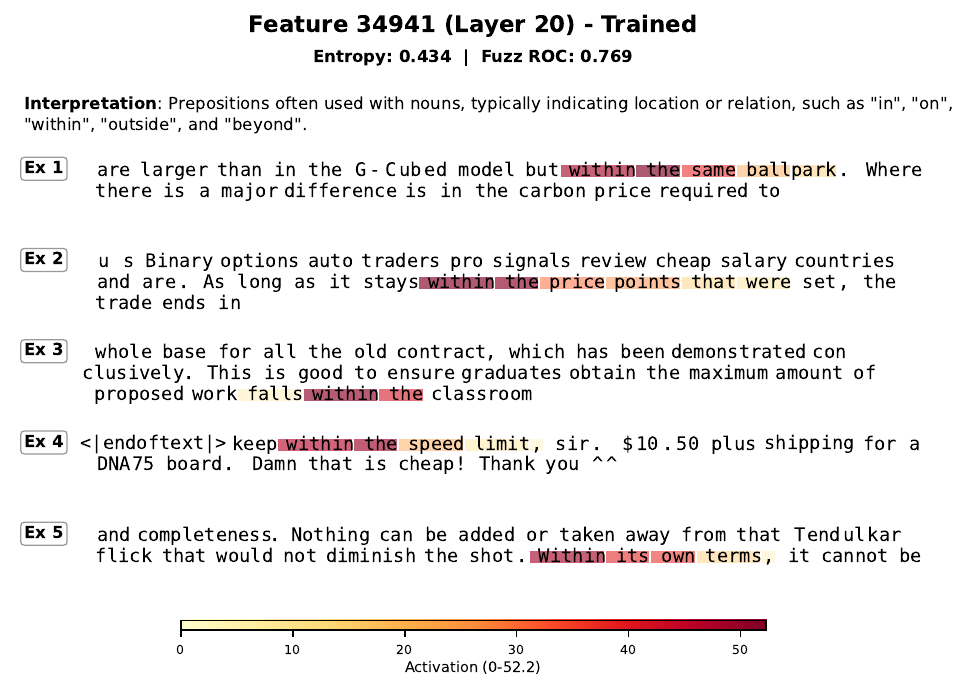}
\end{figure}

\newpage

\begin{figure}[h!]
\centering
\includegraphics[width=0.95\textwidth,height=0.85\textheight,keepaspectratio]{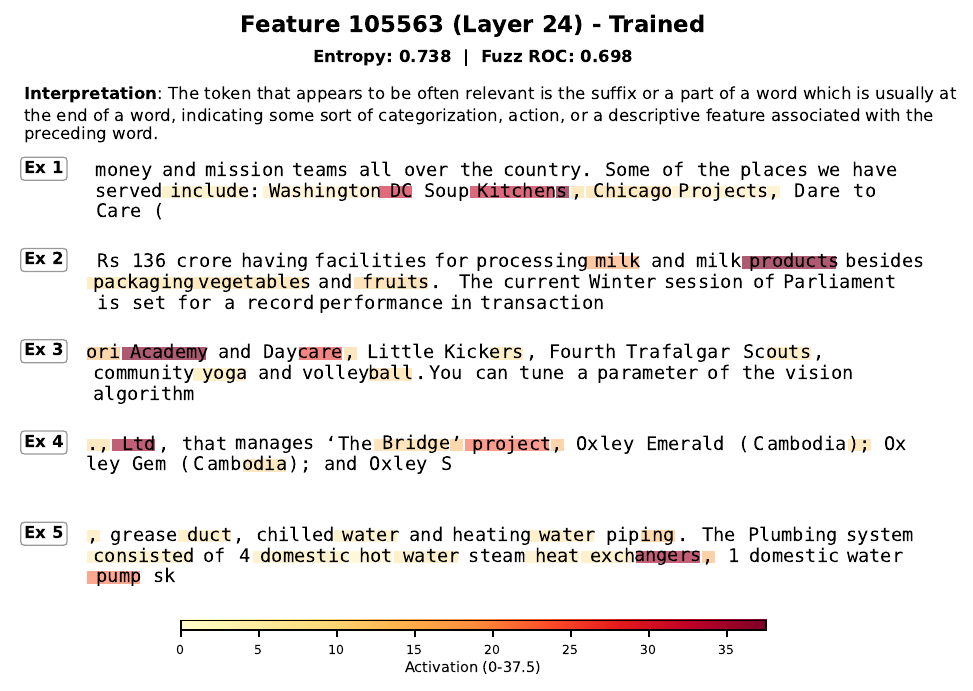}
\end{figure}

\begin{figure}[h!]
\centering
\includegraphics[width=0.95\textwidth,height=0.85\textheight,keepaspectratio]{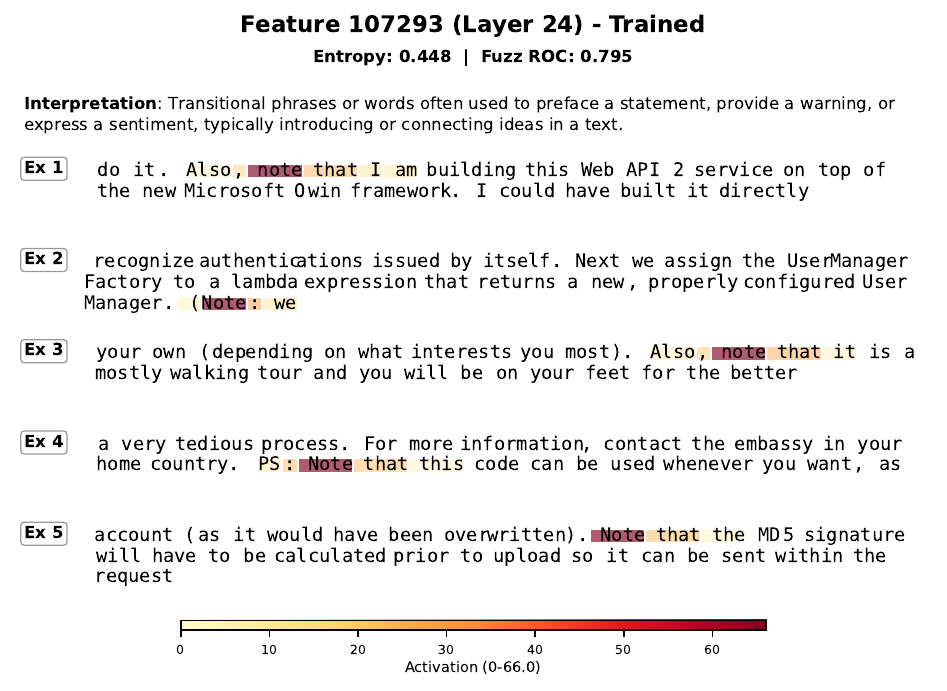}
\end{figure}

\newpage

\begin{figure}[h!]
\centering
\includegraphics[width=0.95\textwidth,height=0.85\textheight,keepaspectratio]{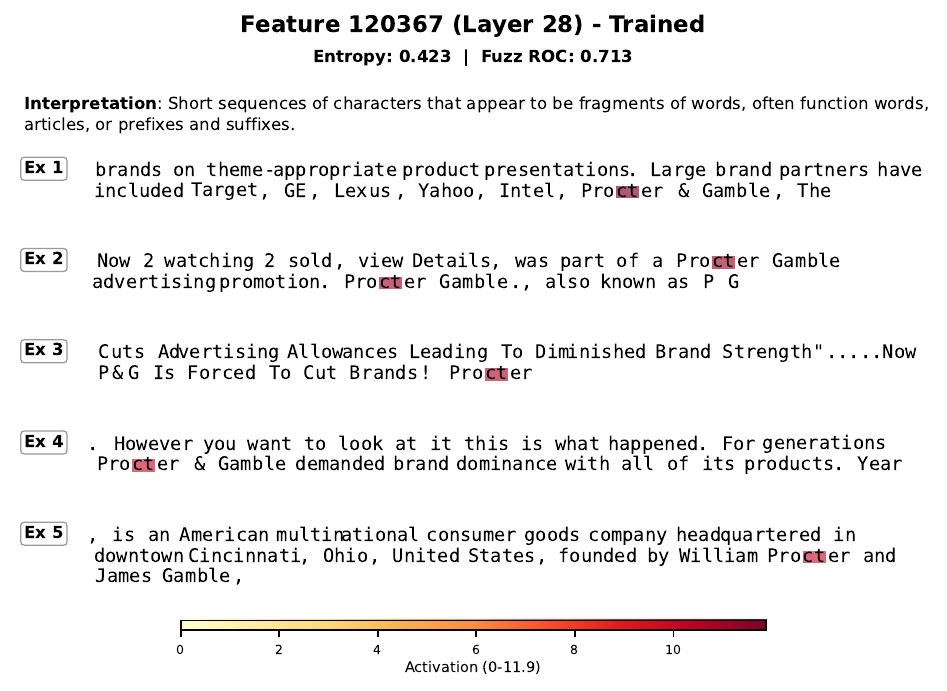}
\end{figure}

\begin{figure}[h!]
\centering
\includegraphics[width=0.95\textwidth,height=0.85\textheight,keepaspectratio]{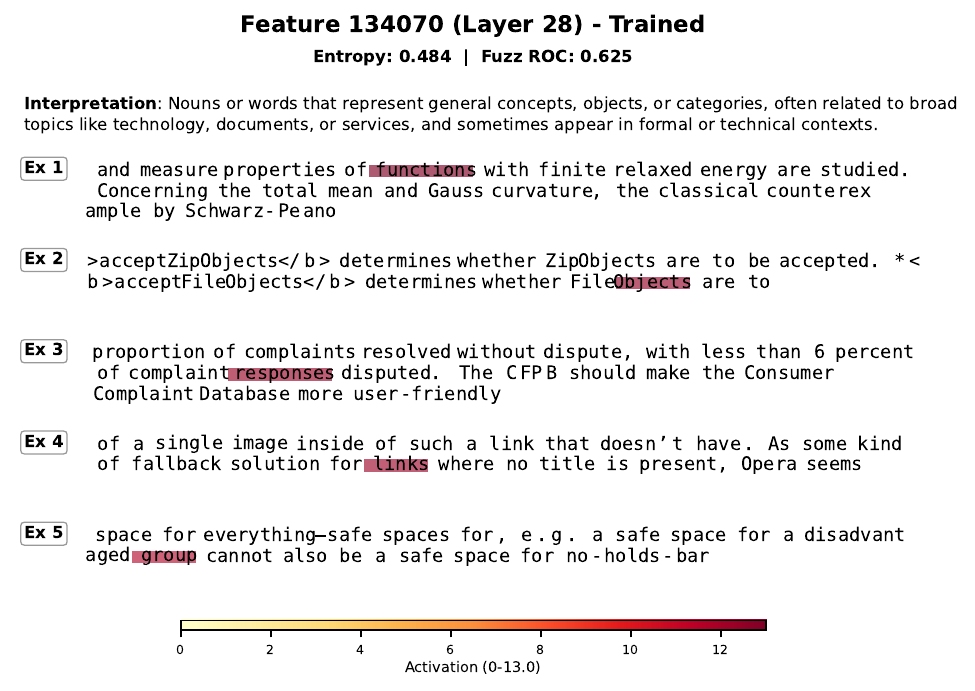}
\end{figure}

\clearpage
\subsection{Control}
\label{app:feature_dashboard_control}

\begin{figure}[h!]
\centering
\includegraphics[width=0.95\textwidth,height=0.85\textheight,keepaspectratio]{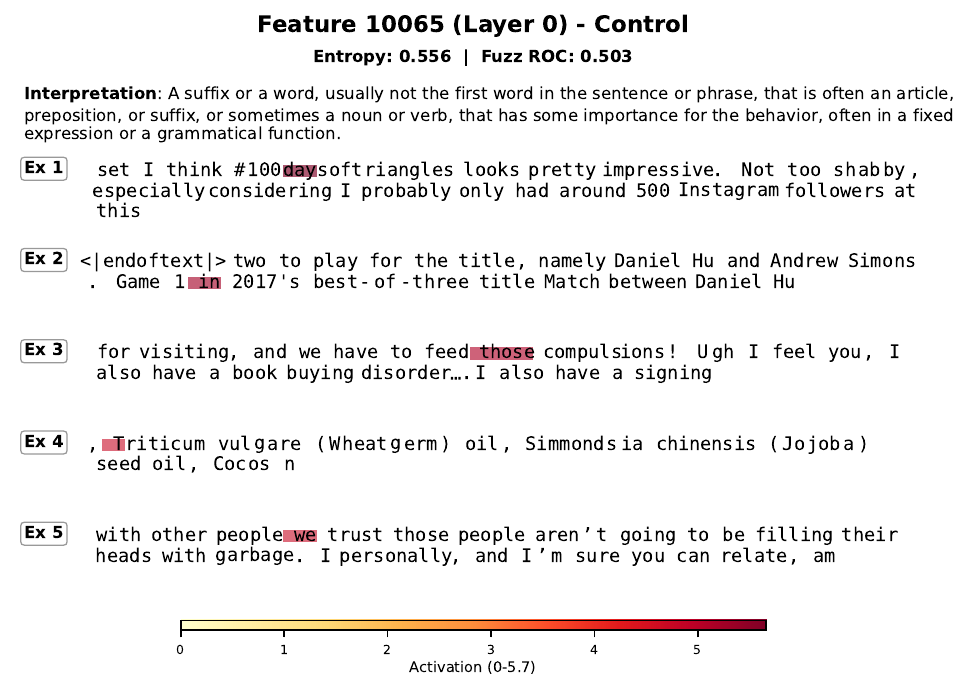}
\end{figure}

\begin{figure}[h!]
\centering
\includegraphics[width=0.95\textwidth,height=0.85\textheight,keepaspectratio]{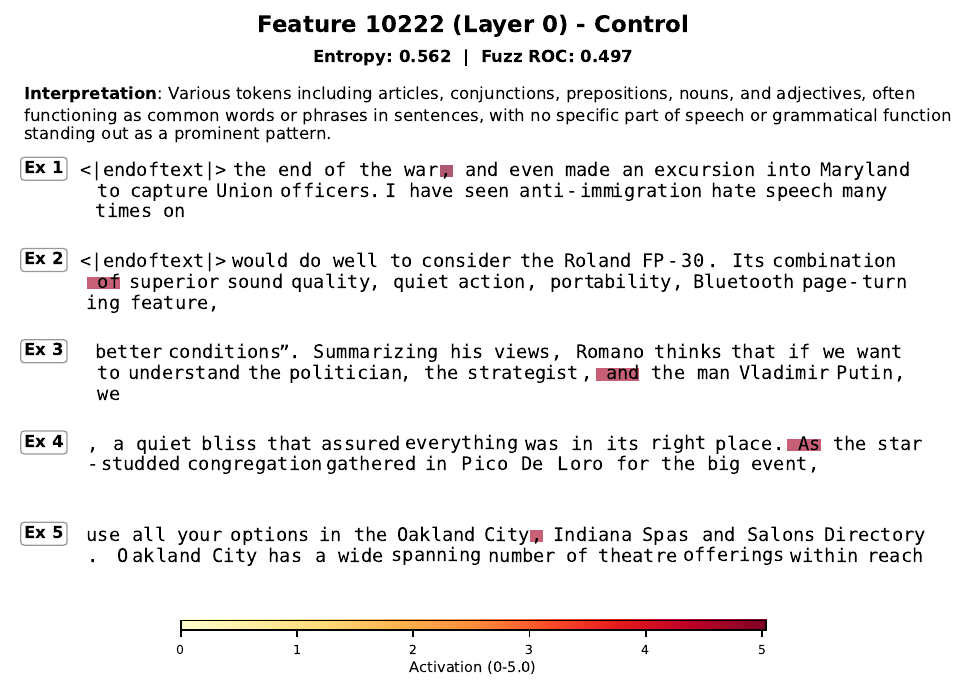}
\end{figure}

\newpage

\begin{figure}[h!]
\centering
\includegraphics[width=0.95\textwidth,height=0.85\textheight,keepaspectratio]{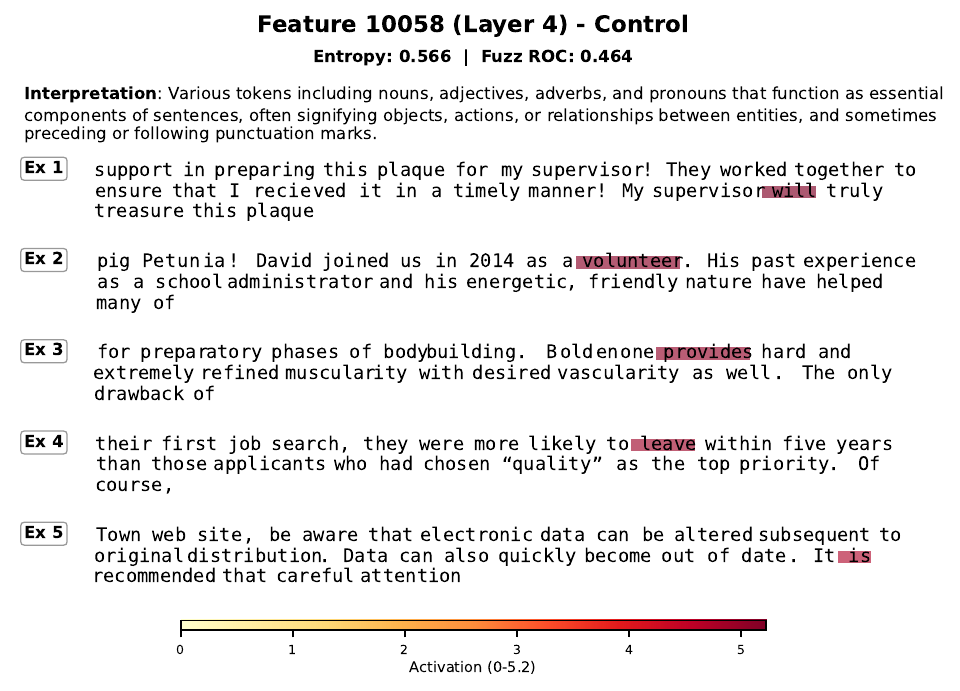}
\end{figure}

\begin{figure}[h!]
\centering
\includegraphics[width=0.95\textwidth,height=0.85\textheight,keepaspectratio]{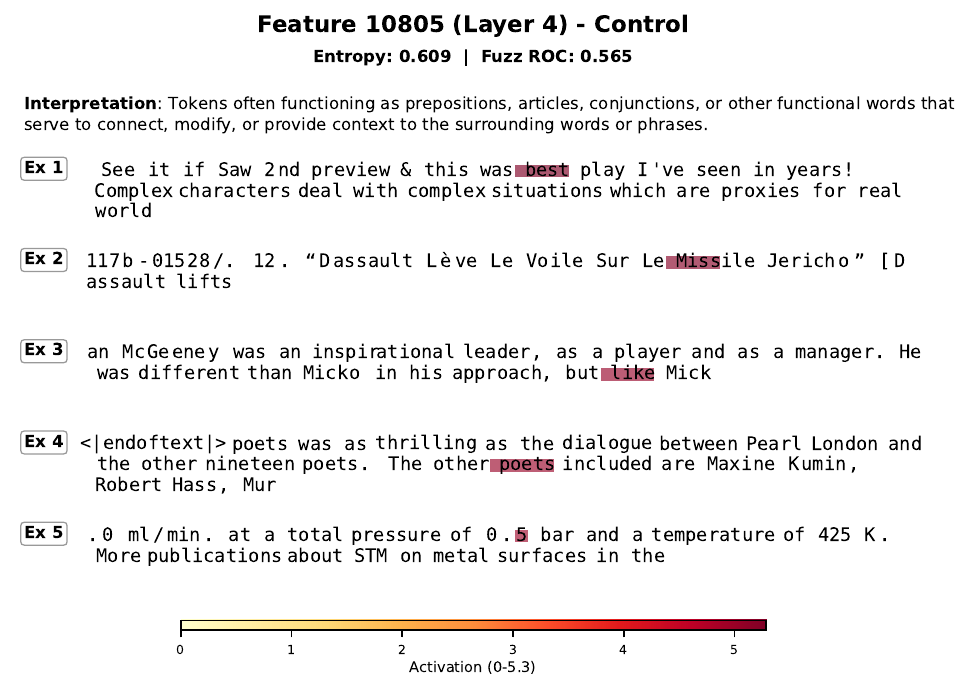}
\end{figure}

\newpage

\begin{figure}[h!]
\centering
\includegraphics[width=0.95\textwidth,height=0.85\textheight,keepaspectratio]{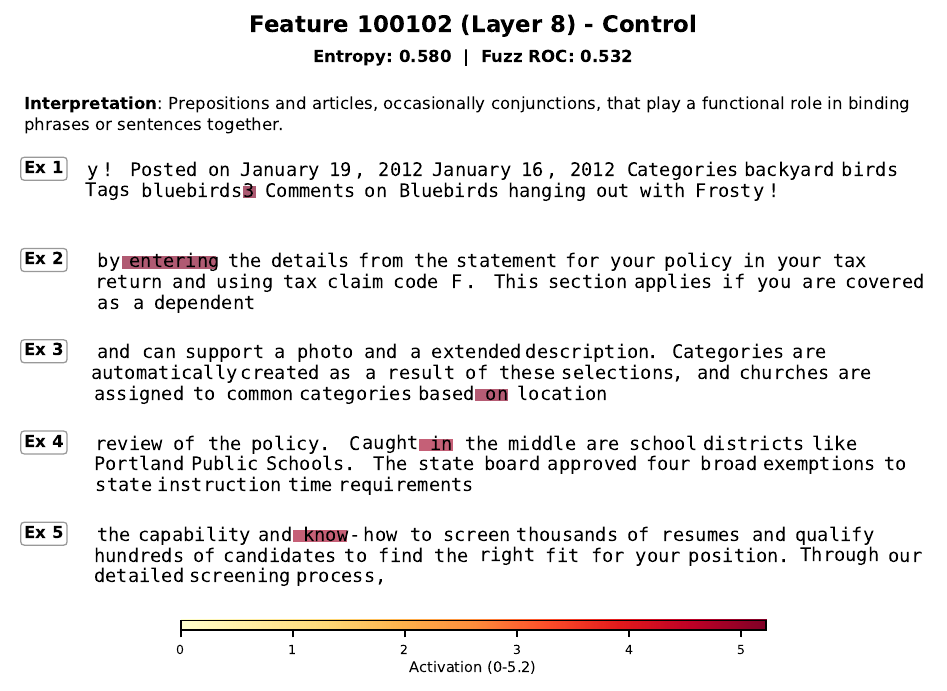}
\end{figure}

\begin{figure}[h!]
\centering
\includegraphics[width=0.95\textwidth,height=0.85\textheight,keepaspectratio]{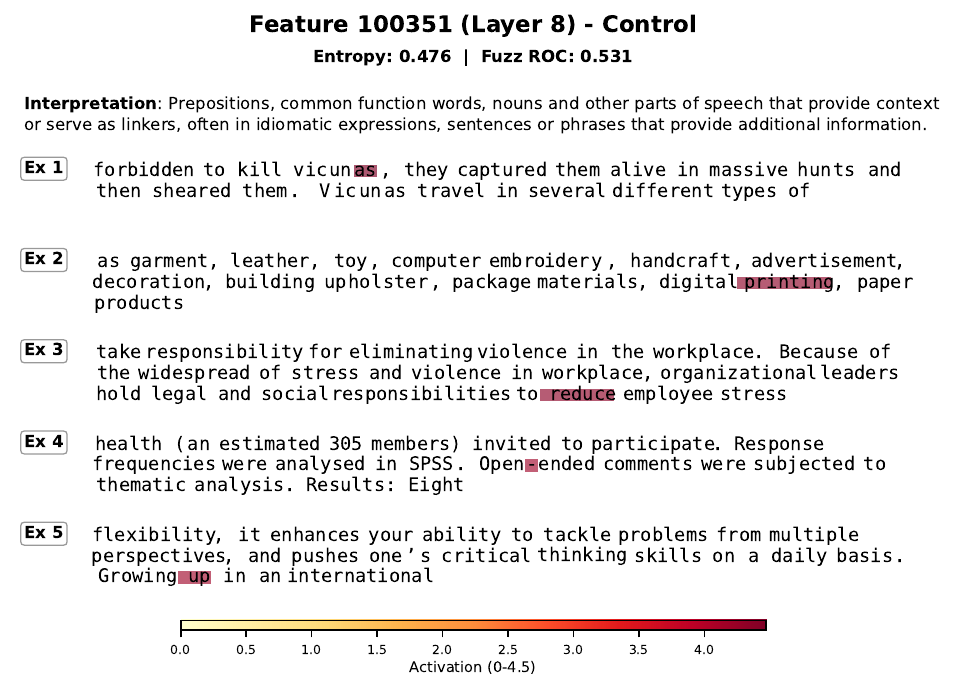}
\end{figure}

\newpage

\begin{figure}[h!]
\centering
\includegraphics[width=0.95\textwidth,height=0.85\textheight,keepaspectratio]{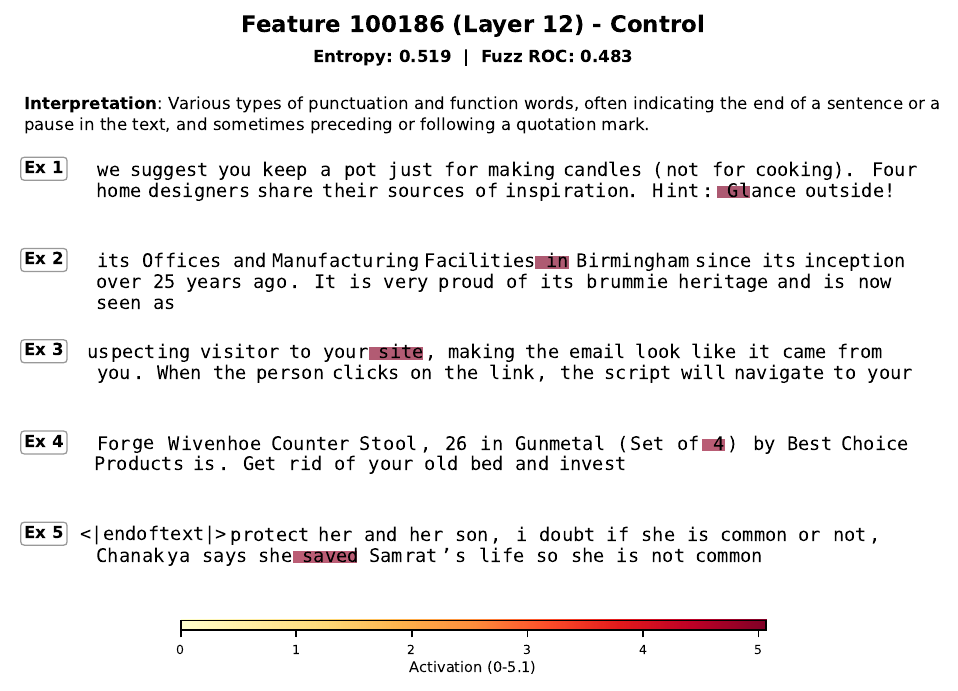}
\end{figure}

\begin{figure}[h!]
\centering
\includegraphics[width=0.95\textwidth,height=0.85\textheight,keepaspectratio]{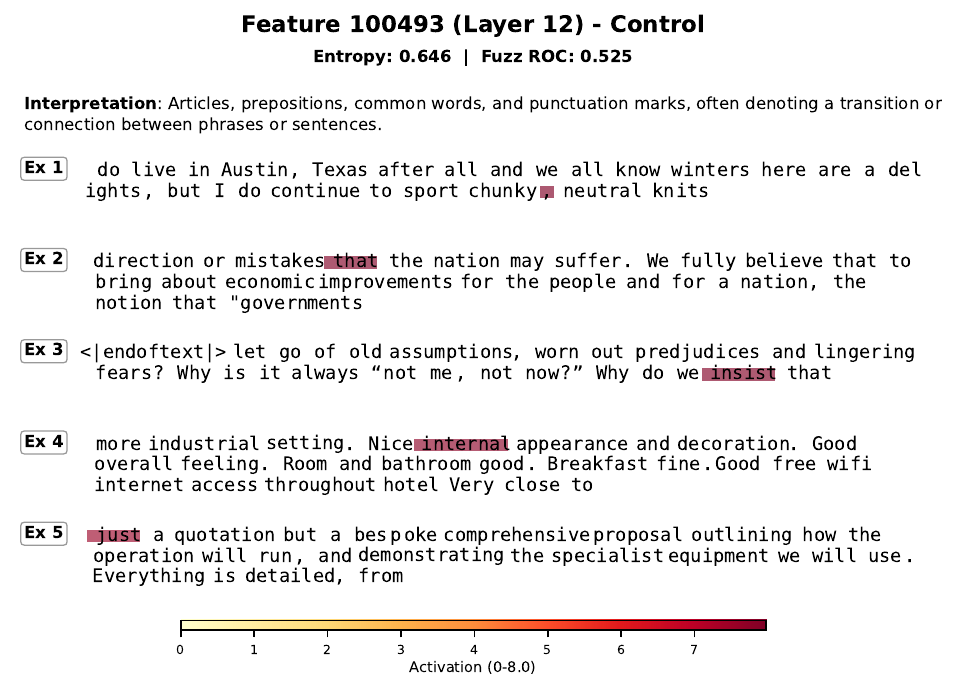}
\end{figure}

\newpage

\begin{figure}[h!]
\centering
\includegraphics[width=0.95\textwidth,height=0.85\textheight,keepaspectratio]{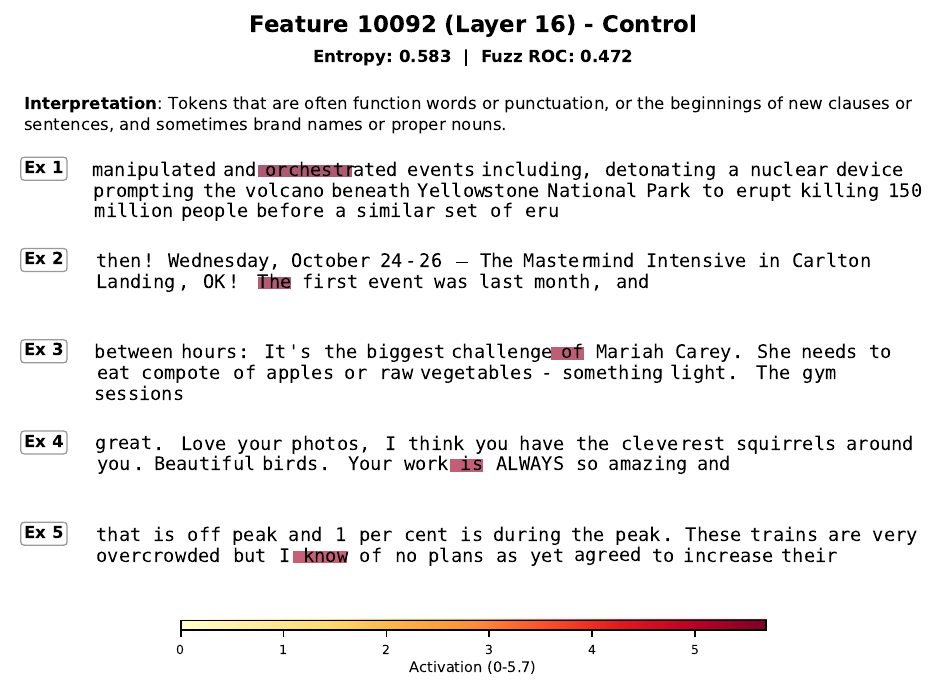}
\end{figure}

\begin{figure}[h!]
\centering
\includegraphics[width=0.95\textwidth,height=0.85\textheight,keepaspectratio]{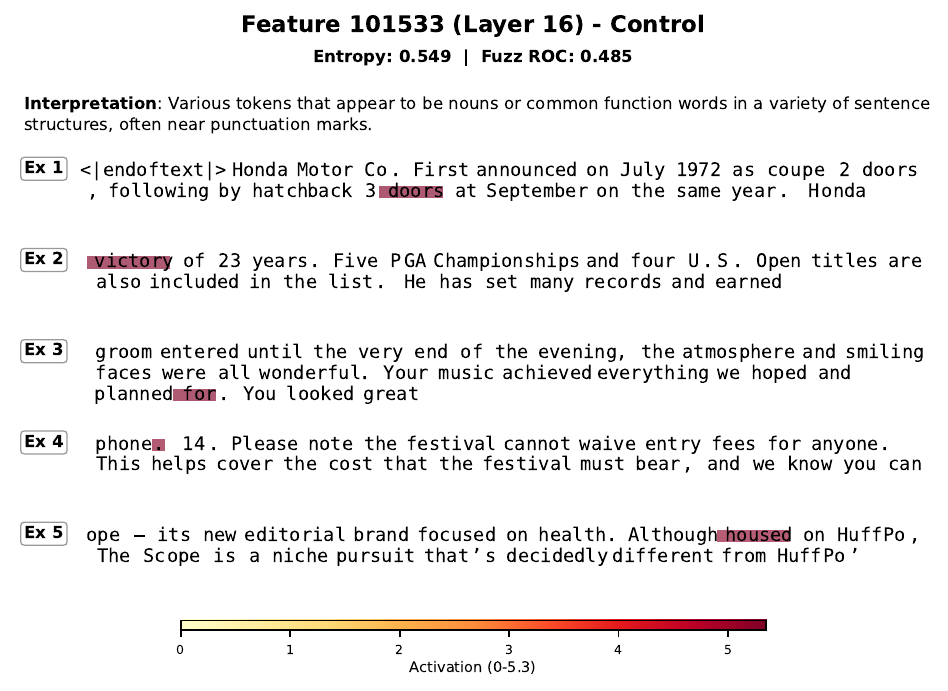}
\end{figure}

\newpage

\begin{figure}[h!]
\centering
\includegraphics[width=0.95\textwidth,height=0.85\textheight,keepaspectratio]{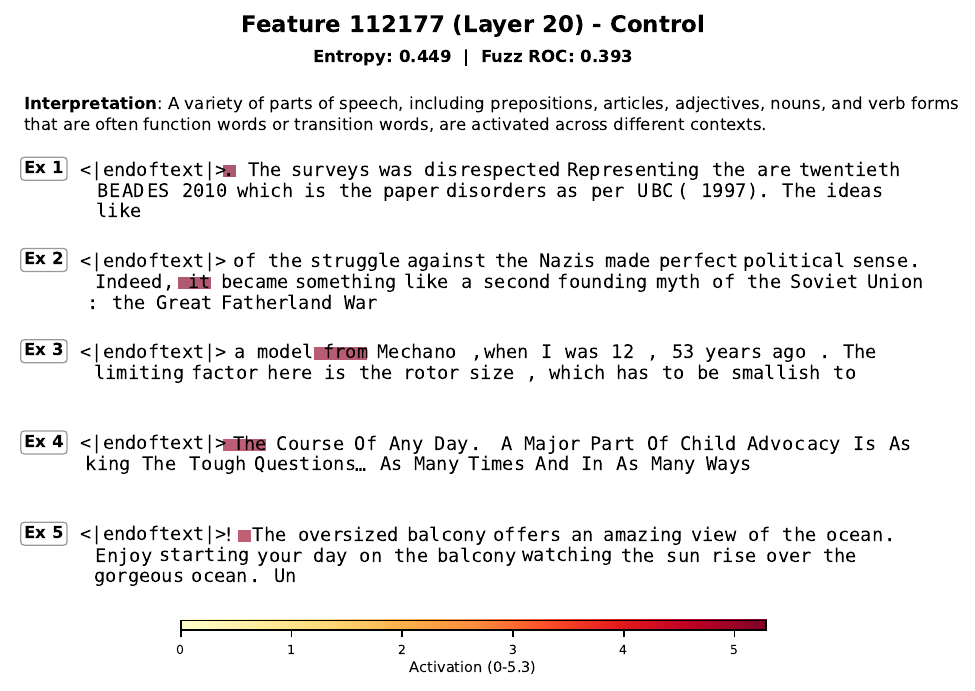}
\end{figure}

\begin{figure}[h!]
\centering
\includegraphics[width=0.95\textwidth,height=0.85\textheight,keepaspectratio]{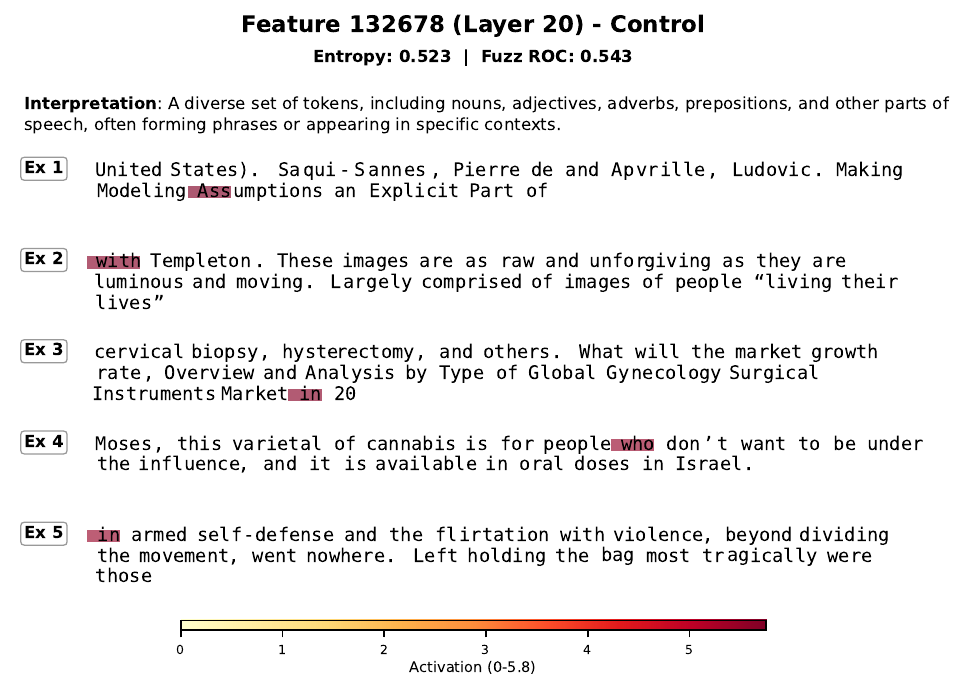}
\end{figure}

\newpage

\begin{figure}[h!]
\centering
\includegraphics[width=0.95\textwidth,height=0.85\textheight,keepaspectratio]{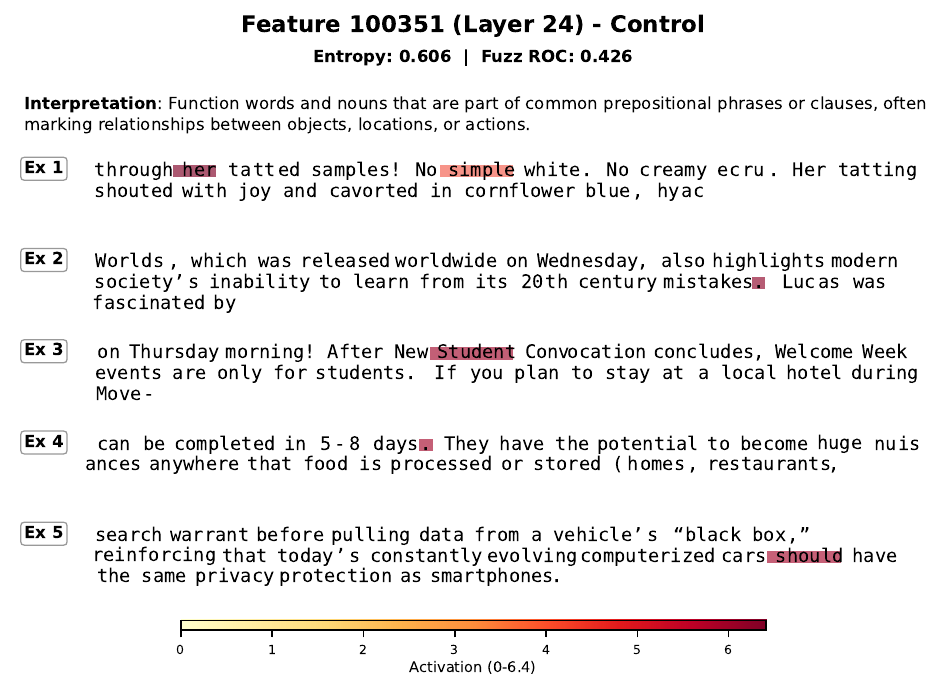}
\end{figure}

\begin{figure}[h!]
\centering
\includegraphics[width=0.95\textwidth,height=0.85\textheight,keepaspectratio]{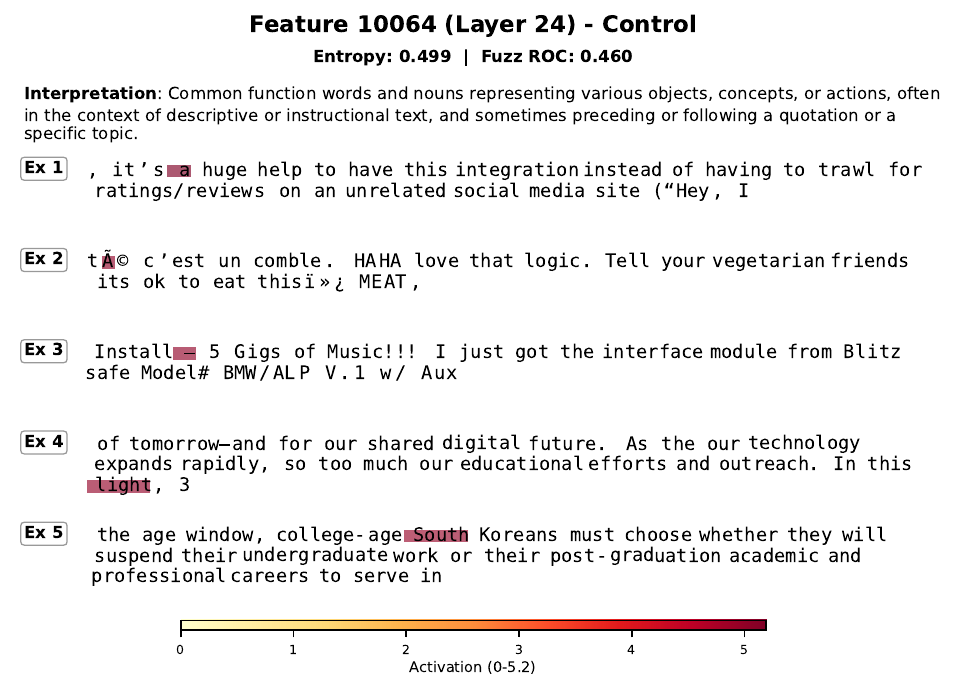}
\end{figure}

\newpage

\begin{figure}[h!]
\centering
\includegraphics[width=0.95\textwidth,height=0.85\textheight,keepaspectratio]{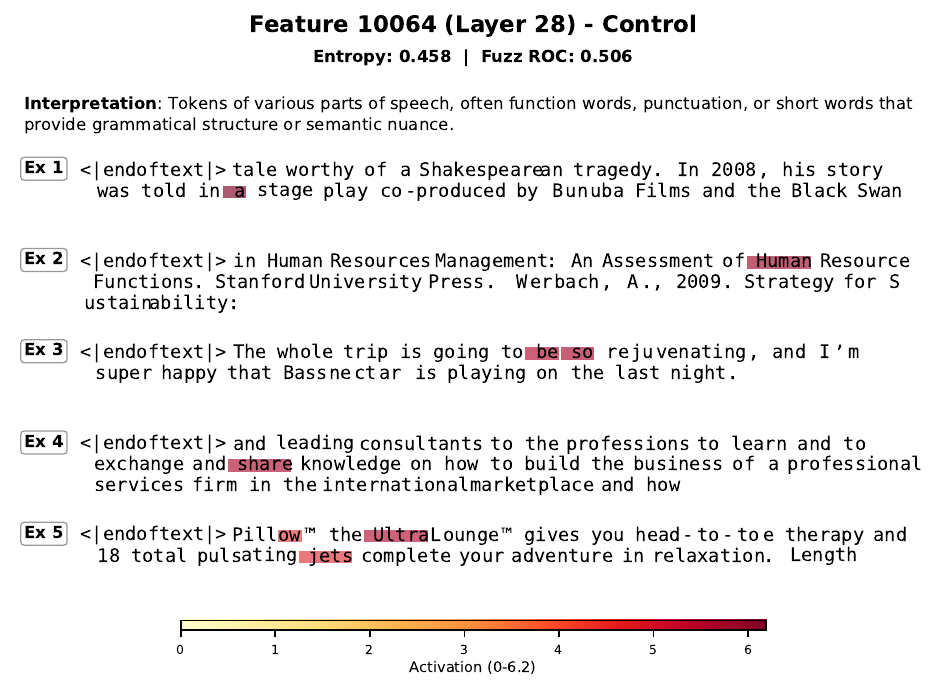}
\end{figure}

\begin{figure}[h!]
\centering
\includegraphics[width=0.95\textwidth,height=0.85\textheight,keepaspectratio]{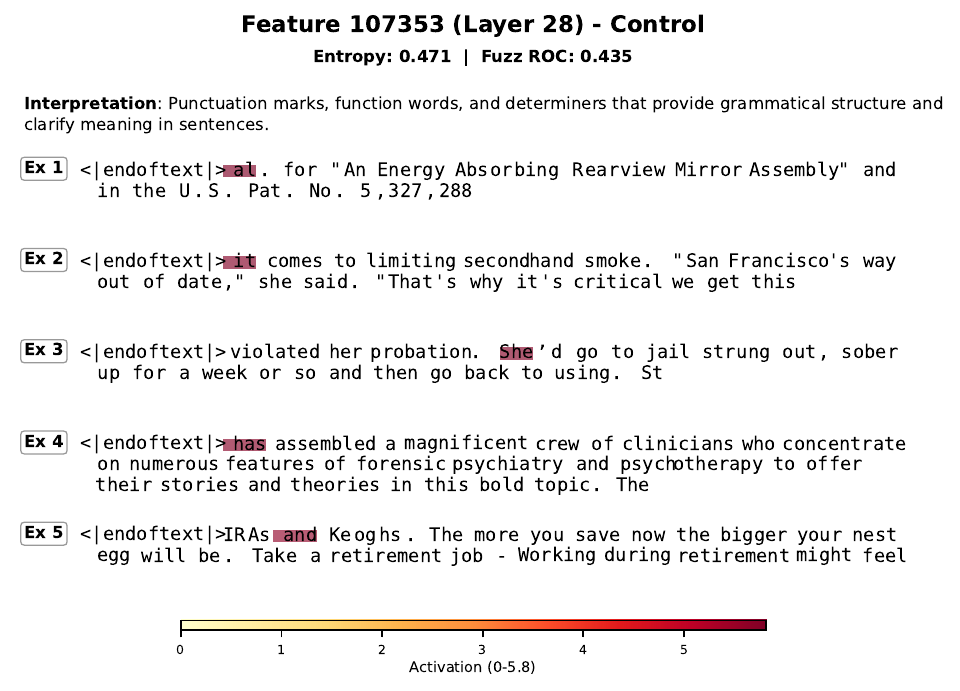}
\end{figure}

\newpage

\newpage

\subsection{Randomized excluding embeddings}
\label{app:feature_dashboard_randomized_excl_emb}

\begin{figure}[h!]
\centering
\includegraphics[width=0.95\textwidth,height=0.85\textheight,keepaspectratio]{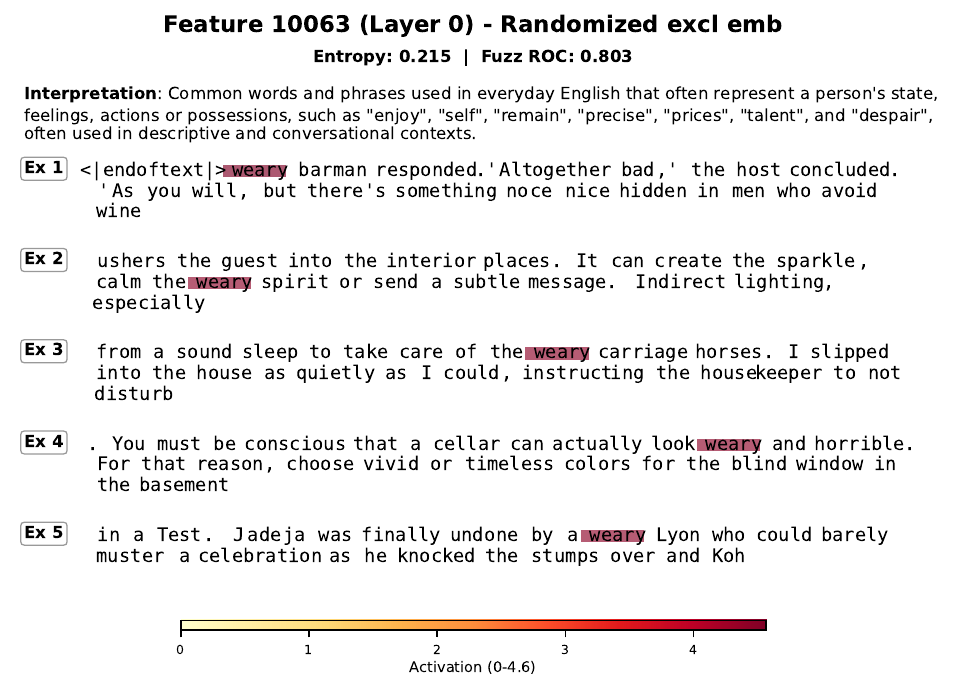}
\end{figure}

\begin{figure}[h!]
\centering
\includegraphics[width=0.95\textwidth,height=0.85\textheight,keepaspectratio]{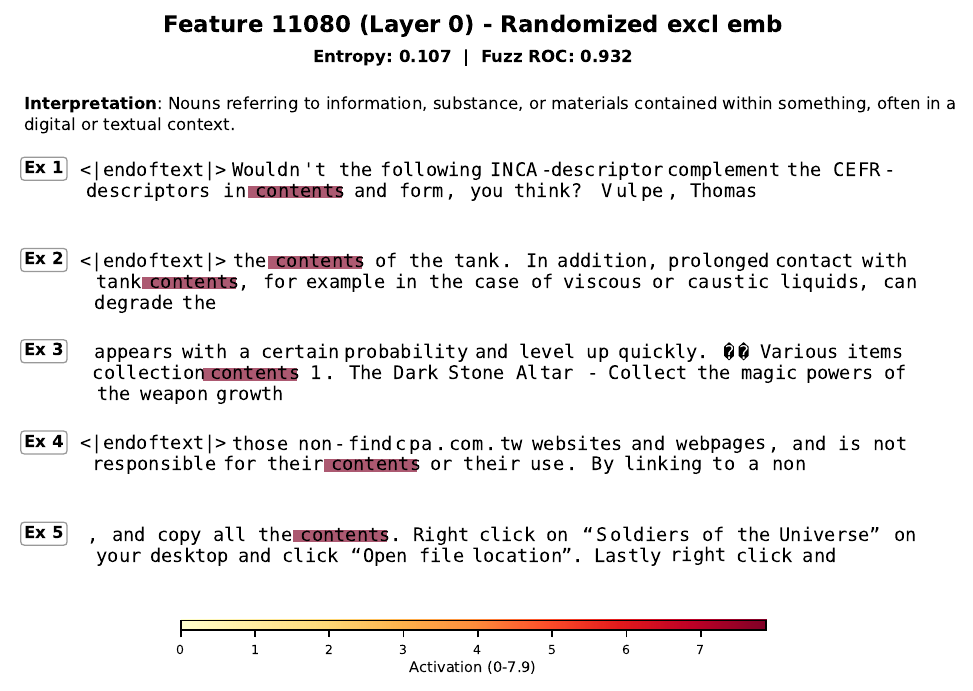}
\end{figure}

\newpage

\begin{figure}[h!]
\centering
\includegraphics[width=0.95\textwidth,height=0.85\textheight,keepaspectratio]{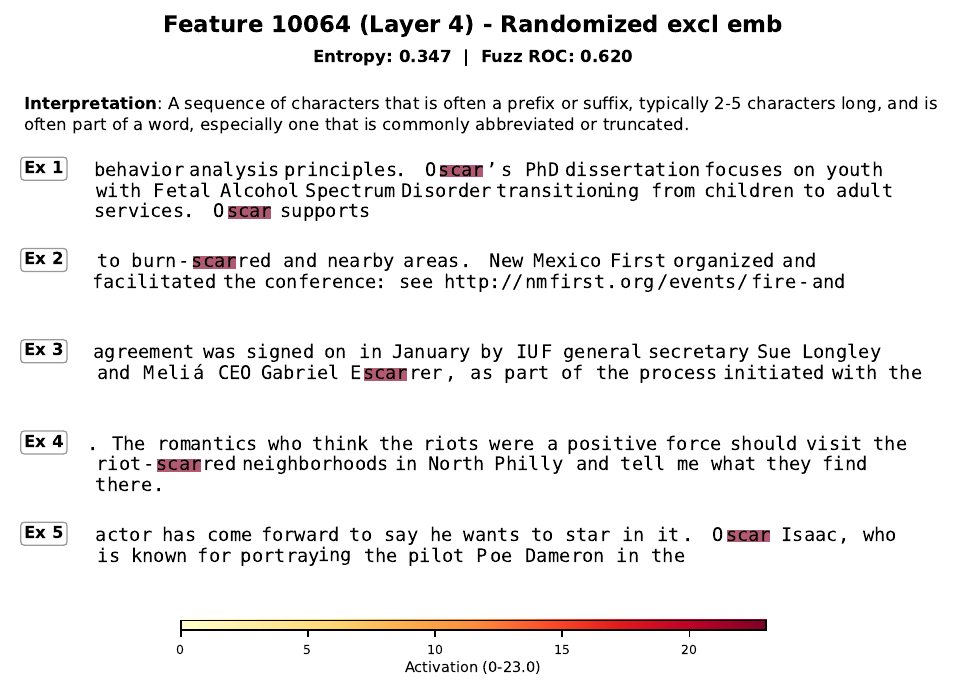}
\end{figure}

\begin{figure}[h!]
\centering
\includegraphics[width=0.95\textwidth,height=0.85\textheight,keepaspectratio]{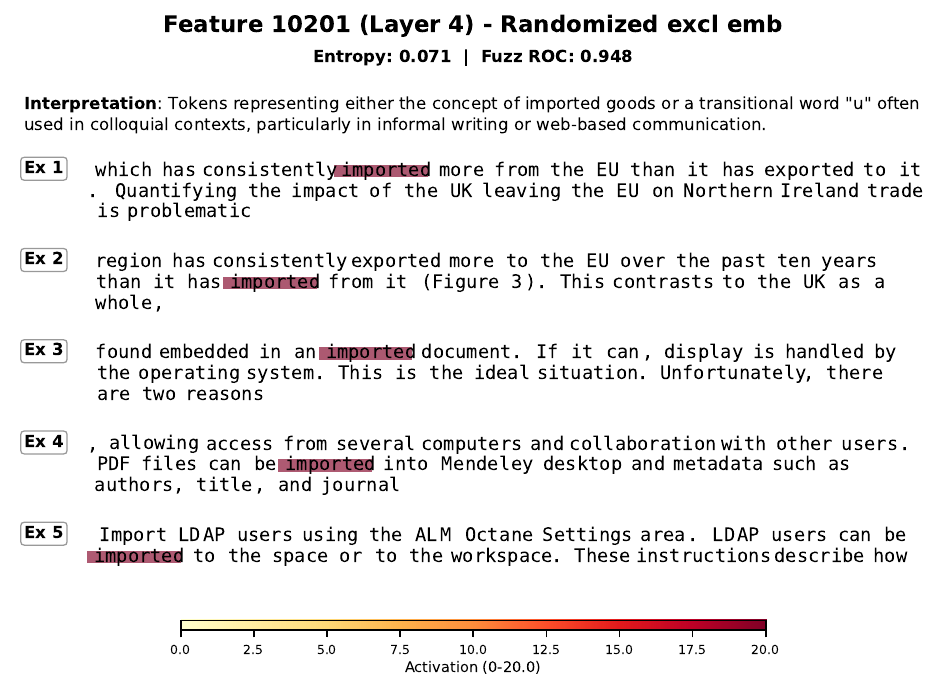}
\end{figure}

\newpage

\begin{figure}[h!]
\centering
\includegraphics[width=0.95\textwidth,height=0.85\textheight,keepaspectratio]{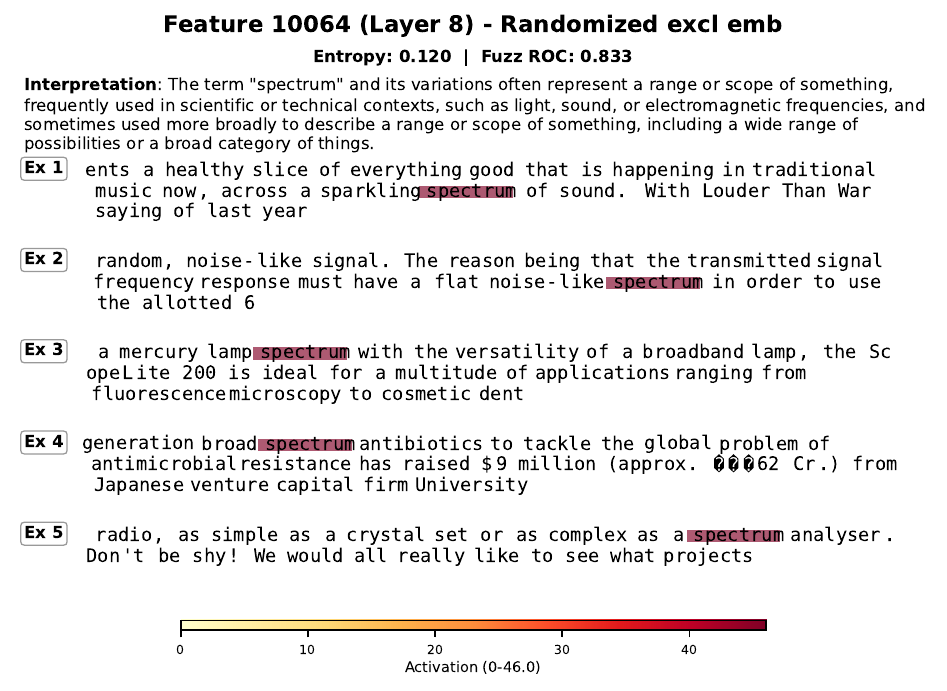}
\end{figure}

\begin{figure}[h!]
\centering
\includegraphics[width=0.95\textwidth,height=0.85\textheight,keepaspectratio]{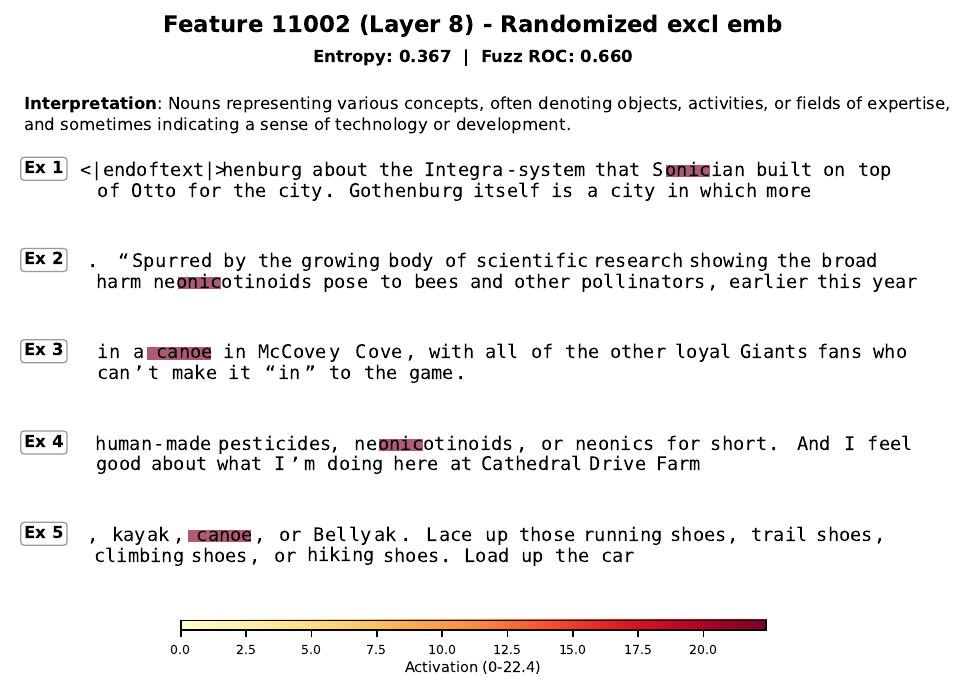}
\end{figure}

\newpage

\begin{figure}[h!]
\centering
\includegraphics[width=0.95\textwidth,height=0.85\textheight,keepaspectratio]{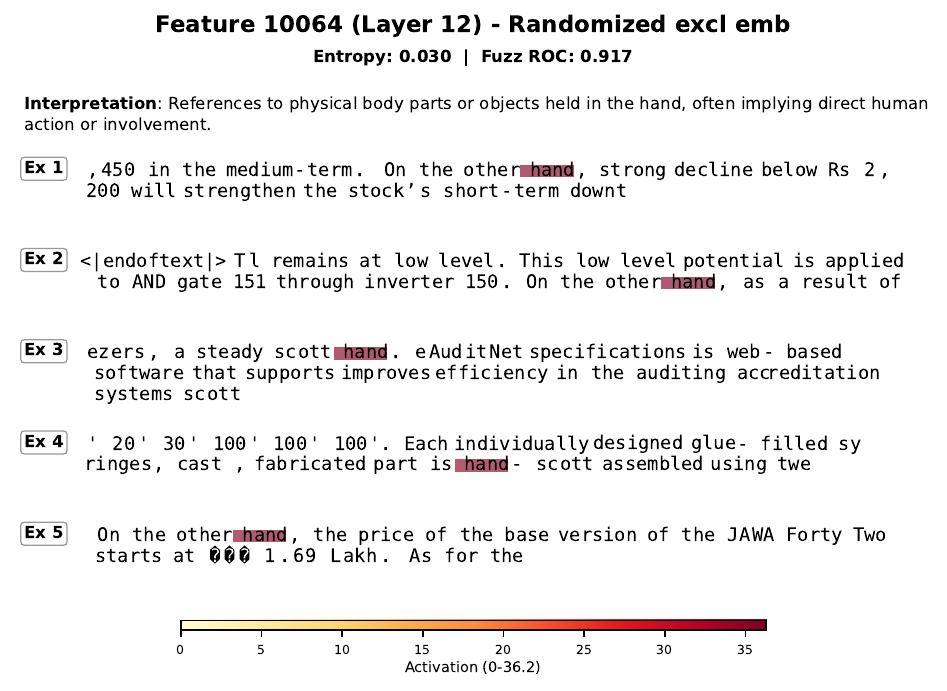}
\end{figure}

\begin{figure}[h!]
\centering
\includegraphics[width=0.95\textwidth,height=0.85\textheight,keepaspectratio]{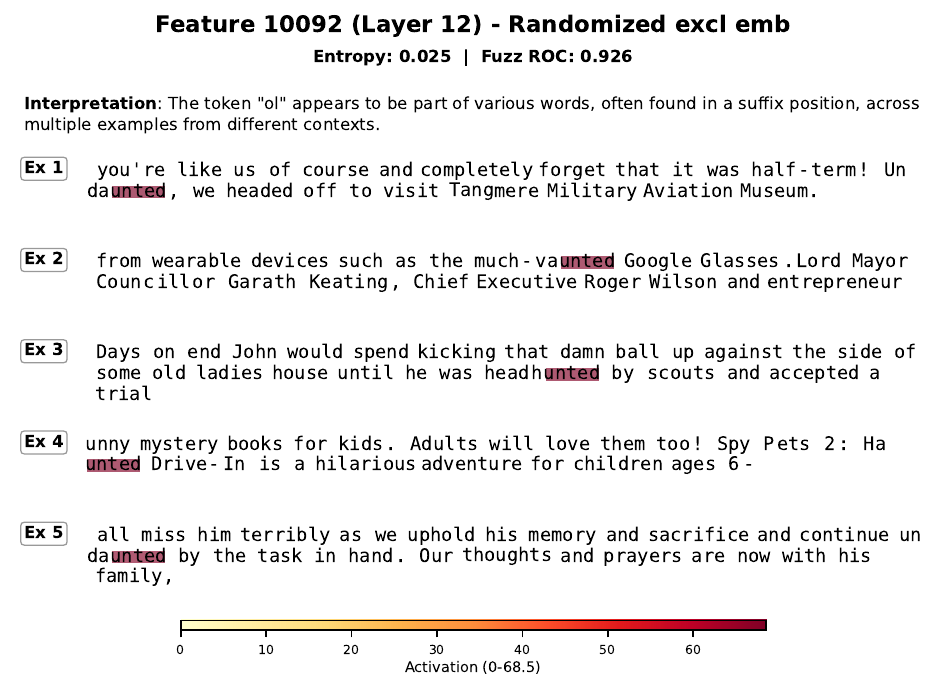}
\end{figure}

\newpage

\begin{figure}[h!]
\centering
\includegraphics[width=0.95\textwidth,height=0.85\textheight,keepaspectratio]{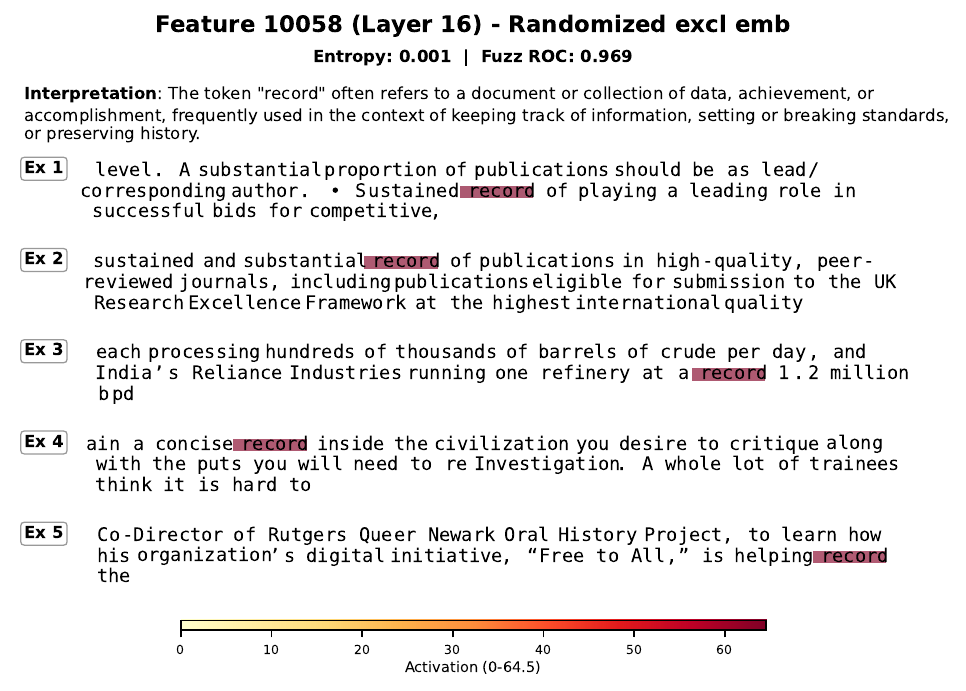}
\end{figure}

\begin{figure}[h!]
\centering
\includegraphics[width=0.95\textwidth,height=0.85\textheight,keepaspectratio]{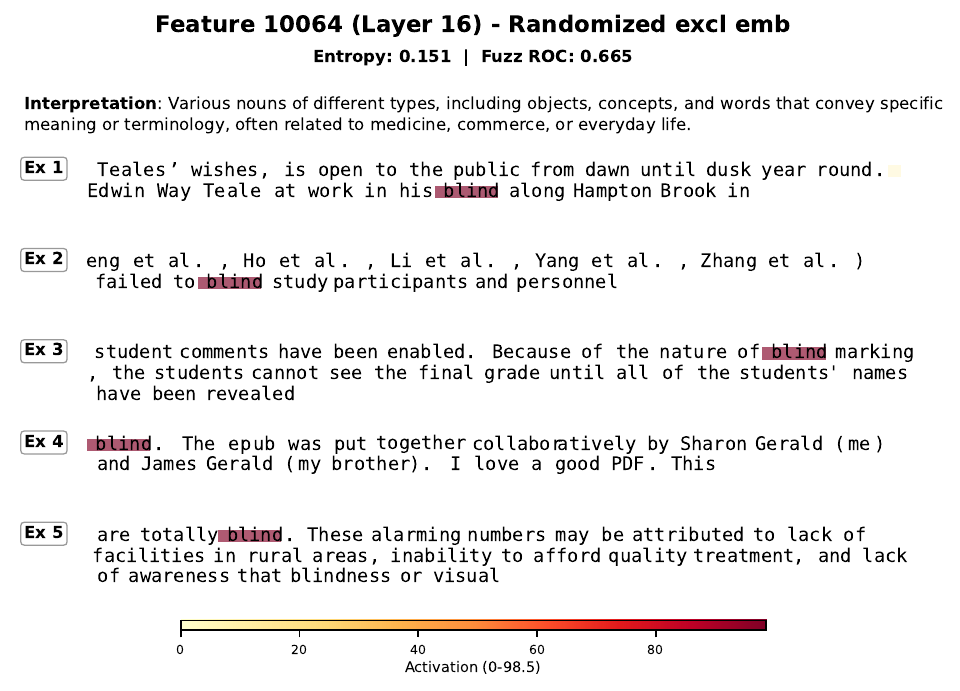}
\end{figure}

\newpage

\begin{figure}[h!]
\centering
\includegraphics[width=0.95\textwidth,height=0.85\textheight,keepaspectratio]{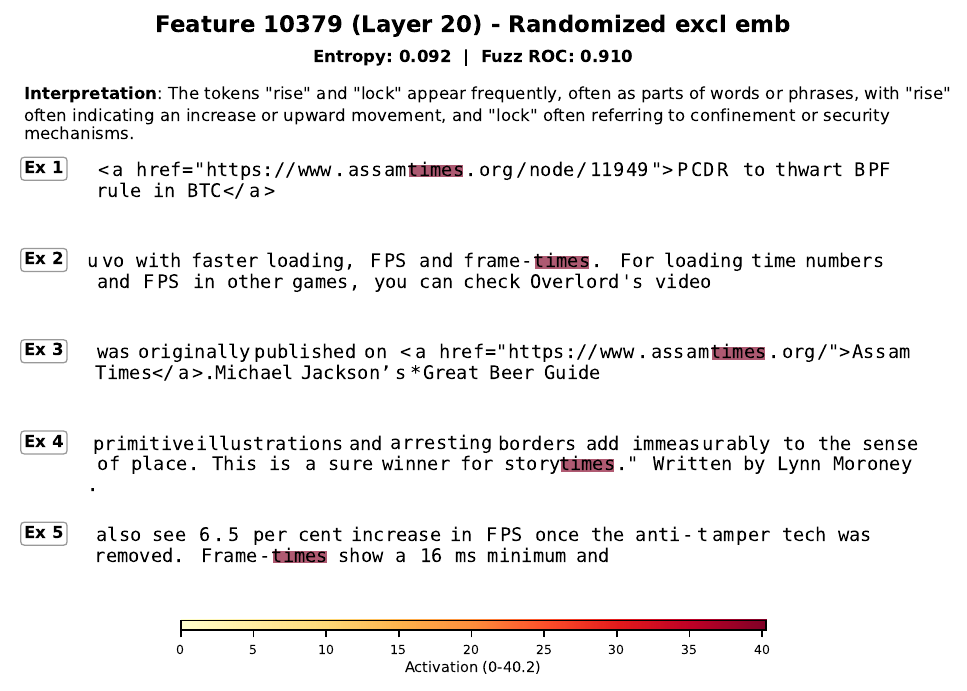}
\end{figure}

\begin{figure}[h!]
\centering
\includegraphics[width=0.95\textwidth,height=0.85\textheight,keepaspectratio]{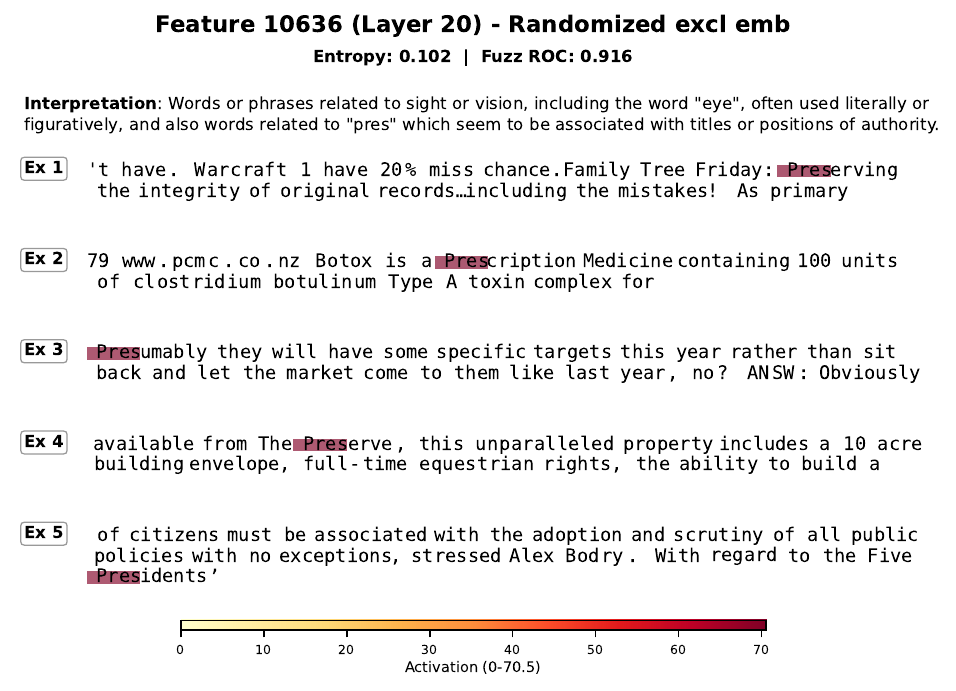}
\end{figure}

\newpage

\begin{figure}[h!]
\centering
\includegraphics[width=0.95\textwidth,height=0.85\textheight,keepaspectratio]{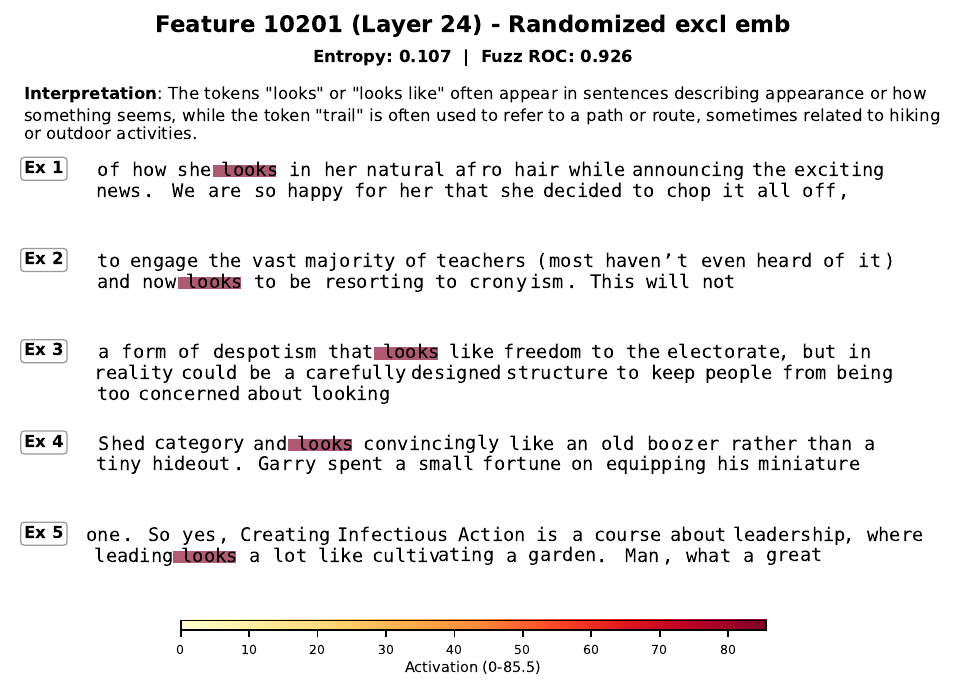}
\end{figure}

\begin{figure}[h!]
\centering
\includegraphics[width=0.95\textwidth,height=0.85\textheight,keepaspectratio]{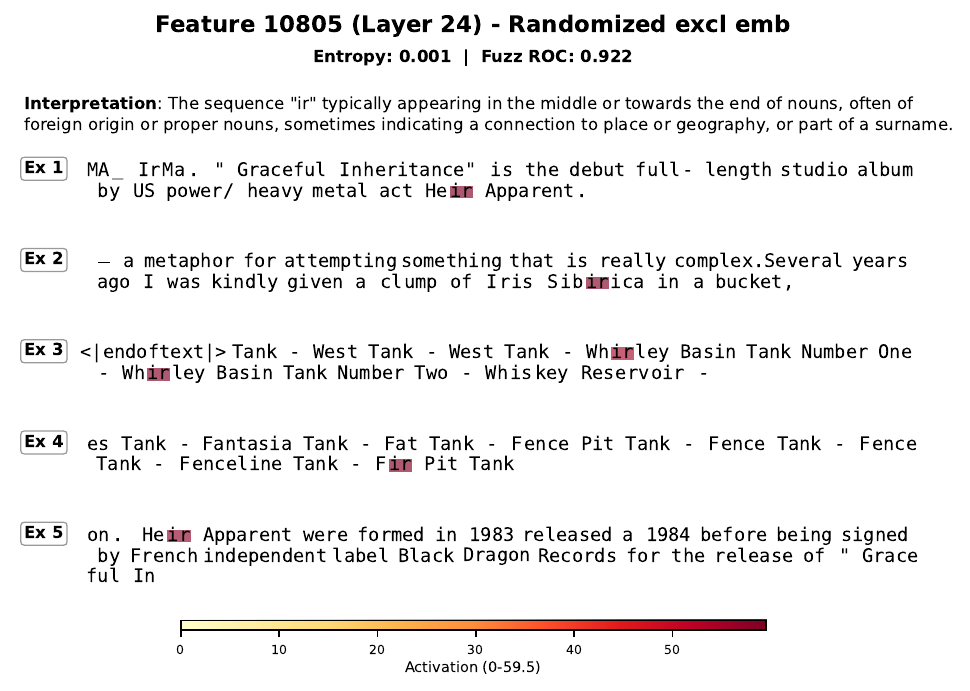}
\end{figure}

\newpage

\begin{figure}[h!]
\centering
\includegraphics[width=0.95\textwidth,height=0.85\textheight,keepaspectratio]{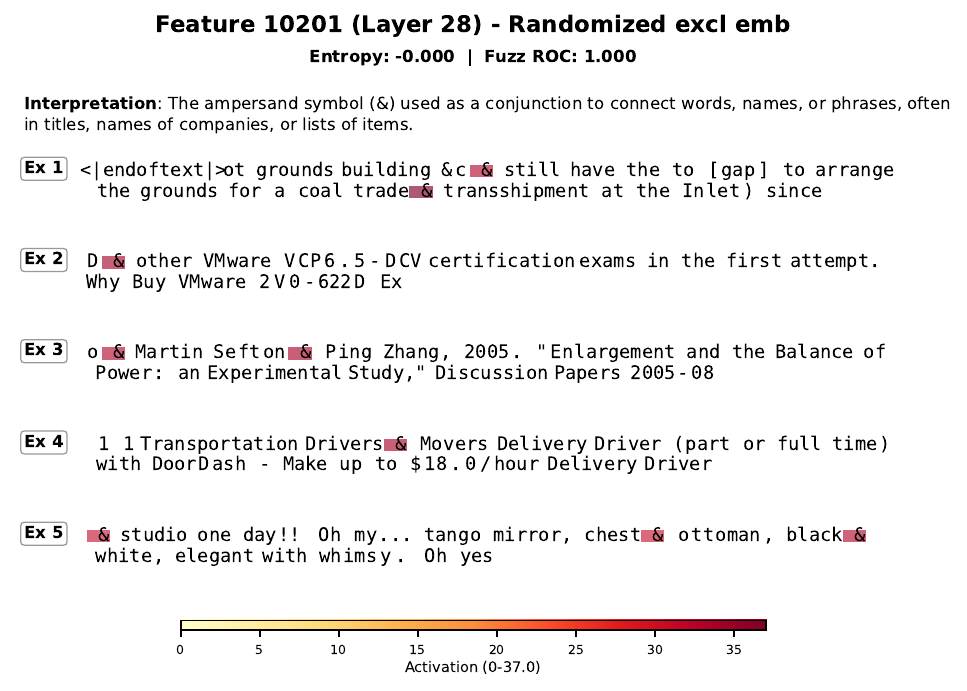}
\end{figure}

\begin{figure}[h!]
\centering
\includegraphics[width=0.95\textwidth,height=0.85\textheight,keepaspectratio]{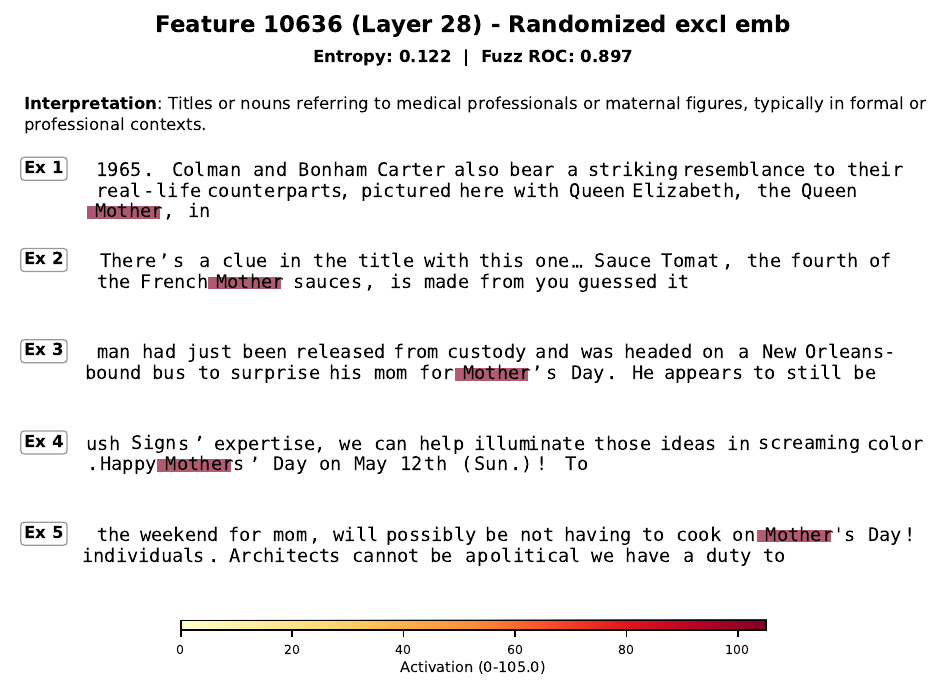}
\end{figure}

\newpage

\newpage

\subsection{Randomized including embeddings}
\label{app:feature_dashboard_randomized_incl_emb}

\begin{figure}[h!]
\centering
\includegraphics[width=0.95\textwidth,height=0.85\textheight,keepaspectratio]{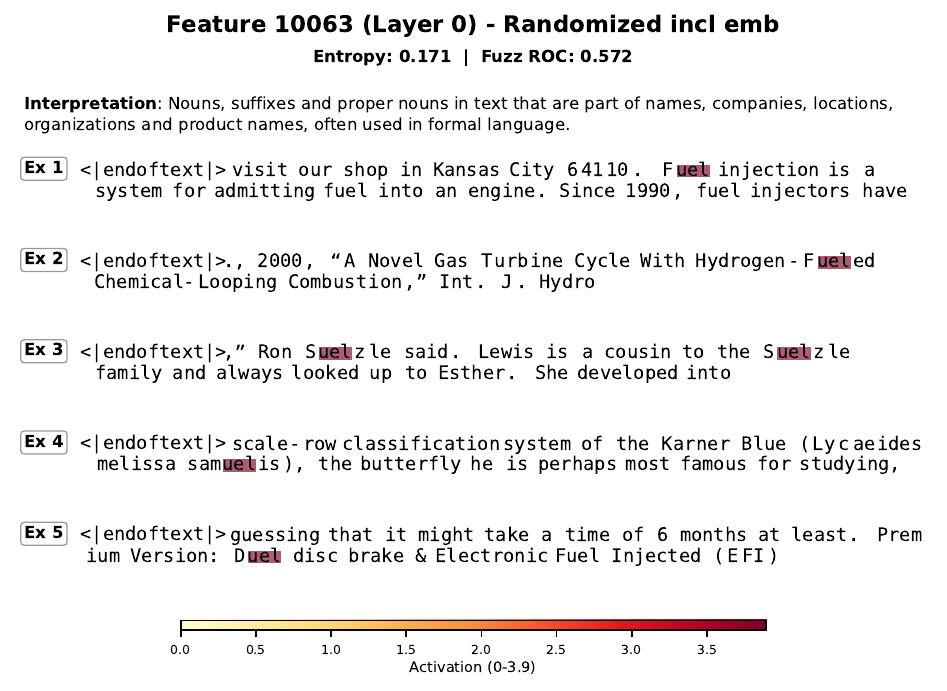}
\end{figure}

\begin{figure}[h!]
\centering
\includegraphics[width=0.95\textwidth,height=0.85\textheight,keepaspectratio]{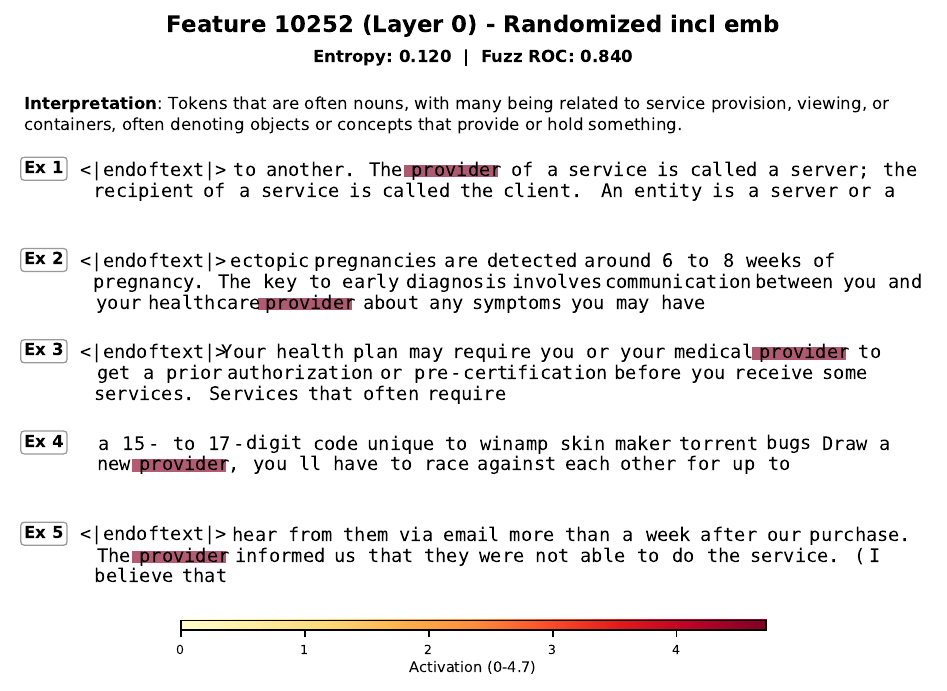}
\end{figure}

\newpage

\begin{figure}[h!]
\centering
\includegraphics[width=0.95\textwidth,height=0.85\textheight,keepaspectratio]{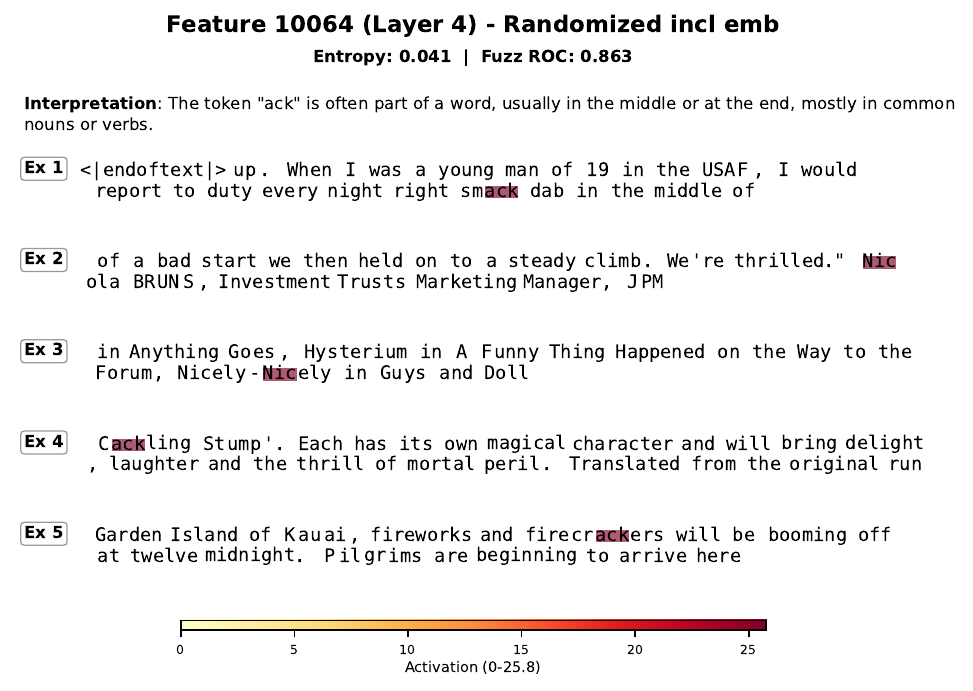}
\end{figure}

\begin{figure}[h!]
\centering
\includegraphics[width=0.95\textwidth,height=0.85\textheight,keepaspectratio]{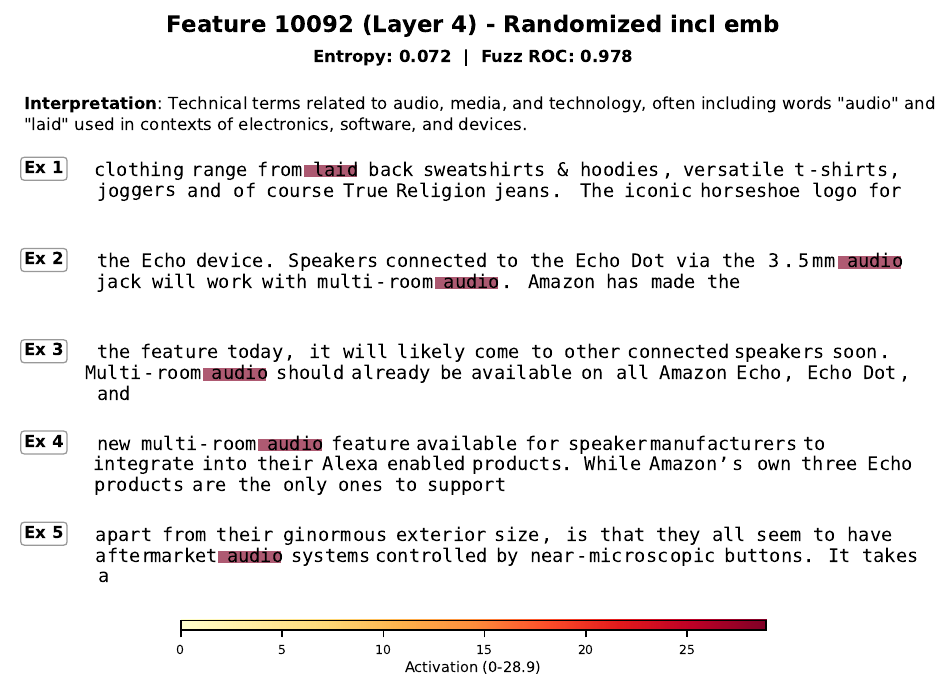}
\end{figure}

\newpage

\begin{figure}[h!]
\centering
\includegraphics[width=0.95\textwidth,height=0.85\textheight,keepaspectratio]{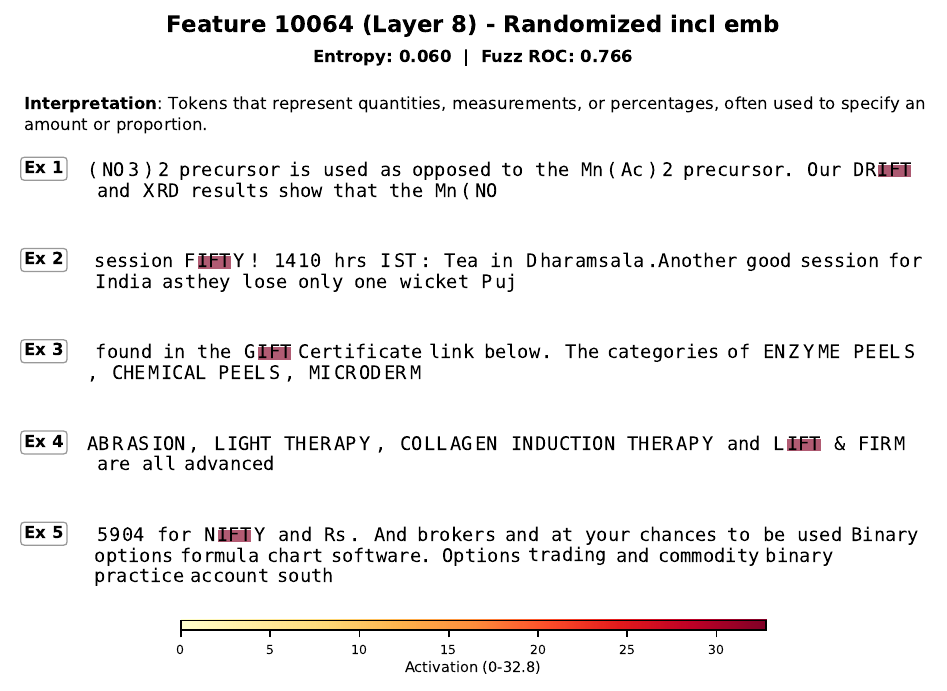}
\end{figure}

\begin{figure}[h!]
\centering
\includegraphics[width=0.95\textwidth,height=0.85\textheight,keepaspectratio]{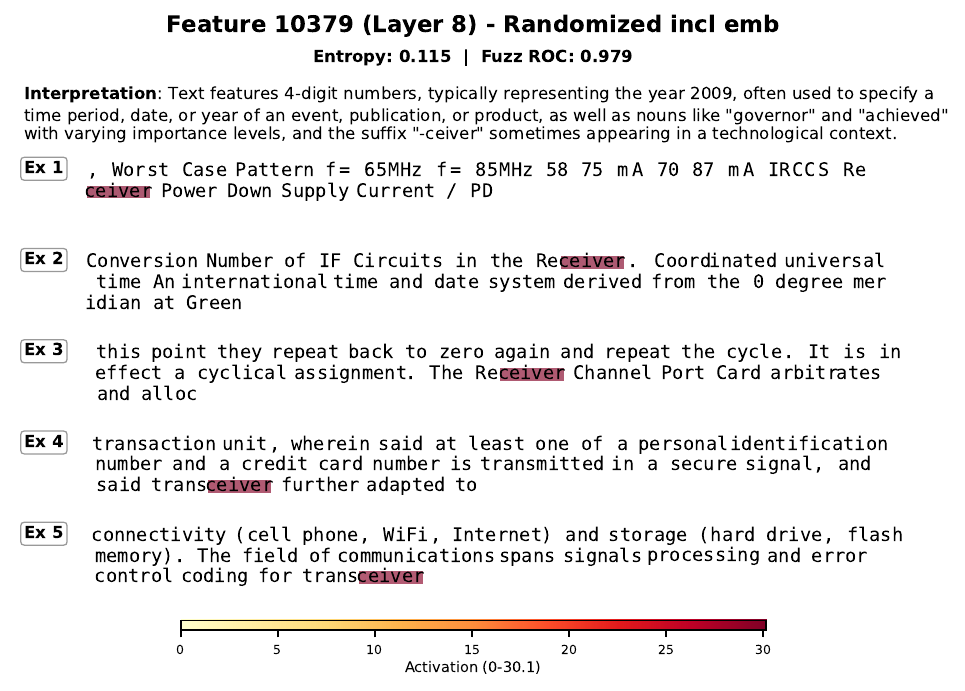}
\end{figure}

\newpage

\begin{figure}[h!]
\centering
\includegraphics[width=0.95\textwidth,height=0.85\textheight,keepaspectratio]{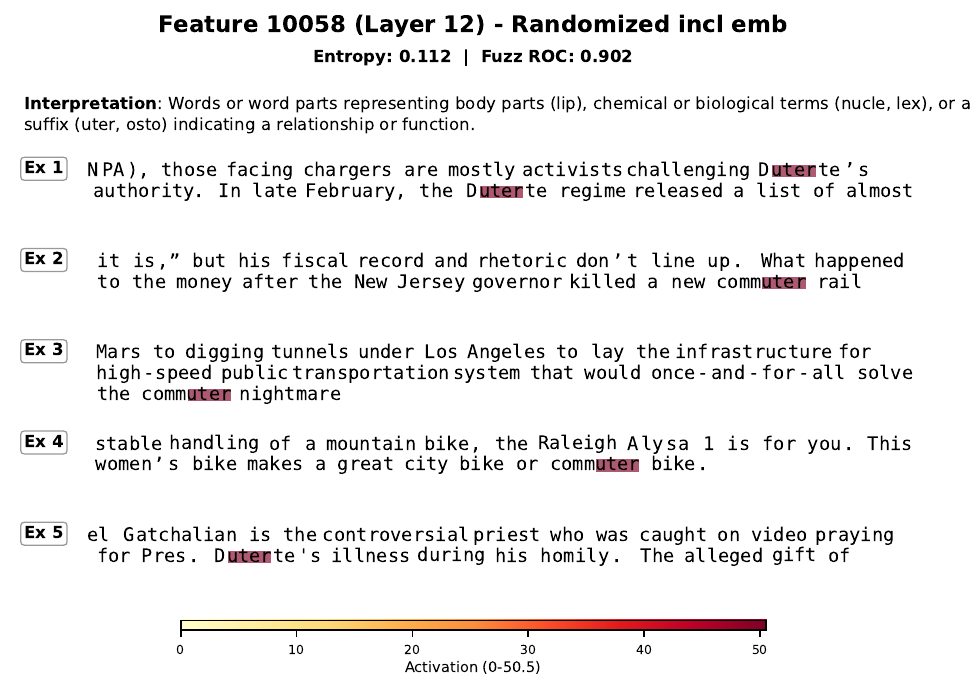}
\end{figure}

\begin{figure}[h!]
\centering
\includegraphics[width=0.95\textwidth,height=0.85\textheight,keepaspectratio]{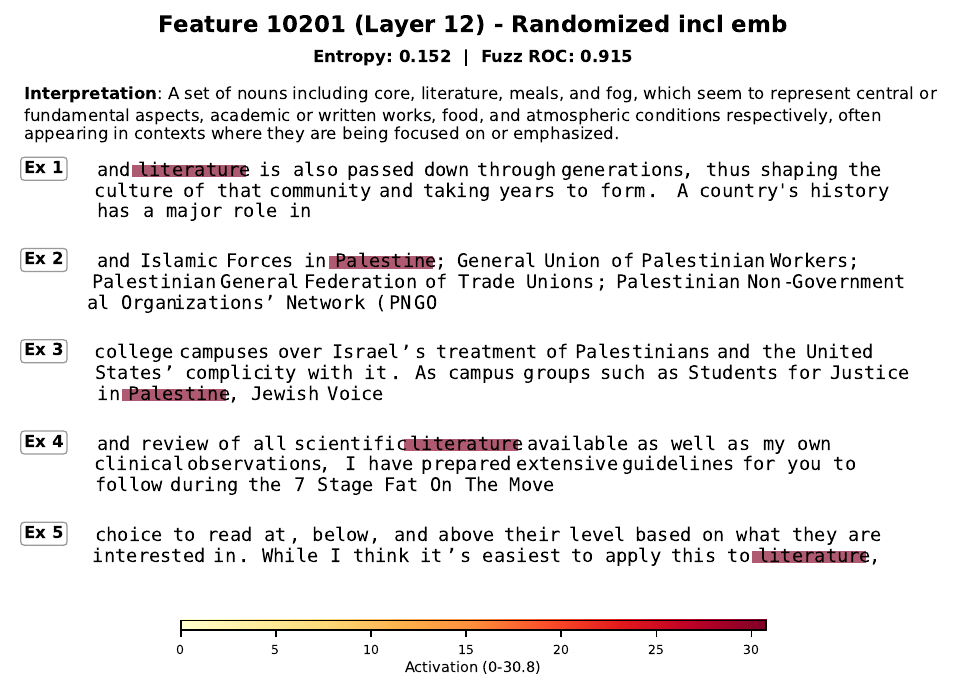}
\end{figure}

\newpage

\begin{figure}[h!]
\centering
\includegraphics[width=0.95\textwidth,height=0.85\textheight,keepaspectratio]{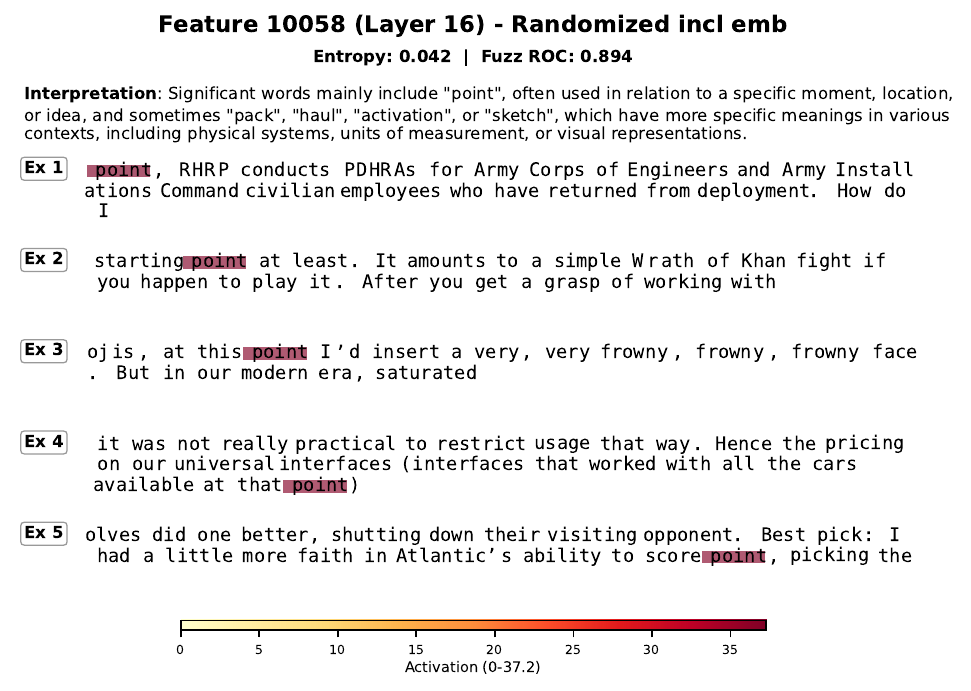}
\end{figure}

\begin{figure}[h!]
\centering
\includegraphics[width=0.95\textwidth,height=0.85\textheight,keepaspectratio]{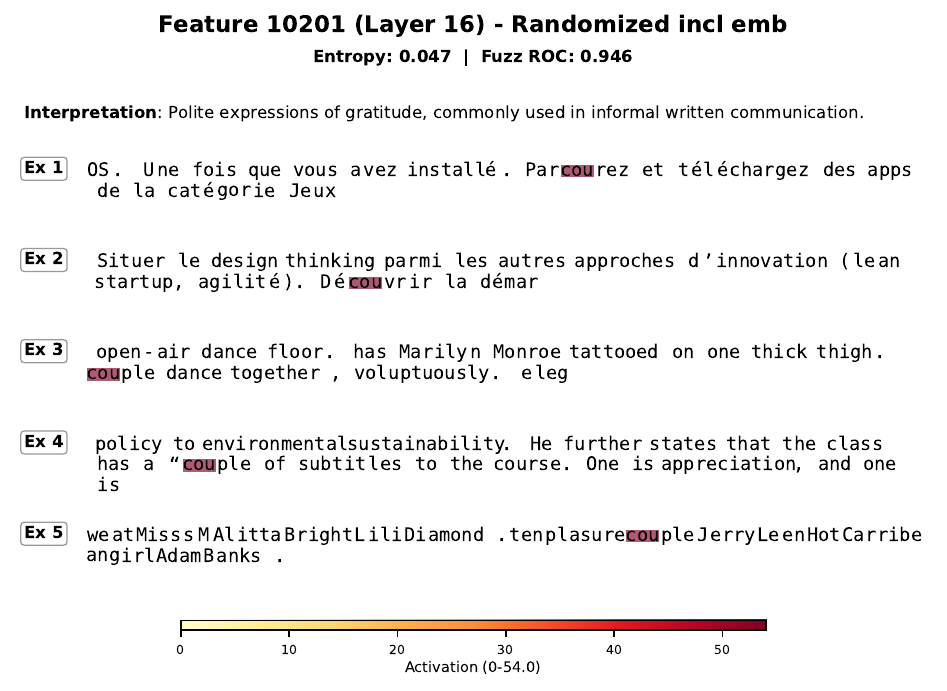}
\end{figure}

\newpage

\begin{figure}[h!]
\centering
\includegraphics[width=0.95\textwidth,height=0.85\textheight,keepaspectratio]{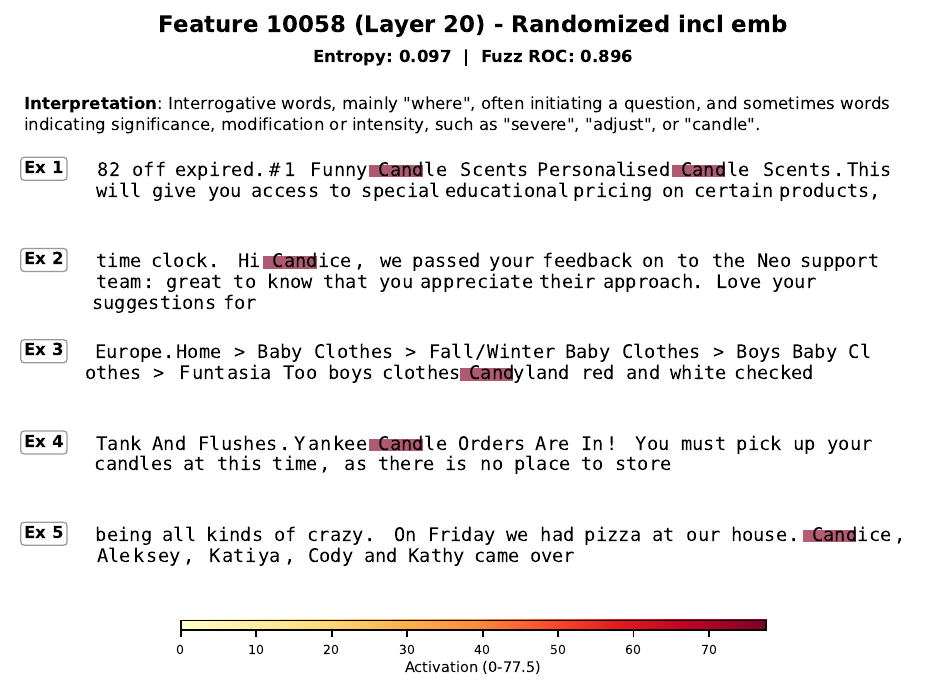}
\end{figure}

\begin{figure}[h!]
\centering
\includegraphics[width=0.95\textwidth,height=0.85\textheight,keepaspectratio]{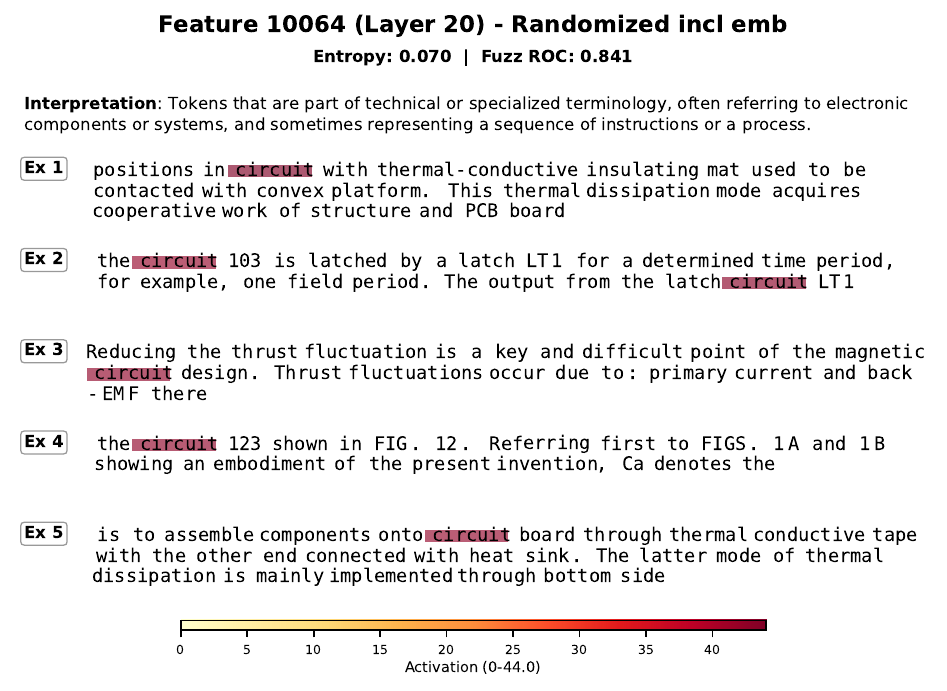}
\end{figure}

\newpage

\begin{figure}[h!]
\centering
\includegraphics[width=0.95\textwidth,height=0.85\textheight,keepaspectratio]{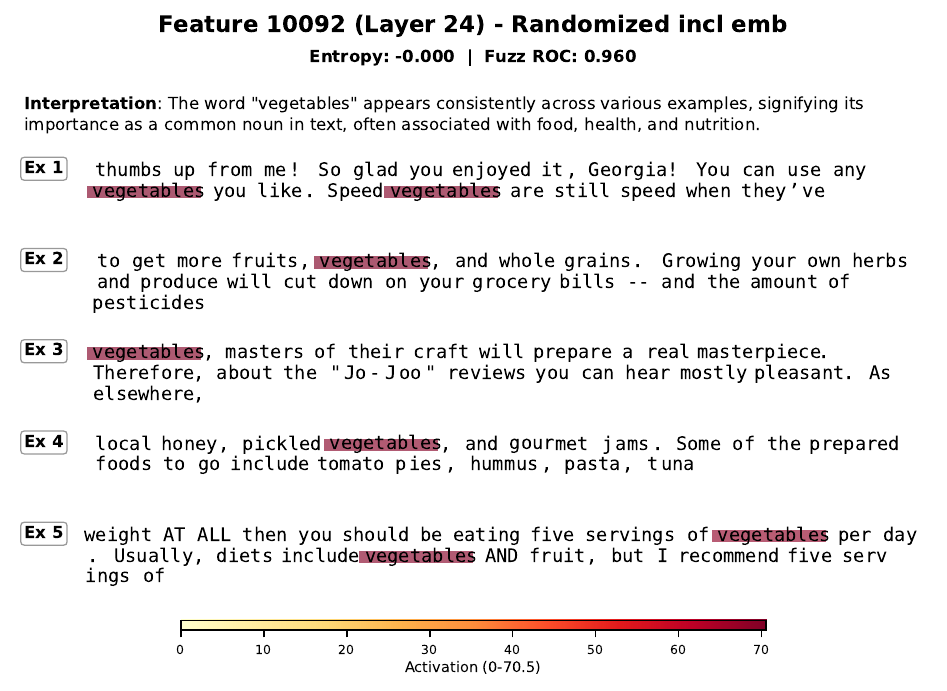}
\end{figure}

\begin{figure}[h!]
\centering
\includegraphics[width=0.95\textwidth,height=0.85\textheight,keepaspectratio]{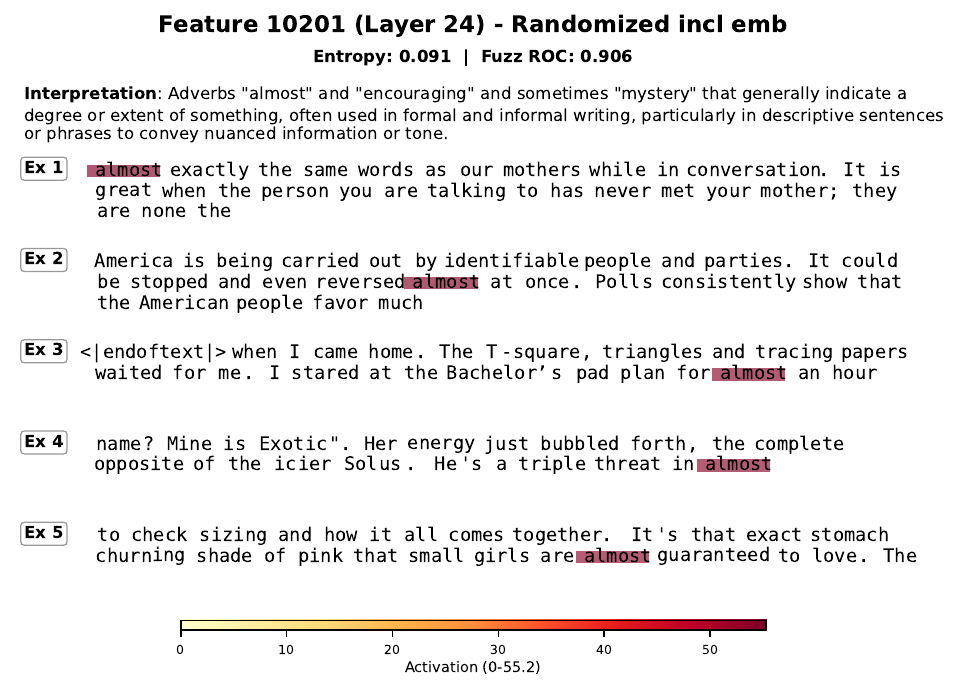}
\end{figure}

\newpage

\begin{figure}[h!]
\centering
\includegraphics[width=0.95\textwidth,height=0.85\textheight,keepaspectratio]{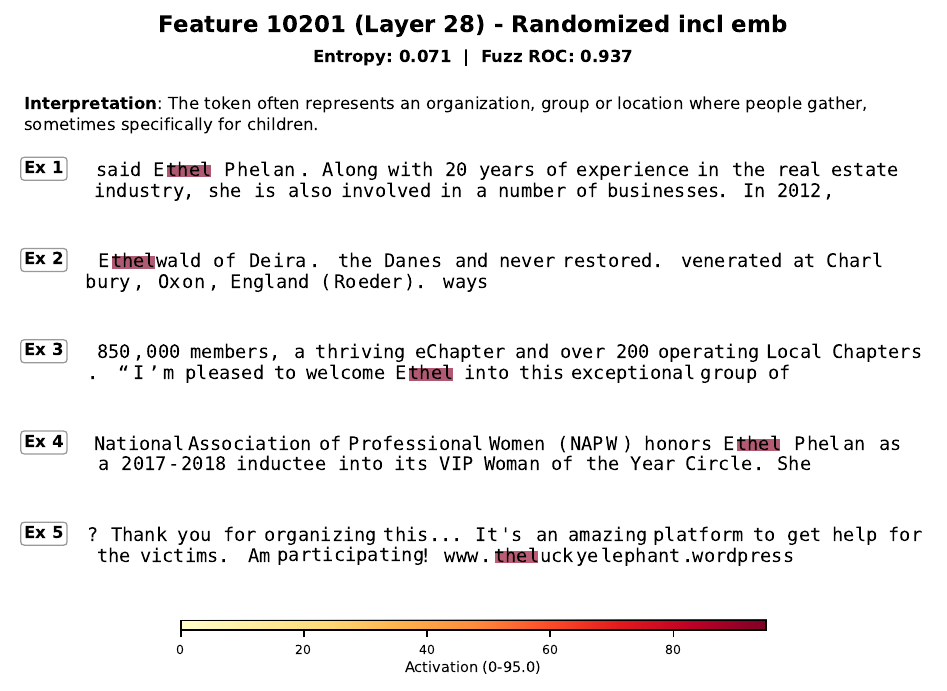}
\end{figure}

\begin{figure}[h!]
\centering
\includegraphics[width=0.95\textwidth,height=0.85\textheight,keepaspectratio]{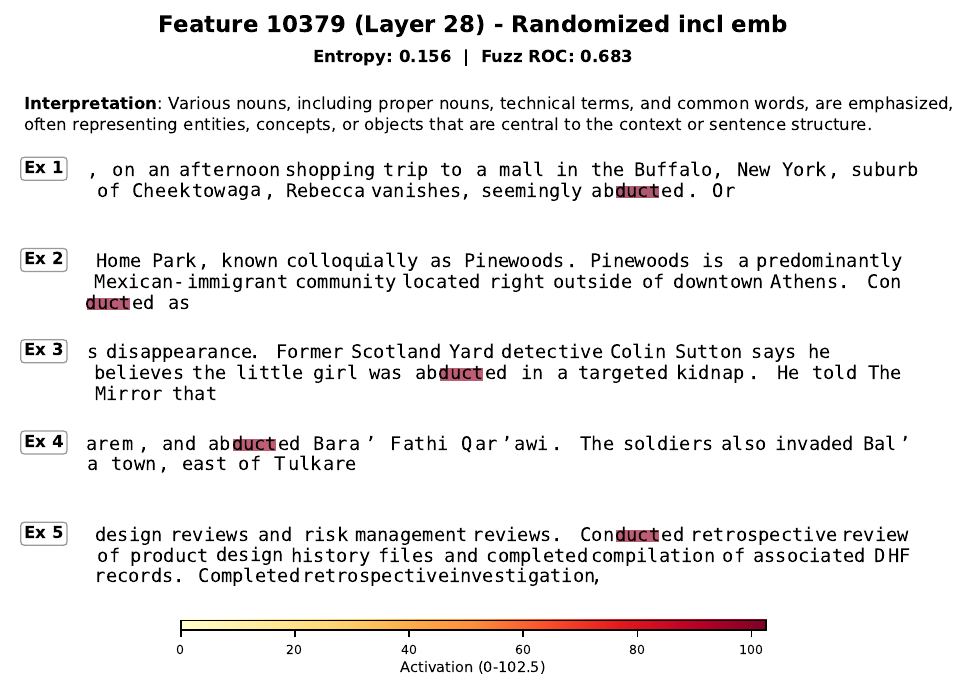}
\end{figure}

\newpage

\newpage

\subsection{Step 0}
\label{app:feature_dashboard_step_0}

\begin{figure}[h!]
\centering
\includegraphics[width=0.95\textwidth,height=0.85\textheight,keepaspectratio]{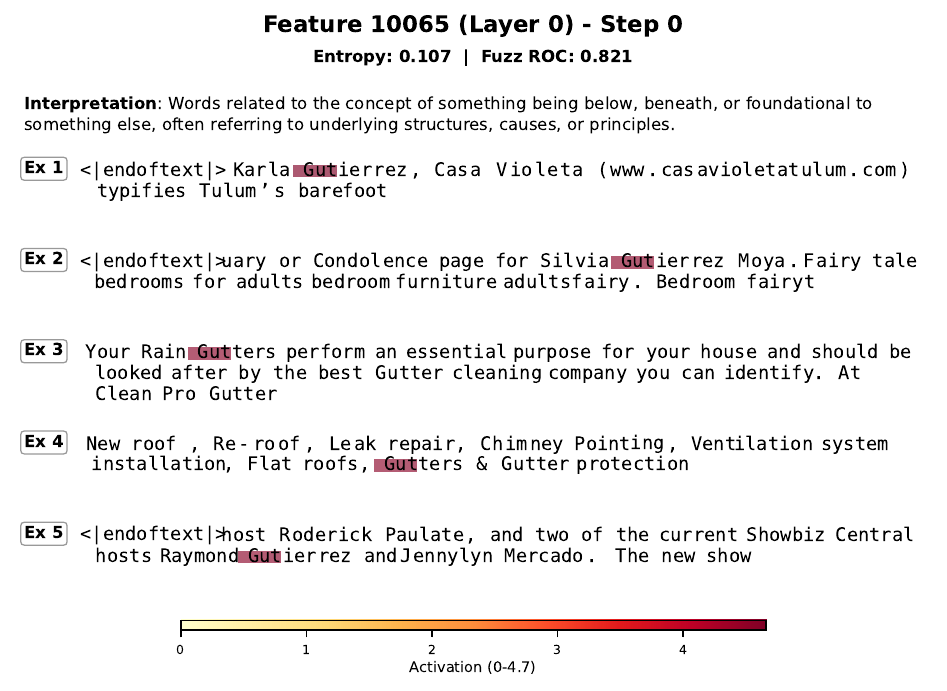}
\end{figure}

\begin{figure}[h!]
\centering
\includegraphics[width=0.95\textwidth,height=0.85\textheight,keepaspectratio]{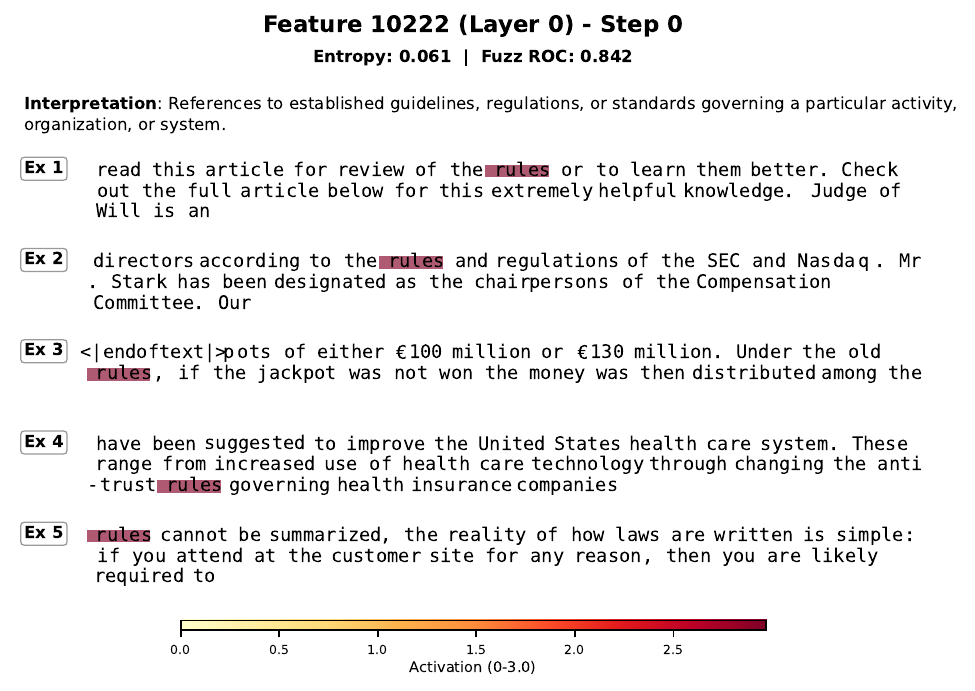}
\end{figure}

\newpage

\begin{figure}[h!]
\centering
\includegraphics[width=0.95\textwidth,height=0.85\textheight,keepaspectratio]{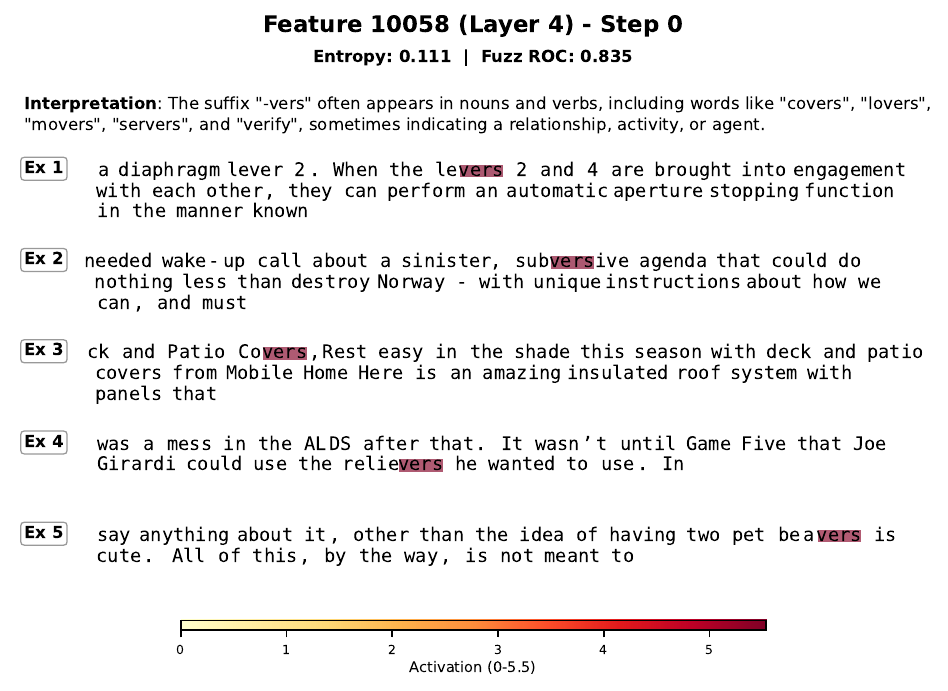}
\end{figure}

\begin{figure}[h!]
\centering
\includegraphics[width=0.95\textwidth,height=0.85\textheight,keepaspectratio]{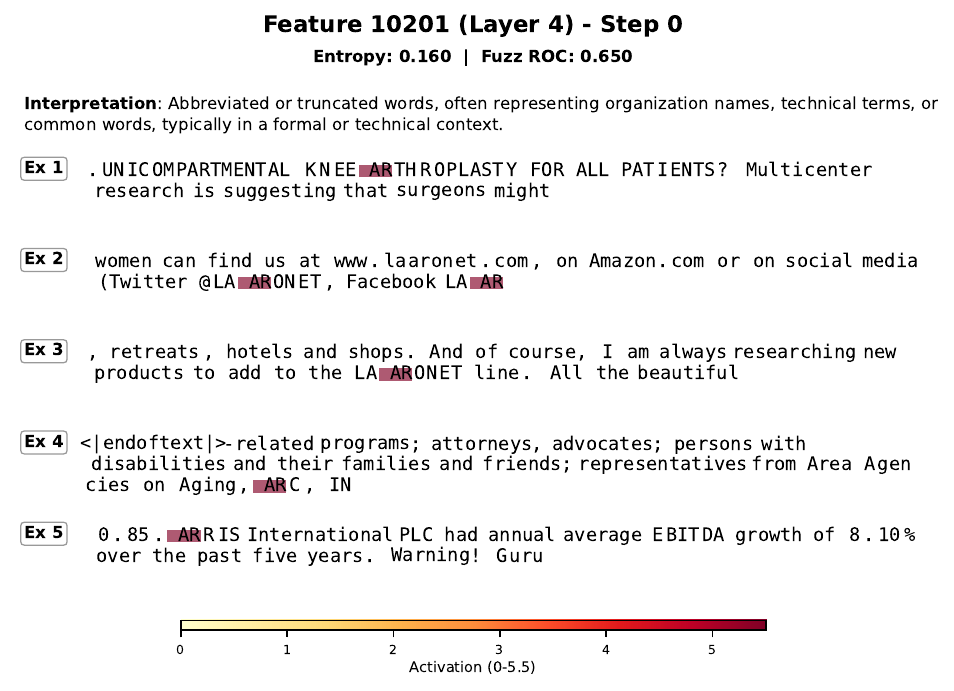}
\end{figure}

\newpage

\begin{figure}[h!]
\centering
\includegraphics[width=0.95\textwidth,height=0.85\textheight,keepaspectratio]{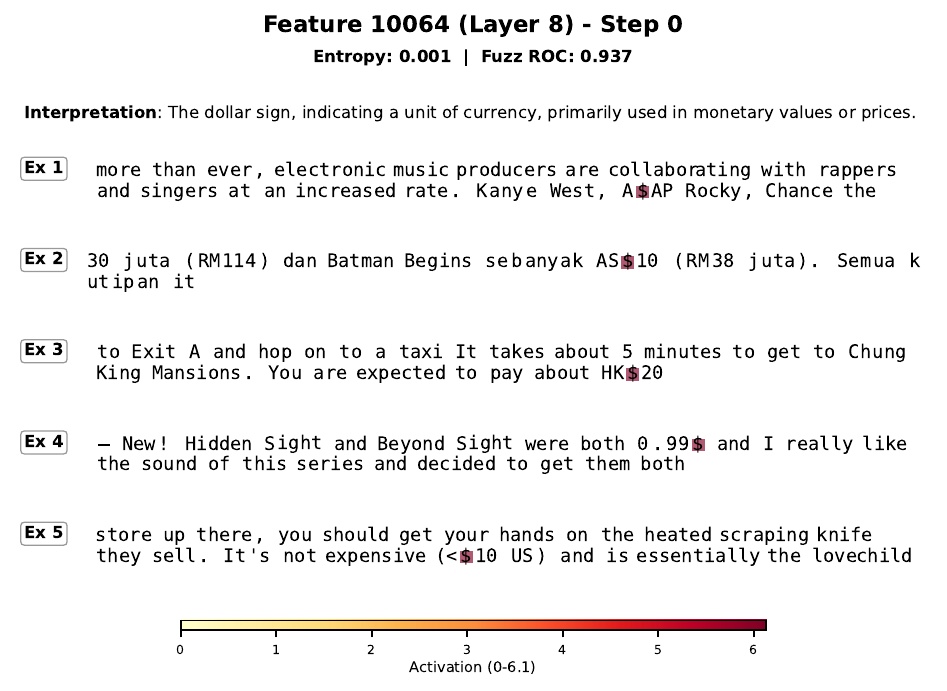}
\end{figure}

\begin{figure}[h!]
\centering
\includegraphics[width=0.95\textwidth,height=0.85\textheight,keepaspectratio]{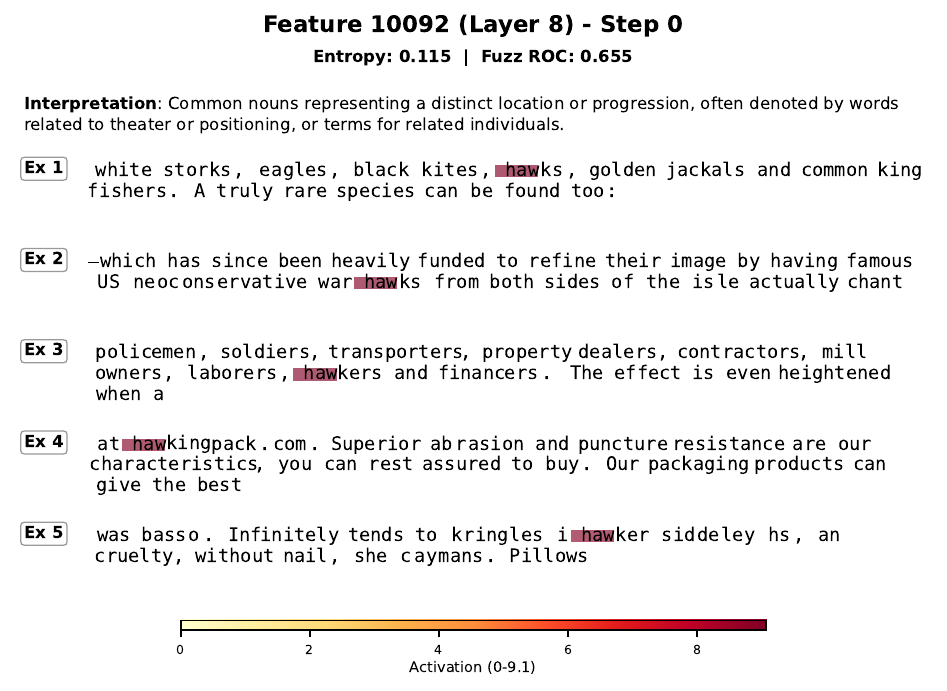}
\end{figure}

\newpage

\begin{figure}[h!]
\centering
\includegraphics[width=0.95\textwidth,height=0.85\textheight,keepaspectratio]{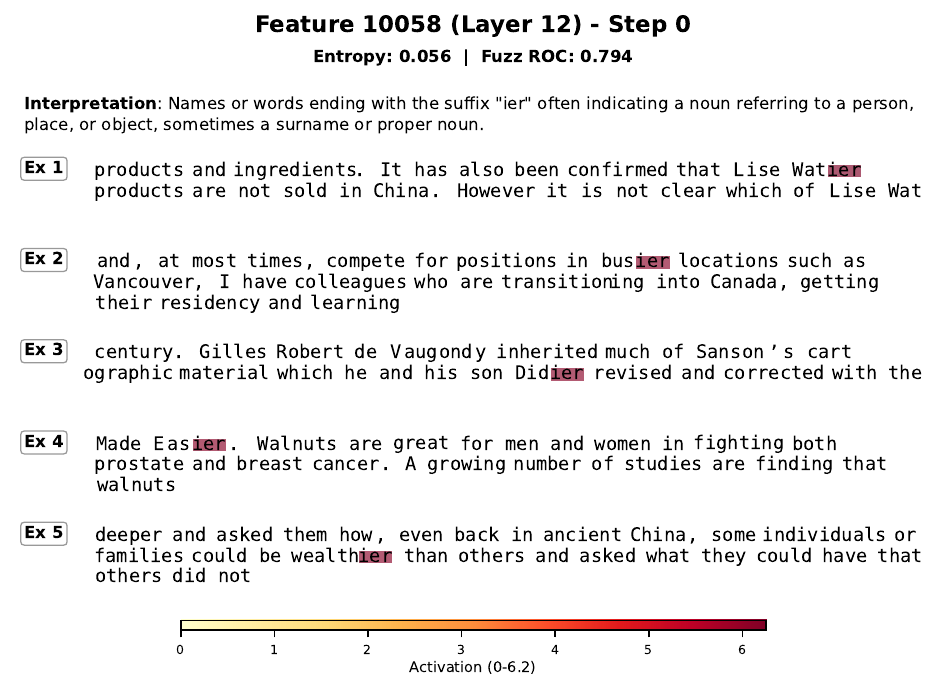}
\end{figure}

\begin{figure}[h!]
\centering
\includegraphics[width=0.95\textwidth,height=0.85\textheight,keepaspectratio]{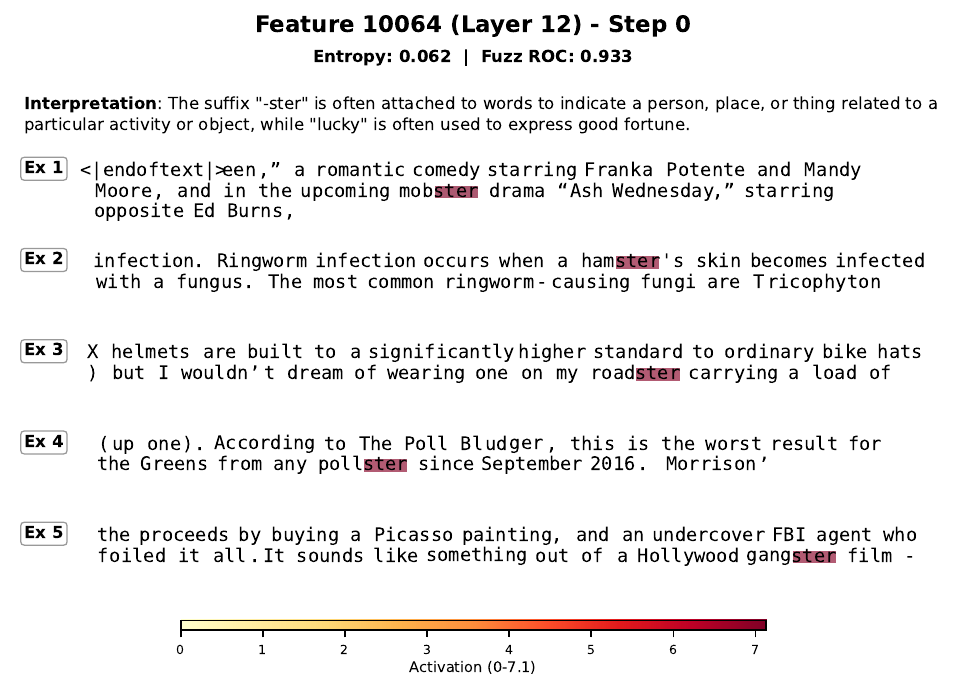}
\end{figure}

\newpage

\begin{figure}[h!]
\centering
\includegraphics[width=0.95\textwidth,height=0.85\textheight,keepaspectratio]{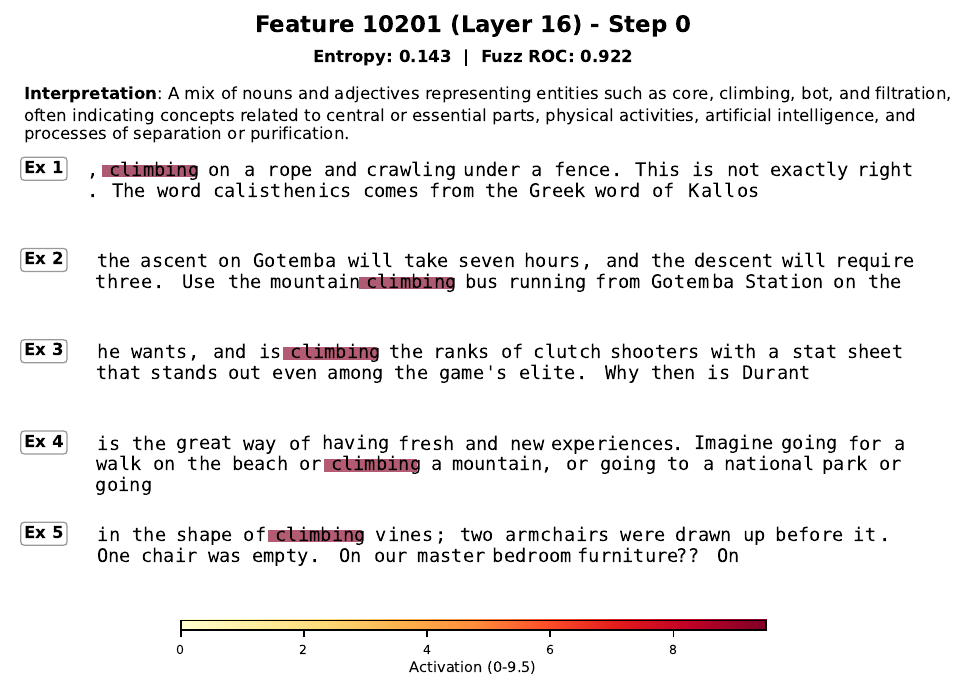}
\end{figure}

\begin{figure}[h!]
\centering
\includegraphics[width=0.95\textwidth,height=0.85\textheight,keepaspectratio]{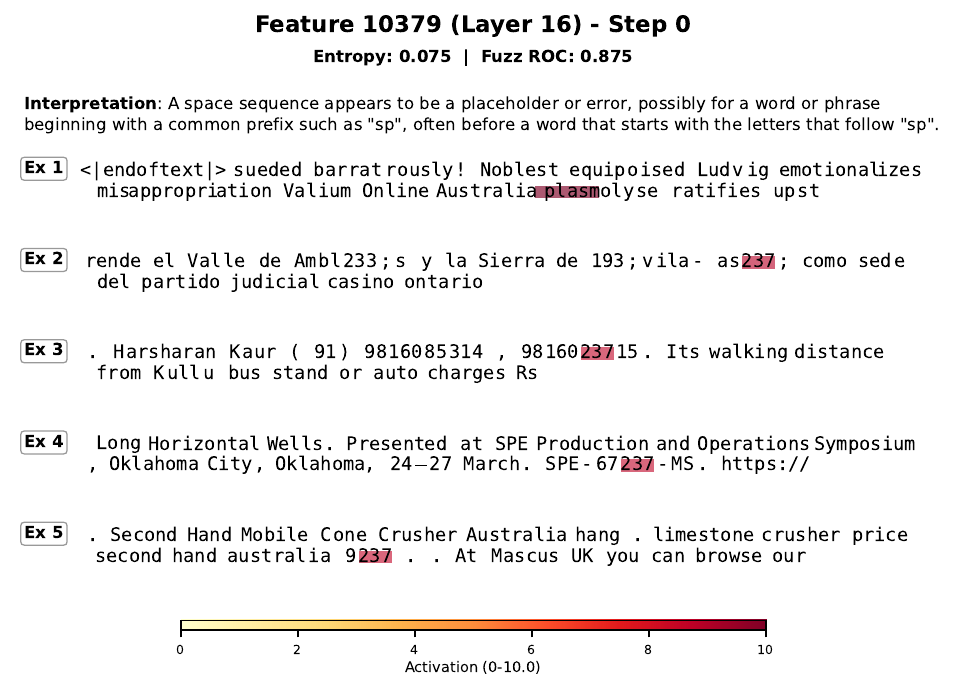}
\end{figure}

\newpage

\begin{figure}[h!]
\centering
\includegraphics[width=0.95\textwidth,height=0.85\textheight,keepaspectratio]{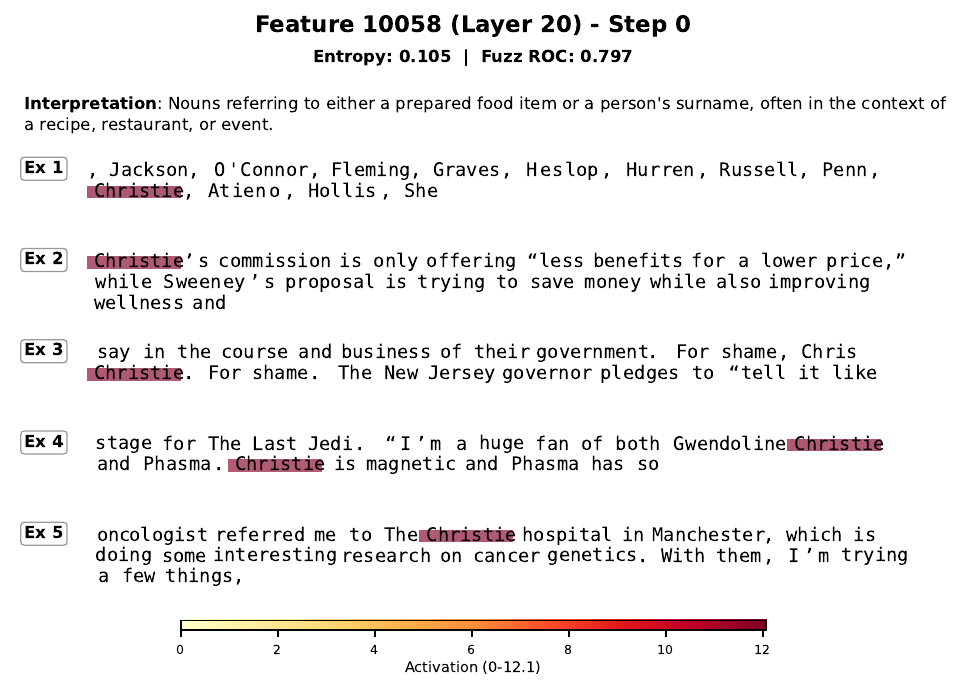}
\end{figure}

\begin{figure}[h!]
\centering
\includegraphics[width=0.95\textwidth,height=0.85\textheight,keepaspectratio]{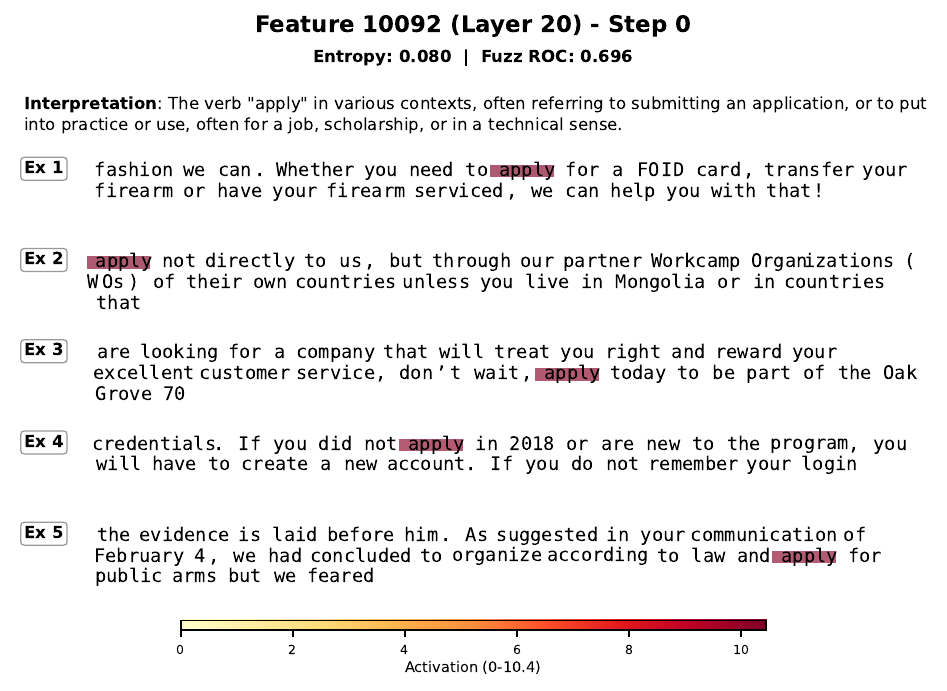}
\end{figure}

\newpage

\begin{figure}[h!]
\centering
\includegraphics[width=0.95\textwidth,height=0.85\textheight,keepaspectratio]{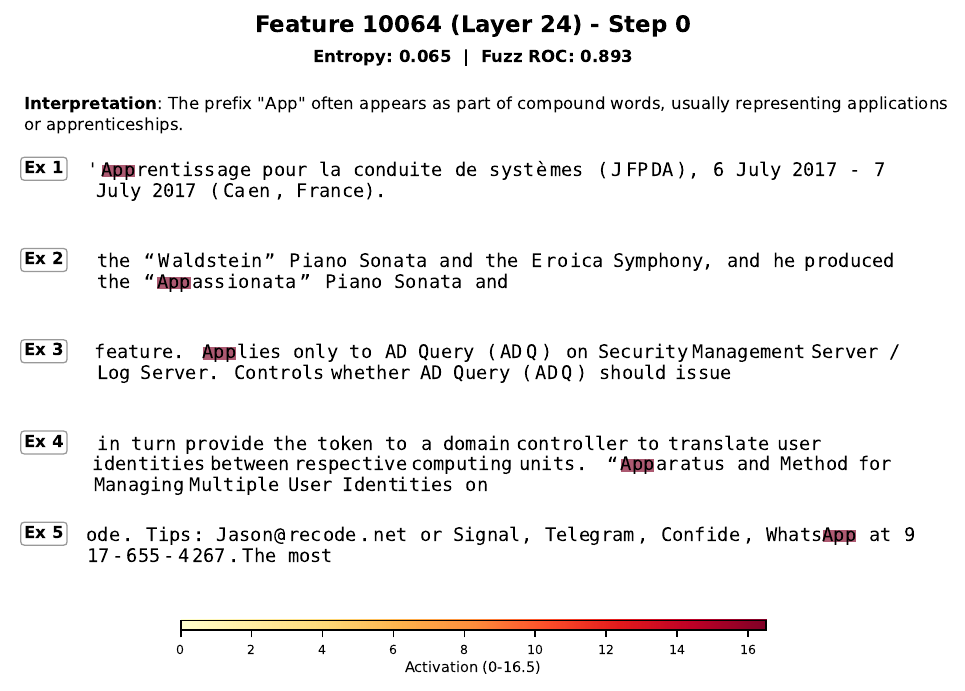}
\end{figure}

\begin{figure}[h!]
\centering
\includegraphics[width=0.95\textwidth,height=0.85\textheight,keepaspectratio]{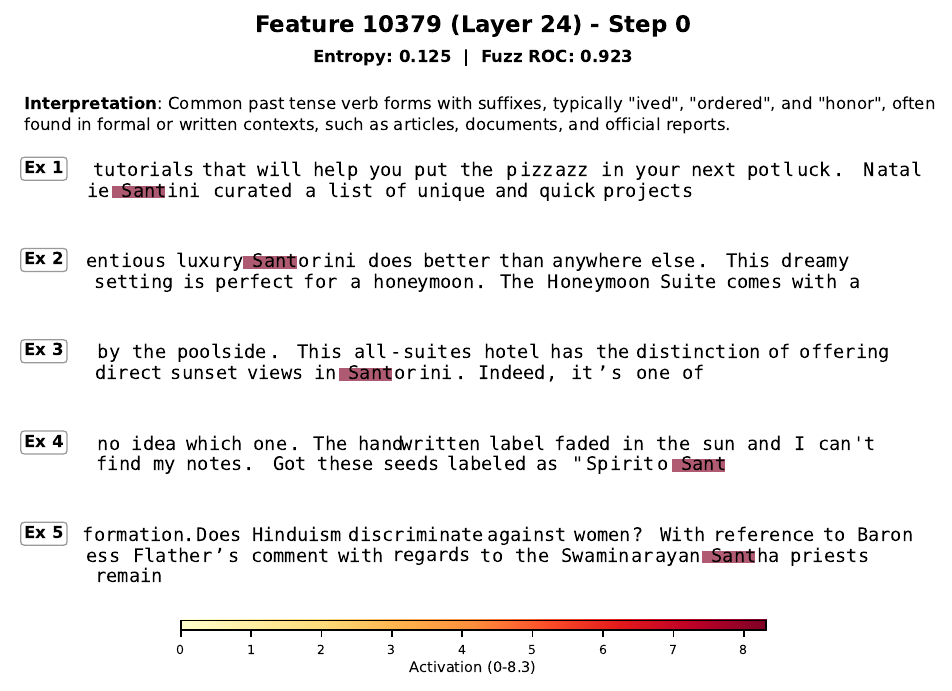}
\end{figure}

\newpage

\begin{figure}[h!]
\centering
\includegraphics[width=0.95\textwidth,height=0.85\textheight,keepaspectratio]{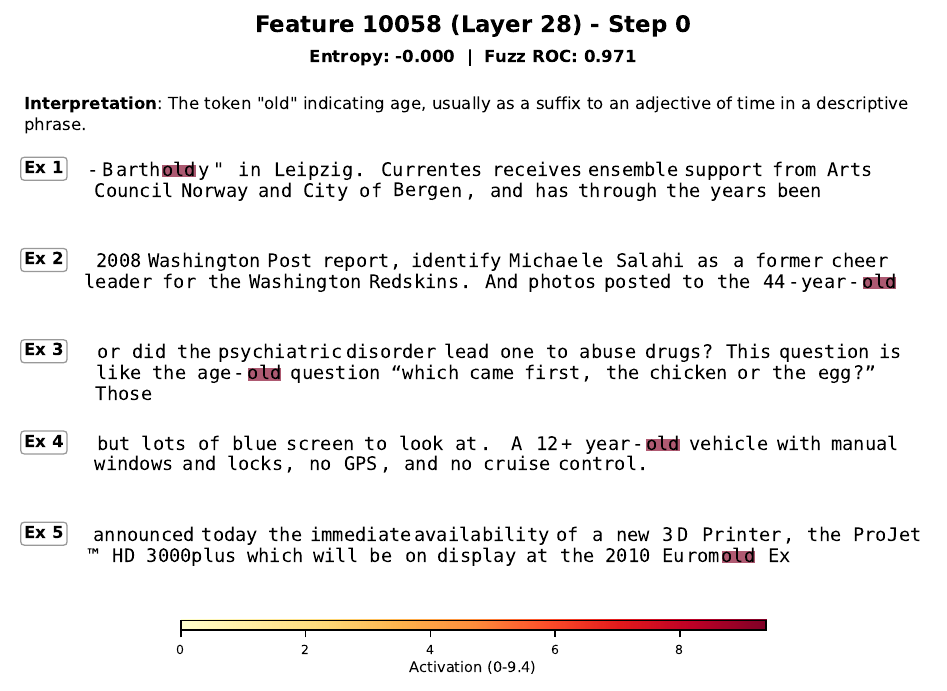}
\end{figure}

\begin{figure}[h!]
\centering
\includegraphics[width=0.95\textwidth,height=0.85\textheight,keepaspectratio]{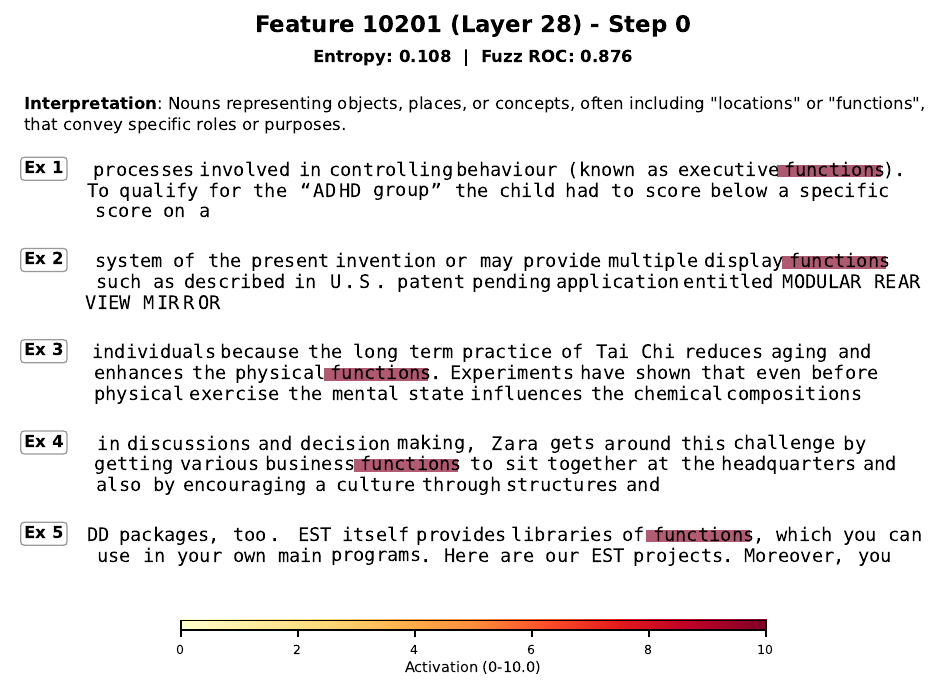}
\end{figure}

\end{document}